\documentclass{article}

\usepackage{microtype}
\usepackage{graphicx}
\usepackage{subfigure}
\usepackage{booktabs} %

\usepackage{hyperref}

\usepackage[accepted]{icml2024}

\usepackage{amsmath}
\usepackage{amssymb}
\usepackage{mathtools}
\usepackage{amsthm}

\usepackage{enumitem}
\setitemize{noitemsep,topsep=0pt,parsep=1.5pt,partopsep=0pt}

\usepackage[capitalize,noabbrev]{cleveref}

\theoremstyle{plain}
\newtheorem{theorem}{Theorem}[section]

\newtheorem{lemma}[theorem]{Lemma}

\theoremstyle{definition}
\newtheorem{definition}[theorem]{Definition}
\newtheorem{assumption}[theorem]{Assumption}
\theoremstyle{remark}

\newtheorem{example}{Example}[section]

\usepackage[textsize=tiny]{todonotes}

\usepackage{notations}
\usepackage{multicol}
\usepackage{multirow}
\usepackage{bm}
\usepackage{comment}
\usepackage{thmtools}
\usepackage{thm-restate}
\usepackage{bbm}
\usepackage[utf8]{inputenc} %
\usepackage[T1]{fontenc}    %
\usepackage{url} 
\usepackage{amsfonts}       %
\usepackage{nicefrac}       %
\usepackage{xcolor}         %
\usepackage{caption}
\usepackage{subcaption}
\usepackage{wrapfig,lipsum}
\usepackage{placeins}

\icmltitlerunning{A Sparsity Principle for Partially Observable Causal Representation Learning}

\begin{document}

\twocolumn[
\icmltitle{A Sparsity Principle for Partially Observable Causal Representation Learning}

\icmlsetsymbol{equal}{*}

\begin{icmlauthorlist}
\icmlauthor{Danru Xu}{uva}
\icmlauthor{Dingling Yao}{ista,mpi}
\icmlauthor{S\'ebastien Lachapelle}{samsung,mila}
\icmlauthor{Perouz Taslakian}{servicenow}
\icmlauthor{Julius von K\"ugelgen}{eth}
\icmlauthor{Francesco Locatello}{ista}
\icmlauthor{Sara Magliacane}{uva}

\end{icmlauthorlist}

\icmlaffiliation{uva}{University of Amsterdam}
\icmlaffiliation{ista}{Institute of Science and Technology Austria}
\icmlaffiliation{mpi}{Max Planck Institute for Intelligent Systems, T\"ubingen, Germany}
\icmlaffiliation{samsung}{Samsung - SAIT AI Lab, Montreal}
\icmlaffiliation{mila}{Mila, Université de Montréal}
\icmlaffiliation{servicenow}{ServiceNow Research}
\icmlaffiliation{eth}{Seminar for Statistics, ETH Z\"urich}
\icmlcorrespondingauthor{Danru Xu}{d.xu3@uva.nl}

\icmlkeywords{Machine Learning, ICML}
\vskip 0.3in
]

\printAffiliationsAndNotice

\begin{abstract}
\looseness-1 
Causal representation learning
aims at identifying high-level causal variables from perceptual data. Most %
methods assume that \textit{all} latent causal variables are captured in the high-dimensional observations.
We instead consider a \textit{partially observed} setting, in which each measurement only provides information about a subset of the underlying causal state.
Prior work has studied this setting with multiple domains or views, each depending on a \textit{fixed} subset of latents.
Here we focus on learning from \emph{unpaired} observations from a dataset with an %
\emph{instance-dependent} partial observability pattern.
Our main contribution is to establish
two identifiability results for this setting: 
one for {linear mixing functions} without parametric assumptions on the underlying causal model, and one for {piecewise linear mixing functions} with Gaussian latent causal variables. %
Based on these insights, we propose two methods for estimating the underlying causal variables by enforcing sparsity in the inferred representation.
Experiments on different simulated datasets and established benchmarks highlight the
effectiveness of our approach in recovering the ground-truth latents.

\end{abstract}

\section{Introduction}
\label{sec: introduction}
Endowing machine learning models with causal reasoning capabilities is a promising direction for 
improving their robustness, generalization, and interpretability~\citep{Spirtes_2000,pearl2009causality,peters2017elements}. Traditional causal inference methods assume that the causal variables are given a priori, but in many real-world settings, we only have unstructured,
high-dimensional observations of a causal system.
Motivated by this shortcoming, causal representation learning~\citep[CRL;][]{scholkopf2021toward} aims to infer high-level causal variables from low-level data such as images.

\looseness-1 A popular approach to identify (i.e., provably recover) high-level latent variables is (nonlinear) independent component analysis (ICA)~\citep{hyvarinen2016unsupervised, hyvarinen2017nonlinear, hyvarinen2019nonlinear, khemakhem2020variational}, which aims to recover independent latent factors from entangled measurements. Several works generalize this setting to the case in which the latent variables can have causal relations~\citep{
yao2022learning,brehmer2022weakly,lippe2022citris,lippe2023biscuit, ahuja2023interventional,ahuja2023multi, lachapelle2022disentanglement, lachapelle2023synergies, lachapelle2024nonparametric,  von2021self,von2023nonparametric,%
Liang2023cca,squires2023linear,buchholz2023learning,zhang2023identifiability}, establishing various {identifiability} results under different assumptions on the available data and the generative process.
However, most existing works assume that \textit{all} causal variables are captured in the high-dimensional observations. 
Notable 
exceptions include \citet{sturma2023unpaired} and \citet{yao2023multi} who study \textit{partially observed} settings with multiple domains (datasets) or views (tuples of observations), respectively, each depending on a \textit{fixed subset} of the latent variables.

In this work, we also focus on learning causal representations 
in such a \emph{partially observed} setting, where not necessarily all causal variables are captured 
in any given observation.
Our setting differs from prior work in two key aspects: (i) we consider learning from a dataset of \textit{unpaired partial} observations; and (ii) we allow for %
\emph{instance-dependent} partial observability patterns, meaning that each measurement depends on an unknown, varying (rather than fixed) subset of the underlying causal state.

\begin{figure*}[!t]
	\centering
	\includegraphics[width=0.8\textwidth]{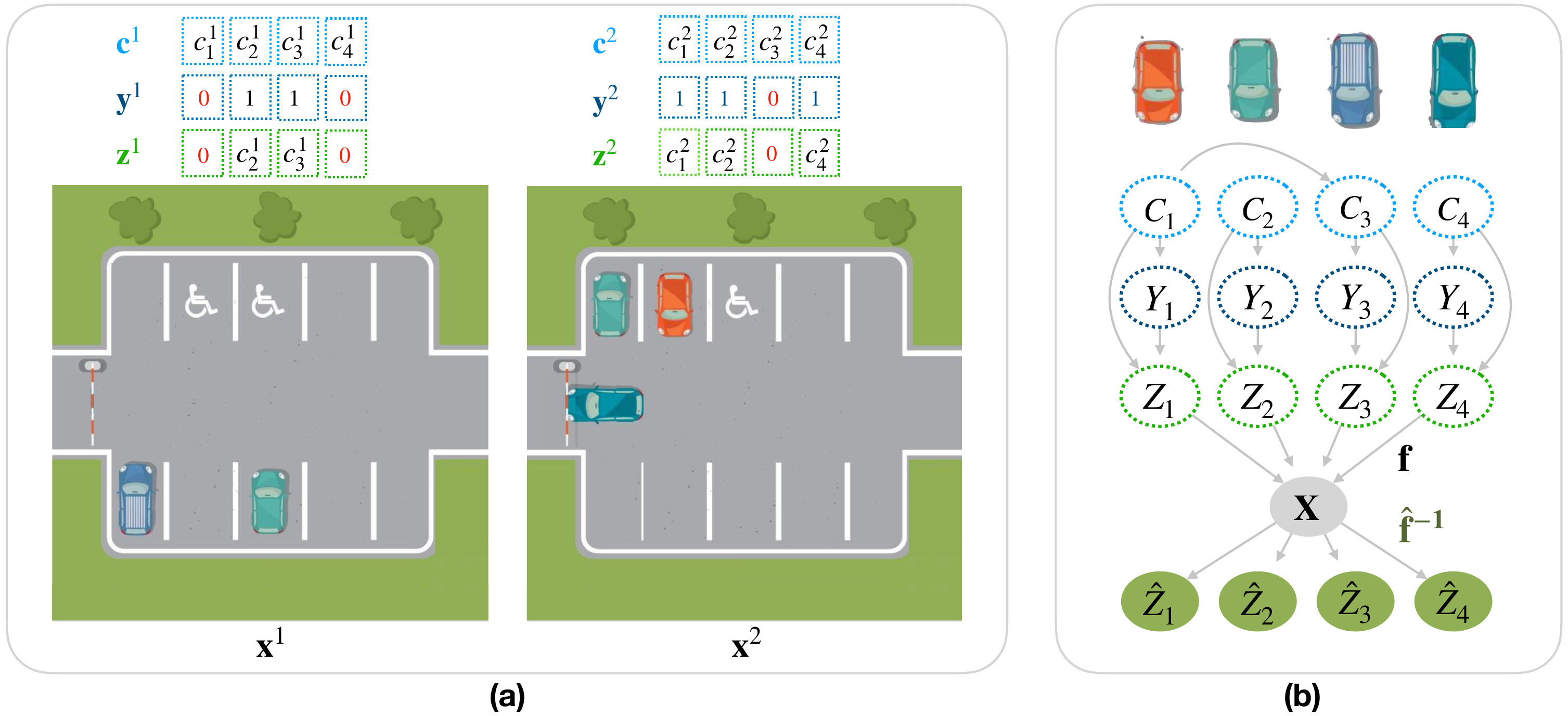} 
	\caption{ \textbf{(a)} Motivating example for the \emph{Unpaired Partial Observation} setting: a stationary camera taking pictures of a car park. We consider $\xb^1$ the image on day 1 and $\xb^2$ the image on day 2. The latent causal variables  $\cb^1$ and $\cb^2$ represent the positions of  four cars on each day. In $\xb^1$ only $Car2$ and $Car3$ are visible, while in $\xb^2$ all cars except $Car3$ are visible. This is represented by the ones in the binary mask variables $\yb^1$ and $\yb^2$. The combination of the values of the latent causal variables $\cb$ and the masks $\yb$ are the \emph{masked causal variables} $\zb$, which used by the mixing function $\fb$ to generate the images~$\xb$. \textbf{(b)}~Causal model of the setting, the dotted line variables are not directly observed, but they are measured only through the observation $\Xb$. Our goal is to learn a representation $\hat{\Zb}$ that identifies $\Zb$  up to permutation and element-wise transformation.}

 \label{fig: example}
\end{figure*}

This setting is motivated by real-world applications in which we cannot at all times observe the complete state of the environment, e.g., because some objects are moving in and out of frame, or are occluded. 
As a motivating example, consider a stationary camera that takes pictures of a parking lot on different days as shown in Fig.~\ref{fig: example}a. On different days, different cars are present in the parking lot, and the same car can be parked in different spots. Our task is to recover the position for each car that is present in a certain image.
In this setting, we only have one observation
for a given state of the system (i.e., one image per day), and the subsets of causal variables that are measured in the observation
(the parked cars), change dynamically across images. 
We highlight the following contributions:
\begin{itemize}[leftmargin=*]
    \item We formalize the \emph{Unpaired Partial Observations} setting for CRL, where each \emph{partial observation} captures only a subset of causal variables and the observations are  unpaired, i.e., we do not have simultaneous partial observations of the same state of the system.
    \item We introduce two theoretical results for identifying causal variables up to permutation and element-wise transformation under partial observability. Both results leverage a sparsity constraint. In particular, Thm.~\ref{thm: linear disentangle} proves
    identifiability 
    for linear mixing function and without parametric assumptions on the underlying causal model. 
    Thm.~\ref{thm: disentanglement piecewise} proves  identifiability %
    for the \emph{piecewise linear mixing function}, when the causal variables are Gaussian and 
    we can group observations by their partial observability patterns. %
    \item Finally, we propose two methods that implement these theoretical results and validate their effectiveness %
    with experiments on simulated data and image benchmarks, e.g. Causal3DIdent \citep{von2021self}, that we modify to test our partial observability setting. 
\end{itemize}

\section{Problem setting}
\label{sec: Problem set up}
In this section, we formalize the \emph{Unpaired Partial Observation} setting, in which we have a set of high-dimensional observations that are functions only of instance-dependent subsets of the true underlying causal variables.
This setting consists of four sets of random variables: the \emph{causal variables}  $\Cb$, the binary \emph{mask variables} $\Yb$ that represent if a variable is measured in a sample, the \emph{masked causal variables} $\Zb$ that combine the information from the causal variables and the masks, and the \emph{observations} $\Xb$. Our goal is to recover the masked causal variables 
 $\Zb$ up to permutation and element-wise transformation, solely from the observations $\Xb$, despite the instance-dependent  partial observability pattern. We show an example of a causal graph of the setting in Fig~\ref{fig: example}b and discuss each component in the following.

\paragraph{Causal variables $\Cb$.} 
We define our latent causal variables as a random vector $\Cb=(C_1,...,C_n)$ that takes values in $\Ccal=\Ccal_1 \times ... \times \Ccal_n \subseteq \RR^n$, which is an open, simply connected latent space. The causal variables follow a distribution with density $p(\Cb)$, which allows for causal relations between them. We assume that $p(\cb) \neq 0$ for all $\cb \in \Ccal$. %

\paragraph{Mask variables $\Yb$.} 
We use a binary mask random variable $\Yb =(Y_1, \dots, Y_n)$ 
with domain $\Ycal \subseteq \{0,1\}^n$ to represent the dynamic partial observability patterns, i.e., the causal variables that are measured in each of the samples. If $Y_i = 1$ then we consider the variable $C_i$ \emph{measured}, i.e. captured in the observation, otherwise it is considered \emph{unmeasured}.
We assume $\Yb$ follows $p(\Yb)$.
Further, we define the \emph{support index} random set %
$\Sb$ as the index of non-zero components of $\Yb$, i.e., $\Sb:=\{i \in [n]: Y_i \neq 0\}$. %
The support index set has a probability mass function $p(\sb)$ and support $\Scal$ defined as:
\begin{align*}
\mathbb{P}(\sb) := \mathbb{P}\left(\bigwedge_{j \in \sb} (\Yb_j = 1) \land \bigwedge_{j \not \in \sb} (\Yb_j = 0)\right)\, \ \ \\
\mathrm{and} \ \
\mathcal{S}:= \{\sb \subseteq [n] \mid p(\sb) > 0\}.
\end{align*}
This definition allows us to model also dependences between the components of $\Yb$ and masking behavior that depends on the values of the underlying causal variables.

\paragraph{Masked causal variables $\Zb$.} The \emph{masked causal variables} $\Zb=(Z_1,...,Z_n) \in \Zcal$ 
are a combination of the causal variables and the masks, and they are the latent inputs to the mixing function $\fb$ that we are trying to recover. In particular, they are  the Hadamard product of the causal variables with the  binary mask variable, i.e., $\Zb=\Yb \odot \Cb$. 
This means that for sample $i \in [ N ]$ and any causal variable $j \in [n]$, if the mask value $y^i_j$ is $1$, then the causal variable $c^i_j$ is measured and $z^i_j$ is $c^i_j$. Instead, if $y^i_j = 0$, then the causal variable is unmeasured and $z^i_j$ takes a fixed masked value $M_j$, which we will consider for simplicity to be $0$. Note that this is not equivalent to do-interventions, since masking variables does not influence any downstream variables, as an intervention would, as explained in an example in App.~\ref{app:relation_partial_obs_interventions}.
Finally, we assume that for all $\sb \in \mathcal{S}$, the probability measure $\mathbb{P}_{\Zb_{\sb} \mid \Sb=\sb}$ has a density w.r.t. the Lebesgue measure on $\mathbb{R}^{|\sb|}$. %

\paragraph{Observations $\Xb$.} We assume that observations $\Xb \in \Xcal \subseteq \RR^d$ are generated by mixing the masked causal variables $\Zb$ with the same \emph{mixing function} $\fb: \Zcal \rightarrow \Xcal$, i.e., $\Xb = \fb(\Zb)$. 
We refer to partitions of observations with the same unknown \emph{partial observability pattern}, i.e., with the same unknown value of $\Yb= \yb$, as \emph{groups}, and we assume that each observation $\xb^i$ for $i \in [N]$ is part of a group $g^i \in \Gcal$.

Our goal is to identify the masked causal variables $\Zb$ from a set of observations $\Xb$. %
In CRL we usually cannot recover the exact value of the latent variables, but we can only identify them up to some transformation. %
Our results guarantee that each ground truth  variable is represented by a single estimated variable up to a linear transformation. Similar notions of identifiability were used in previous works \citep{COMON1994, khemakhem2020variational, lachapelle2023synergies}.%
\begin{definition}
\label{def: identi one2one}
The ground truth representation vector $\Zb$ 
is said to be identified \emph{up to permutation and element-wise linear transformation} by a learned representation vector $\hat{\Zb}$ 
when there exists a permutation matrix $\Pb$ and an invertible diagonal matrix $\Db$ %
such that %
$\hat{\Zb}=\Pb\Db\Zb$ almost surely. 
\end{definition}

To prove our results, we describe the sufficient support index variability, which was originally defined by \citet{lachapelle2023synergies} for sparse multitask learning for disentanglement. %
\begin{assumption} (Sufficient support index variability \citep{lachapelle2023synergies})
\label{assump: sufficient support}
For all $i\in [n]$,
we assume that the union of the support indices $\sb$ that do not contain $i$ covers all of the other causal variables, or more formally:
    $$\bigcup_{\sb \in \Scal \mid i \notin \sb} \sb = [n] \setminus \{i\} \,.$$
\end{assumption}
This assumption avoids cases in which two variables are always missing at the same time, since then we would not be able to disentangle them from the observations.
A trivial set $\mathcal{S}$ that satisfies this assumption is $\mathcal{S} = {{1}, {2}, ..., {n}}$, which contains $n$ distinct masks.
We conjecture that if this assumption is not satisfied for all $i$, but only for blocks of variables, we would instead get block identifiability. 

\section{Identifiability via %
a Sparsity Principle}
\label{sec:disentanglement via sparisty}

In this section, we show how a simple sparsity constraint on the learned representations allows us to learn a disentangled representation of the ground truth variables. We first consider linear mixing functions and prove identifiability without any parametric assumption on the causal variables and while allowing for partial observability patterns that can depend on the value of the causal variables (Thm.~\ref{thm: linear disentangle}).%
We then investigate if this \emph{sparsity principle} can also identify variables for nonlinear mixing functions $\fb$. This is not the case in general, as we show in
Example~\ref{ex:example}. 

Our linear result hinges on the existence of a function $\gb$ such that the composition of $\gb$ and~$\fb$ is affine. Based on this intuition, we consider \emph{piecewise linear} mixing functions $\fb$, since for Gaussian causal variables they can be composed with an appropriate~$\gb$ to obtain an affine function. %
In this setting, we prove that we can learn a disentangled representation of the latent variables (Def.~\ref{def: identi one2one}) 
given that the masks are independent of the causal variables and given that we know the groups of the observations (Thm.~\ref{thm: disentanglement piecewise}).

\subsection{Linear Mixing Function}
We show that for linear mixing functions under a perfect reconstruction (Eq.~\ref{eq:zero_reconstruction2}), a simple \emph{sparsity constraint} on the learned representation (Eq.~\ref{eq:sparsity_constraint}) allows us to learn a disentangled representation of the ground truth latent variables.%
\begin{restatable}[Element-wise Identifiability for Linear $\fb$]{theorem}{thmlinear}
\label{thm: linear disentangle}
Assume the observation $\Xb = \fb(\Zb)$ follows the data-generating process in Sec.~\ref{sec: Problem set up}, where $\fb: \Zcal \to \Xcal$ is an injective linear function, and Ass.~\ref{assump: sufficient support} holds. 
Let $\gb: \Xcal\rightarrow \mathbb{R}^n$ be an invertible linear function onto its image and let $\hat{\fb}: \mathbb{R}^n \rightarrow \mathbb{R}^d$ be an invertible %
continuous function onto its image.
If both of the following conditions hold,
\begin{align}
    &\EE\norm{\Xb - \hat\fb(\gb(\Xb))}^2_2 = 0 \,,\quad\text{and} \label{eq:zero_reconstruction2}\\ &\EE\norm{\gb(\Xb)}_0 \leq \EE\norm{\Zb}_0 , \label{eq:sparsity_constraint}
\end{align}
then $\Zb$ is identified by $\hat\fb^{-1} (\Xb)$ up to a permutation and element-wise linear transformations (Def.~\ref{def: identi one2one}), i.e., $\hat\fb^{-1} \circ \fb$ is a permutation composed with element-wise invertible linear transformations on $\Zcal$. %
\end{restatable}

We provide a proof in App.~\ref{app:proofs:thm_linear} and now give an intuitive explanation for why it holds. 
The zero reconstruction loss ensures that no information is lost in the encoding $\gb(\Xb)$, which implies that $\gb(\Xb)$ is not sparser than $\Zb$.
Hence, incorporating Eq.~\eqref{eq:sparsity_constraint} as a constraint or regularization term in our methods enables our estimators to match the sparsity of the ground truth variables, which breaks indeterminacies due to rotations of the latent space. 

The idea of using a sparsity constraint or regularization is similar to previous work~\citep{lachapelle2023synergies} in the context of sparse multitask learning. In this paper we leverage these ideas in the distinct setting of partial observability. 
This result requires that the mixing function $\fb$, is injective, but not necessarily bijective.
Notably, it  does not require any parametric assumptions on the distribution of $\Zb$, thus allowing for causal relations or other statistical dependencies among the latent variables. Finally, this result also allows for mask variables that potentially depend on the values of the latent causal variables.

\subsection{Is sparsity enough for identification for nonlinear $\fb$?}
\looseness-1 Since linearity of $\fb$ is a strong assumption that may not hold in many applications, an obvious question is whether we can extend this result to nonlinear mixing functions. Unfortunately, this is not the case without making further assumptions, as demonstrated by the following example.

\begin{example}
\label{ex:example}
Consider $\Cb \sim \mathcal{N}(0, \Ib_2)$, where $\Ib_2$ is the identity matrix.
Assume an independent mask $\Yb$ with distribution $p(\Yb = \yb) = 1 / 4$ for any $\yb \in \{0,1\}^2$, satisfying  
Ass.~\ref{assump: sufficient support}. 
Let the nonlinear mixing function $\fb:\RR^2 \rightarrow \RR^2$ be %
    \begin{align}
        \fb(\zb) := \sinh(\Rb_{\frac{\pi}{4}}\zb) + \sinh(\Rb_{-\frac{\pi}{4}}\zb)\, ,
    \end{align}
where $\Rb_{\theta}$ is a rotation matrix. Consider $\hat\fb$ and $\gb$ to be the identity function, which trivially satisfy Eq.~\ref{eq:zero_reconstruction2}, since $\hat{\fb} \circ \gb (\Xb) = \Xb$. We show in Appendix~\ref{app:counterexample} that $\gb$ satisfies the sparsity constraint (Eq.~\ref{eq:sparsity_constraint}). However, despite satisfying all requirements of Thm~\ref{thm: linear disentangle} except for the linearity of $\fb$, we can show that each component of $\hat{\fb}^{-1} \circ \fb = \fb$ depends on both components of $\zb$; or in other words, the learned representation does not identify $\Zb$ up to permutation and element-wise transformations. We refer to App.~\ref{app:counterexample} for details. %
\end{example}

\subsection{Piecewise Linear Mixing Function}
In light of Example~\ref{ex:example}, we consider the role of linearity in Thm.~\ref{thm: linear disentangle}. We notice that it is a sufficient condition for ensuring that there exists a $\gb$ such that $\gb \circ \fb$ is affine on $\Zcal$. %
We then consider the question: even if $\fb$ is not affine itself, can we consider a restricted class of $\fb$ and latent variables $\Zb$ such that the
composition of $\fb$ and an appropriate $\gb$ is affine?
As a first step we consider a piecewise linear $\fb$ and  %
assume that the  causal variables are Gaussian and that the masks are independent from the causal variables. 

\begin{assumption}
\label{assump: gaussian dist}
We assume $\Cb$ follows a non-degenerate \emph{multivariate normal distribution}, i.e.
    $\Cb  \sim  N(\mub, \Sigmab)$,
where $\mub \in \RR^{n}$ and $\Sigmab \in \RR^{n\times n} $ is a positive definite matrix.
\end{assumption}
\begin{assumption}
\label{assump:independent masks}
We assume $\Cb$ and $\Yb$ are independent from each other, i.e. the partial observability pattern does not depend on the values of the causal variables.
\end{assumption}
These assumptions represent a classical linear Gaussian Structural Causal Model setting for the causal variables and a missing-at-random assumption in terms of which variables are measured in the observation. On the other hand, in our setting, we do not directly observe the causal variables or the masks, but we only observe them mixed in an observation.

Under these assumptions, the conditional distribution of the masked causal variables $\Zb$ given the binary mask vector $\Yb$ is defined as a multivariate normal distribution:
\begin{align*}%
\Zb~|~\Yb &\sim  N(\mub_{\Yb}, \Sigmab_{\Yb})\\
\text{where }  \mub_{\Yb}=(\mu_1 Y_1, \dots &, \mu_n Y_n ), \quad \Sigmab_{\Yb(ij)}=\Sigmab_{ij}Y_iY_j
\end{align*}
This distribution is a degenerate multivariate normal (De-MVN), i.e., a normal with a singular covariance matrix, if at least one of the causal variables is masked by $\Yb$.

Intuitively, we can leverage the Gaussianity of $\Zb|\Yb$ to enforce that the reconstructed $\hat{\Zb} |\Yb = \gb(\Xb)$ is also Gaussian. We now show that this allows us to identify the latent factors $\Zb$ 
through our \emph{sparsity constraint} on the learned representations (Eq.~\ref{eq:sparsity_constraint1}), given that we are able to partition the data according to the unknown value of $\Yb= \yb$, or in other words, given that we know the group $g_i$ for each observation $\xb_i$ for $i\in [N]$. 
The rationale of this requirement is that we do not need to know the value of the latent mask $\Yb$, but we do need to be able to separate observations that are generated by different 
distributions of $\Zb$,
so we can effectively enforce the Gaussianity constraint (Eq.~\ref{eq:group gauss}).

\begin{restatable}[Element-wise Identifiability for Piecewise Linear $\fb$]{theorem}{thmpiecewiselinear}
\label{thm: disentanglement piecewise}
Assume the observation $\Xb$ follows the data-generating process in Sec~\ref{sec: Problem set up}, Ass.~\ref{assump: sufficient support}, \ref{assump: gaussian dist} and \ref{assump:independent masks} hold and $\fb: \Zcal \to \Xcal$ is an injective continuous piecewise linear function. Let $\gb: \Xcal \rightarrow \mathbb{R}^n$ be a continuous invertible piecewise linear function and let $\hat\fb: \mathbb{R}^n \rightarrow \mathbb{R}^d$ be a continuous invertible %
piecewise linear function onto its image. %
If all following conditions hold:  
\begin{align}
    &\mathbb{E}\norm{\Xb - \hat\fb(\gb(\Xb))}^2_2 = 0 \,
    \label{eq:zero_reconstruction1},\\ 
    &\mathbb{E}\norm{\gb(\Xb)}_0 \leq \mathbb{E}\norm{\Zb}_0 \text{ and }
    \label{eq:sparsity_constraint1}\\ 
    &\gb(\Xb)~|~(\Yb=\yb) \sim N(\mub_{\yb},\Sigmab_{\yb}) \qquad \forall \yb \in \Ycal,
    \label{eq:group gauss}
\end{align}
for some $ \mub_{\yb} \in \mathbb{R}^{n}, \Sigmab_{\yb} \in \mathbb{R}^{n \times n}$, then $\Zb$ is identified by $\hat\fb^{-1}(\Xb)$, 
i.e., $\hat\fb^{-1} \circ \fb$ is a permutation composed with element-wise invertible linear transformations (Def.~\ref{def: identi one2one}).  
\end{restatable}

We provide the complete proof in App.~\ref{app:proof element-wise piecewise linear}. 
We first provide some results for a weaker notion of identifiability: \emph{identifiability up to affine transformations} (Def.~\ref{def: linear ident}). To this end, we first extend a theorem by \citet{kivva2022identifiability} from the case of non-degenerate to potentially \emph{degenerate} multivariate normal variables (Thm.~\ref{adap_thm:identif-inv-affine}). Such variables are crucial in our setting because partial observability potentially introduces degenerate cases. The crux of our proof involves handling the case of degenerate variables that do not have probability density.
We then show that
given the information of the binary mask $\Yb$, we can identify the latent factors $\Zb$ up to an affine function $\hb_{\Yb}$ (Lemma~\ref{lemma: affine given mask}). We then show that all of these affine functions can be represented by a single affine function $\vb:= \hat{\fb}^{-1}(\fb(\Zb))$ defined on $\Zcal$.

Compared to linear case in Thm.~\ref{thm: linear disentangle}, the additional constraint in this case is \emph{Gaussianity} on both $\Zb|\Yb$ and the estimator $\gb(\Xb)|\Yb$. This ensures that the composition of two piecewise linear functions $\gb$ and $\fb$ remains affine on $\Zcal$, extending the results from linear $\fb$ to piecewise linear $\fb$.

\paragraph{Non-zero mask values.} %
Our theoretical analysis assumes that we mask the unmeasured latent variables with a mask value of $0$, i.e. when $\yb^i =0$ for $i \in [N]$, we set $\zb_i = 0$. %
One can wonder whether allowing to set $\zb^i = \Mb$ when $\yb_i = 0$ for some potentially nonzero constant vector $\Mb$ would make the model more expressive. It turns out that this is not the case, since the decoder $\fb$ can always shift $\Zb$ in arbitrary ways, making the specific value of $\Mb$ irrelevant. %

\section{Implementation}
\label{sec: implementation}

We implement our two  theoretical results asconstrained optimization problems in Cooper
~\citep{gallegoPosada2022cooper}.
We approximate the sparsity constraint, i.e. Eq.~\ref{eq:sparsity_constraint} in Thm.~\ref{thm: linear disentangle} and Eq.~\ref{eq:sparsity_constraint1} in Thm.~\ref{thm: disentanglement piecewise}, by replacing the $L_0$ norm with $L_1$ norm, which is differentiable except at zero. In practice, the $L_1$ norm of the ground truth variables $\Zb$ is unknown, so we instead set a hyperparameter $\epsilon$ for the sparsity constraint
$\mathbb{E}\norm{\gb(\Xb)}_1 \leq \epsilon$. 
In our experiments, we use $\epsilon=0.01$ or $0.001$, details are provided in App.~\ref{app: sensitivity analysis}.

For linear $\fb$, we can reconstruct the latent variables directly from a dataset of observations $\{\xb^i\}_{i \in [N]}$ by minimizing the reconstruction error and adding the sparsity constraint (Thm~\ref{thm: linear disentangle}).
For piecewise linear $\fb$, Thm.~\ref{thm: disentanglement piecewise} requires that we know how to partition the data with the same partial observability \emph{at training time}, i.e., we have information about the group of each observation.
At test time, we have already learned $\gb$, so we can instead use only the observations without the group information. %
To encourage the Gaussianity condition on $\gb(\Xb)|\Yb = \yb$ (Eq.~\refeq{eq:group gauss}) in Thm.~\ref{thm: disentanglement piecewise}, we add two 
regularization terms %
that push the estimated skewness of each latent variable in each group to be 0 and the estimated kurtosis to be 3, which are the values of these moments in the Gaussian distribution.
We learn the encoder $\gb_\psi$ and the decoder $\hat\fb_\theta$ by solving the following optimization problem:
\begin{align}
\label{equ: loss_pw}
     \min_{(\theta, \psi)} & \frac{1}{N} \sum_{i \in [N]}
     \norm{\xb^i - \hat\fb_\theta(\gb_\psi(\xb^i))}^2_2  \nonumber \\
&+ \sum_{g \in \Gcal} \Big(\left|\mathrm{\widehat{skew}}_{g}\left(\gb_\psi(\xb)\right)\right| 
 + \left|\mathrm{\widehat{kurt}}_{g}\left(\gb_\psi(\xb)\right)-3\right|
 \Big) 
\nonumber \\
    &\text{subject to: \quad}
    \frac{1}{nN} \sum_{i \in [N]} \norm{\gb_\psi(\xb^i)}_1 \leq \epsilon, 
\end{align}
where $\mathrm{\widehat{skew}}_{g}\left(\gb_\psi(\xb)\right)$ and $\mathrm{\widehat{kurt}}_{g}\left(\gb_\psi(\xb)\right)$ are the estimated skewness and estimated kurtosis for group $g \in \Gcal$.
We solve both problems with the extra-gradient version of Adam \cite{gidel2018variational}. 
We provide all details in App.~\ref{app:implementation}.

\section{Experimental Results}
\label{sec: Experiments}

We perform three sets of experiments, one with numerical data in Sec.~\ref{sec:numericalexp} and two with image datasets, a dataset with multiple balls in Sec.~\ref{sec:multiballexp} and \emph{PartialCausal3DIdent}, a partially observable version of Causal3DIdent \citep{von2021self} in Sec.~\ref{sec:causal3didentexp}. We provide all the code for our method and the experiments at \href{https://github.com/danrux/sparsity-crl}{https://github.com/danrux/sparsity-crl}.

Following previous work \citep{hyvarinen2019nonlinear, khemakhem2020variational}, we report the \emph{mean coefficient of determination (MCC)} to assess that the learned representations match the ground truth up to a permutation and element-wise linear transformations. 
This metric is based on the Pearson correlation matrix $\mathrm{Corr}^{n\times n}$ between the learned representations $\hat{\Zb}=\gb(\Xb)$ and ground truth masked causal variables $\Zb$. 
Since our results are up to permutation, 
we compute the MCC on the permutation $\pi$ that maximizes the average of $|\mathrm{Corr}_{i,\pi(i)}|$ for each index of a ground truth variable $i \in [n]$, i.e. MCC=$\frac{1}{n}\max_{\pi \in perm([n])} \sum_{i=1}^{n}|\mathrm{Corr}_{i,\pi(i)}|$. We denote the correlation matrix with the permutation $\pi$ as $\mathrm{Corr}^{n\times n}_{\pi}$.

\subsection{Numerical Experiments}
\label{sec:numericalexp}
We generate numerical data following  Sec.~\ref{sec: Problem set up}. We investigate varying the partial observability patterns, the underlying causal model and the type of mixing function $\fb$. 
For the encoder $\gb$ and the decoder $\hat{\fb}$, we use 7-layer MLPs %
with $[10, 50, 50, 50, 50, 10]\times n$ %
units per layer, where $n$ is the number of causal variables, with LeakyReLU activations for each layer except the final one in the piecewise linear case. We apply batch normalization to control the norm of $\gb(\Xb)$. For simplicity, for most experiments we set the output dimension of $\gb$ to $n$. In an ablation in App.~~\ref{app: over-parameterization} we set the output dimension of $\gb$ to an overestimate of $n$, showing that our method learns to only use $n$ dimensions and provides similar performance to the other experiments.
In each setup, we average over 20 random seeds.

We experiment with different ratios of measured variables 
$\rho=\{1 \textsc{var}, 50\%, 75\%\}$, where $1 \textsc{var}$ is one measured variable, while $50\%$ and $75\%$ are the percentages of measured variables in each sample.
Based on each $\rho$ we predefine a set of $K$ masks $\Yb$ that satisfies Ass.~\ref{assump: sufficient support}. %
As discussed in Sec.~\ref{sec:disentanglement via sparisty}, 
while the theoretical results assume a mask value of $\Mb = \mathbf{0}$, the specific value used is inconsequential from the theoretical point of view. On the other hand, the optimization procedure improves for mask values that are out of distribution w.r.t. the unmasked distribution of each variable, since it is then easier to recover which variables are masked in each group. 
We investigate different masks values and consider $\Mb=\mub+ \delta \cdot \sigma$, where $\mub, \sigma \in \RR^n$ are the mean and standard deviation of $\Zb$ and $\delta \in \{0,2,3,5,10\}$. 

We consider $n=\{3,5,10,20,40\}$ causal variables $\Cb$. In each experiment, we  generate a random directed acyclic graph $\Dcal$ from a random graph Erd\"os-R\'enyi (ER)-$k$ model, where $k\in \{0,1,2,3\}$ and ER-$k$ is a graph with $n\cdot k$ edges. In particular, ER-0 implies independent $\Cb$ and therefore $\Zb$. 
Based on $\Dcal$, we consider three types of structure causal models (SCM): i) a \emph{linear Gaussian SCM} where edge weights are sampled uniformly from $[-2,-0.5]\cup[0.5,2]$ and we use standard Gaussian noises; ii) a \emph{linear exponential SCM}, where we have a similar setup, but with exponential noises with scale 1; iii) a \emph{nonlinear SCM}, where we simulate a nonlinear function with a linear layer, a sigmoid activation and a fully connected layer with 100 hidden units, %
 where edge weights are sampled uniformly from $[-2,-0.5]\cup[0.5,2]$ and we use standard Gaussian noises. %

\begin{figure*}[t]
\centering        
        {
        \includegraphics[width=0.3\textwidth]{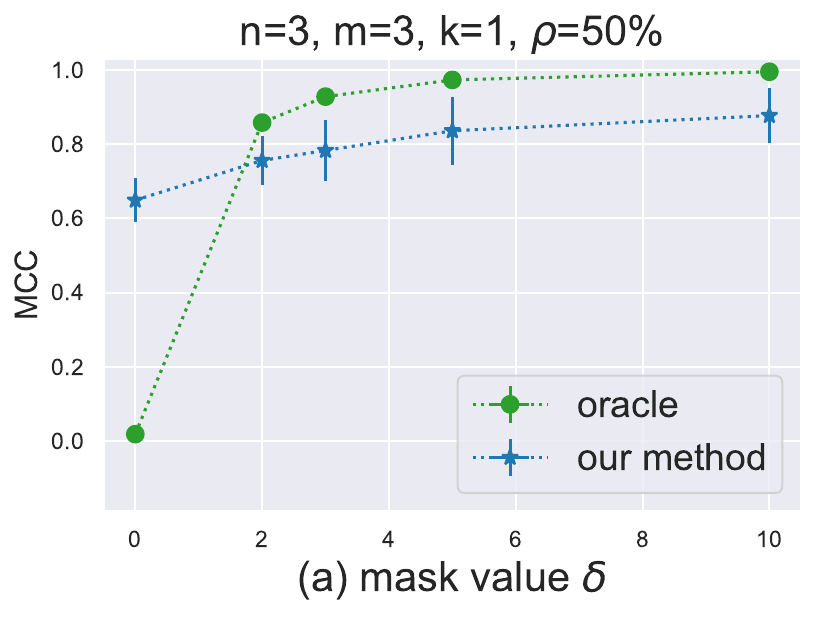} 
        \hspace{0cm} 
        \includegraphics[width=0.3\textwidth]{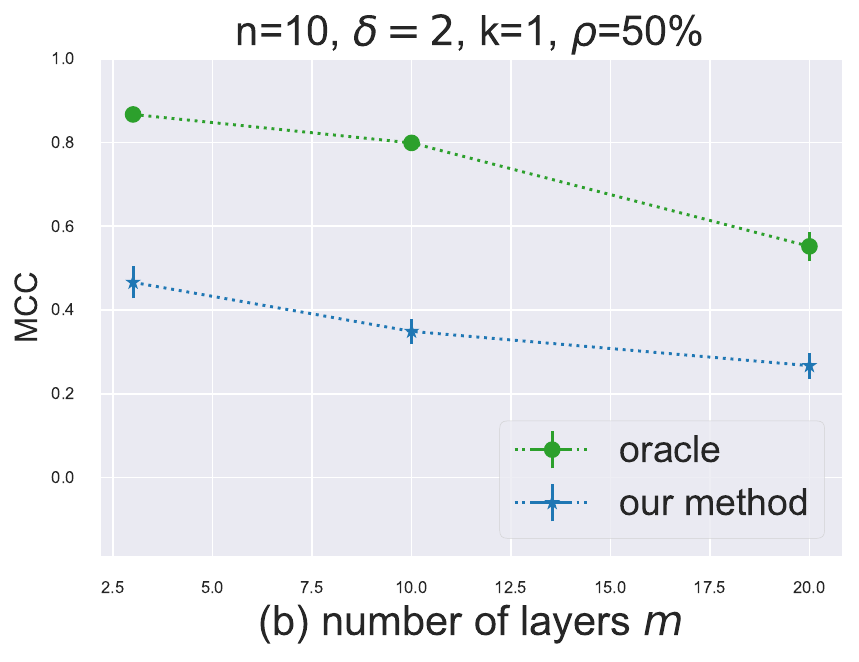}
        \hspace{0cm} 
        \includegraphics[width=0.3\textwidth]{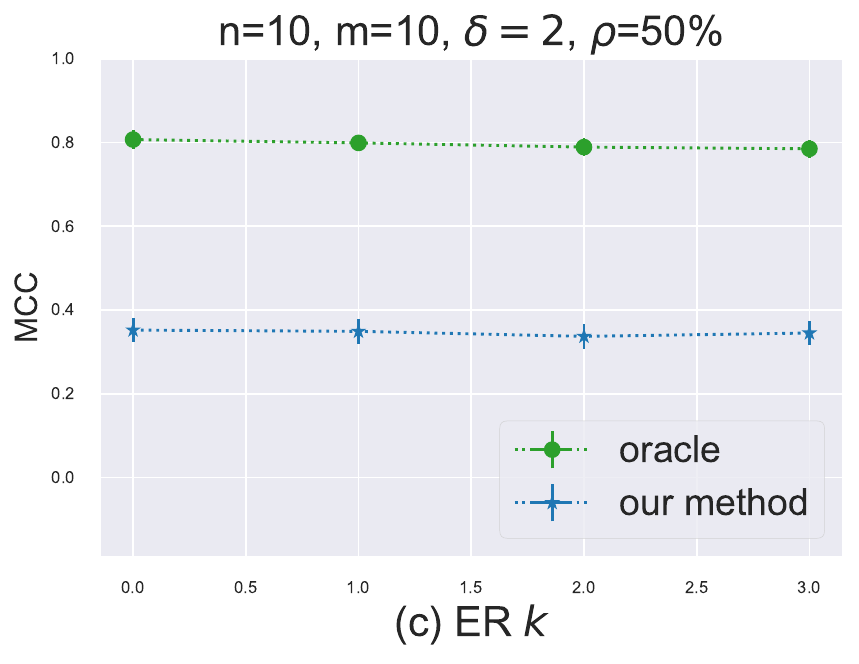}
        \hspace{0cm}
        \includegraphics[width=0.3\textwidth]{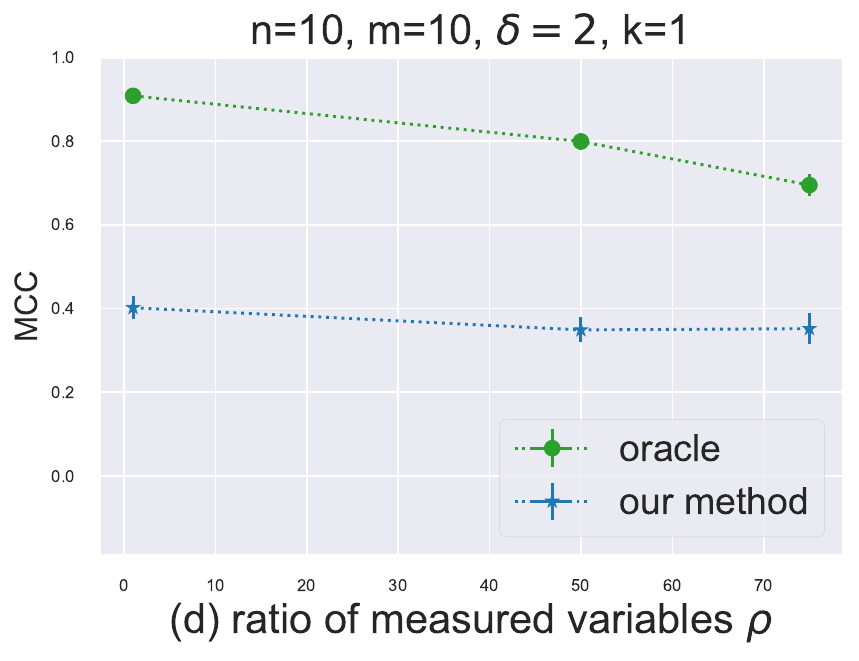}
        \hspace{0cm} 
        \includegraphics[width=0.3\textwidth]{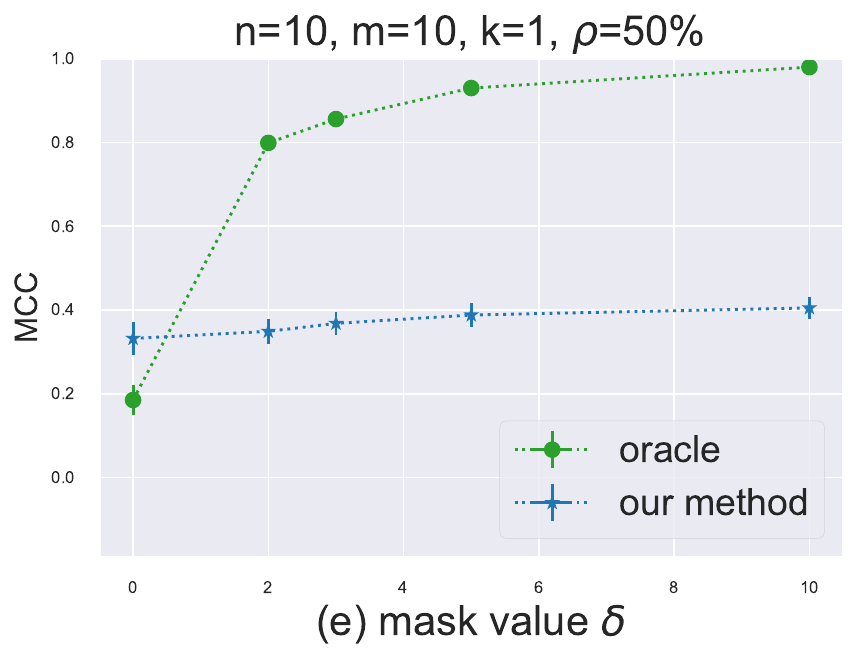}
        \hspace{0cm} 
        \includegraphics[width=0.3\textwidth]{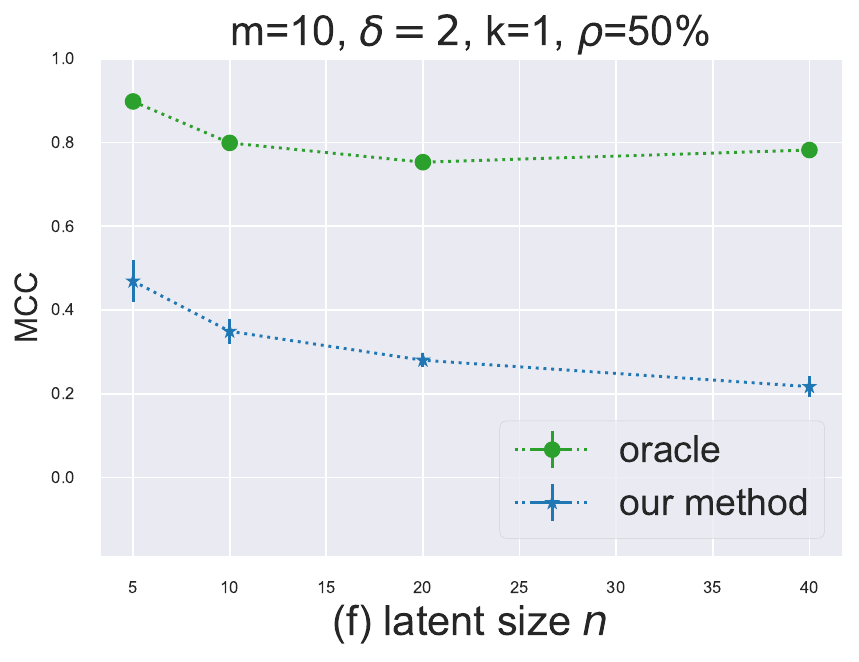}
        \hspace{0cm}
         }
        \caption{Results for different parameters in the piecewise linear numerical experiments. Our method implements  Eq.~\ref{equ: loss_pw} based on the group information. The \emph{oracle} method implements the same loss, but with as additional information  the mask $\yb_i$, which it uses to  assign a low variance to the masked variables in each sample for the skewness and kurtosis regularization terms. This method showcases the potential of our theoretical results with a stronger Gaussianity constraint.}.%
        
        \label{fig:num_pw}
\end{figure*}

\begin{table}[t]
\caption{Results for the numerical experiments for  linear mixing functions with $\delta=0$.
The bold font indicates which parameters are varying in each block of rows. 
}
\label{tab: num_linear}
\begin{center}
\begin{small}
\begin{sc}
\begin{tabular}{ccccc}
\hline
$n$ & $k$ & SCM &  $\rho$ & \textbf{MCC}     \\ \hline
\textbf{5}    & 1 &Lin. Gauss  & 50 \%  &  0.997$\pm$0.002 \\
\textbf{10}   & 1 &Lin. Gauss &  50 \%   & 0.996$\pm$0.001\\
\textbf{20}   &  1 & Lin. Gauss  & 50 \%  & 0.987$\pm$0.029 \\
\textbf{40}   & 1 &Lin. Gauss   & 50 \%   & 0.714$\pm$0.153 \\
\hline
10    & \textbf{0} & Indep. Gauss  & 50 \%  &  0.998$\pm$0.001 \\
10    & \textbf{1} &Lin. Gauss  & 50 \%   &  0.996$\pm$0.001 \\
10   &  \textbf{2} &Lin. Gauss & 50 \%   &  0.904$\pm$0.113\\
10   &  \textbf{3} &Lin. Gauss  & 50 \%   & 0.793$\pm$0.142\\
\hline
10    & \textbf{0} & Indep. Exp  & 50 \%   &  0.998$\pm$0.001 \\
10   &  \textbf{1} & Lin. Exp   & 50 \%   & 0.998$\pm$0.002\\
10    & \textbf{2} & Lin. Exp  & 50 \%  &   0.910$\pm$0.108 \\
10   & \textbf{3} &Lin. Exp & 50 \%   & 0.825$\pm$0.123  \\
\hline
10   &  \textbf{1} & Nonlinear  & 50 \%  &  0.997$\pm$0.001\\
10   &  \textbf{2} & Nonlinear  & 50 \%  & 0.997$\pm$0.001\\
10   &  \textbf{3} & Nonlinear  & 50 \%  & 0.996$\pm$0.001\\
\hline
10    & 1 &Lin. Gauss  & \textbf{1\text{var}}  &  0.998 $\pm$0.002  \\
10   &  1 &Lin. Gauss & \textbf{50 \%}   & 0.996$\pm$0.001 \\
10   &  1 &Lin. Gauss  & \textbf{75 \%}  & 0.877 $\pm$0.096\\
\hline
\end{tabular}
\end{sc}
\end{small}
\end{center}
\end{table}

\paragraph{Results for linear mixing function (Thm.~\ref{thm: linear disentangle}).}
We use a fully connected layer to model the linear mixing function $\fb$. We show the performance, measured in the average MCC over three random seeds in \cref{tab: num_linear}. In the first four rows, we investigate how the number of the latent causal variables influences the performance for a linear Gaussian SCM with an average degree $k = 1$ and a ratio of measured variables $\rho=50 \%$. In this case, the method achieves excellent performances for smaller $n$, but these degrade as $n$ increases. In the second group of rows, we consider the effect of $k$, the average degree in the causal graph in the linear Gaussian case for $n=10$ causal variables and $\rho=50 \%$. Also in this case, the performances are excellent for low $k$, including $k=0$ which represents independent variables, but they degrade with higher $k$. The third and fourth group of rows show how the performance varies for different $k$ for the linear exponential and  nonlinear SCM, showing a similar performance. Finally, we show the results for varying ratio of measured variables $\rho$, showing a small degradation when we measure more variables at the same time. Intuitively, measuring a smaller number of variables for each sample is the easier setting for disentangling them from the observations.

\paragraph{Results for piecewise linear mixing function (Thm.~\ref{thm: disentanglement piecewise}).} 
For the piecewise linear $\fb$, we use a $m=\{3,10,20\}$-hidden-layer MLP with $m-1$ LeakyReLU ($\alpha=0.2$) activation functions and a final linear layer, to model the piecewise linear mixing function. The number for layers $m$ in the $\fb$ mixing function is a proxy for the complexity of the function $\fb$, a linear function has $m=0$, while the higher the $m$ the higher the non-linearity. Following~\citep{lachapelle2022disentanglement}, the weight matrices are sampled from a standard Gaussian distribution and  are orthogonalized by their columns to ensure $\fb$ is injective. In this setting, we evaluate with a linear Gaussian SCM, since this satisfies Ass.~\ref{assump: gaussian dist} in Thm.~\ref{thm: disentanglement piecewise}.

We start by showing  results in Fig.~\ref{fig:num_pw}a for the simple case of number of causal variables $n=3$ and number of  layers $m=3$. %
Through optimization of Eq.~\eqref{equ: loss_pw}, our approach achieves good performances in these simple scenarios. However, in more complex cases,  %
e.g. more latent variables or more complicated $\fb$, there is a decline in performance, as shown in  Fig.~\ref{fig:num_pw}b-f. %
We hypothesize that the reason for this drop is that the estimated skewness and kurtosis cannot guarantee Gaussianity, which is crucial to ensure identifiability in Thm.~\ref{thm: disentanglement piecewise}. We show that this is empirically the case for $n=5$ in Fig.~\ref{fig:non_gaussian} in App.~\ref{app: oracle}.
We attribute this issue to the per-group sample variance used to calculate both sample skewness and kurtosis. We test this assumption and potential of our theoretical results, by comparing with an \emph{oracle} that has access to the masks $\yb^i$ for each observation $\xb^i$. This information is used to set the group sample variance to a low value for the masked variables in each group. As shown in Fig.~\ref{fig:num_pw}b-f this is effective in improving the performances of our sparsity constraint.

We test the performance of our method and the oracle for various parameters in Fig.~\ref{fig:num_pw}. In particular, we see in Fig.~\ref{fig:num_pw}a and Fig.~\ref{fig:num_pw}e that when the distance $\delta$ between masked and unmasked variables increases, it enables a more distinct separation of $\Zb$, providing better identification results. Similar to the linear case, we see that an increase in complexity, e.g., in the number of causal variable $n$, as shown in Fig.~\ref{fig:num_pw}f, or the number of layers $m$, as shown in Fig.~\ref{fig:num_pw}b, lowers the performance. Similarly, as shown in Fig.~\ref{fig:num_pw}d, the performance drops when the ratio of measured variables  $\rho$ increases.
Interestingly, the average degree of the causal graph $k$ does not have an impact, as shown in Fig.~\ref{fig:num_pw}c.
We provide more results and  visualizations in App.~\ref{app:experiments_numerical}.
In App.~\ref{app: graph_learning} we show the results of applying standard causal discovery methods on the learned representations, assuming the causal Markov and faithfulness assumption, as well as causal sufficiency.Intuitively, the closer the learned representations are to the ground truth, the more accurate are the learned causal relations between them.

\subsection{Image dataset: Multiple Balls} 
\label{sec:multiballexp}
We create a new image dataset in which we render in a 2D space $b$ moving balls, as shown in Fig.~\ref{fig:balls} in App.~\ref{app:experiments_multipleballs}. Our latent causal variables are the $(x,y)$ position of each ball $\bb$, which we model as Gaussian. We consider two settings: i) a \emph{missing ball} setting in which the balls can only move along the $x$-axis and they can move out of view, and ii) a \emph{masked position} setting in which the balls can move freely inside the frame and each of their coordinates can be masked by being set to an unknown, but fixed, $\Mb$ value.
Both of these settings showcase possible applications of our approach.
The first setting is a simplified version of a setting with a fixed camera capturing a set of objects that might move out of view.
The second setting is a simplified version of occlusion, in which an occluded object is not captured in the image, but it still interacts with other objects.

The \emph{missing ball} setting represents the intuitive setting for partial observability, i.e. when an object is out of the frame or occluded. While we model each object with two causal variables, its $x$ and $y$ coordinates,  our methods do not allow that masks for two variables are deterministically related (Ass.~\ref{assump: sufficient support}). Thus, we constrain the balls to move only on the $x$-axis, which we then identify from the observations. 
In order to test also the identifiability of each variable of the same object, we devise the \emph{masked position} setting. In this setting, we can still use our method with a non-zero mask value for each variable, which in this case represents a specific value for one of the $x$ or $y$ coordinates of the ball. 
We generate datasets for both settings by varying the number of balls $b = 2, 5, 8$. 
We use a predefined set of $K$ masks $\yb$ that satisfies Ass.~\ref{assump: sufficient support}.
 For the missing ball setting, we generate the $x$-coordinates of each ball %
 from a truncated normal distribution $\Ncal(0.5,0.1^{2})$ with bounds $(0.1,0.9)$. For the masked position setting, we generate the $x$ and $y$-coordinates of each ball %
 $i\in[b]$ 
 independently from a truncated 2-dimensional normal distribution $\Ncal(\mub_i,\Sigmab_i)$ with bounds $(0.1,0.9)^2$, where $\mub_i\sim Unif(0.4,0.6)^2$, and $\Sigmab_i= ((0.01,0.005)(0.005,0.01))$ for all groups. For both settings, we generate the masked causal variables as $(\Zb_k)^K_{k=1} = \yb \cdot \Cb + (1- \yb) \cdot \Mb \in \RR^{K \times n}$, where $\Mb=\mathbf{0}$ for the missed ball setting, and $\Mb = \mub_i + \delta \cdot \sigma_i$ for the masked position setting. Instead of a fixed-size training dataset, we generate images online until convergence. We provide more details in App.~\ref{app:experiments_multipleballs}.

\paragraph{Results.} 
As illustrated in Table~\ref{tab: ball_mcc}, while the MCC decreases with an increase in the number of balls $b$, all MCCs remain above $0.90$. The results are consistently higher in the missing ball setting, where there are $b$ variables to reconstruct, while in the masked position setting there are $2b$. 
\begin{table}
\centering
\small
\caption{Results for the multiple balls dataset.\label{tab: ball_mcc}}
\begin{tabular}{ccc}
\hline
$b$ & MCC Missing   & MCC Masked  \\ \hline
2     & 0.963$\pm$0.012 & 0.946$\pm$0.005 \\
5     & 0.950$\pm$0.011 & 0.939$\pm$0.003 \\
8     & 0.928$\pm$0.004 & 0.901$\pm$0.002 \\ \hline
\end{tabular}
\end{table}
Additionally, in the masked position setting, we have dependence between the $x$ and $y$ coordinates, which can make the problem more challenging.

\subsection{Image dataset: PartialCausal3DIdent}

\label{sec:causal3didentexp}
We explore the capability of our method on a partial observability version of Causal3DIdent~\citep{von2021self} that we create.
Causal3DIdent collects images of 7 object classes rendered from 10 %
causal variables, including object color, object positions, spotlight positions, etc. Each latent variable is rescaled into an interval of $[-1, 1]$. The mixing function, i.e., the rendering process, is not piecewise linear (which violates the assumption in Thm.~\ref{thm: disentanglement piecewise}). We still evaluate our piecewise linear method, following the intuition that non-linear functions can be approximated up to an arbitrary precision by an adequate number of linear pieces.

\paragraph{Dataset generation.}
Since Causal3DIdent is fully observable, we sample from it to create \emph{PartialCausal3DIdent}, %
a dataset in which some of the latent variables are masked to a predefined value.
For each datum, we sample a latent vector $\Cb \sim \Ncal(0, \sigma^2 \Ib)$ of $n=10$ independent causal variables. We apply the set of $K$ predefined masks $\yb \in \{0, 1\}^{K \times n}$ %
to get a set of masked latent variables $(\Zb_k)^K_{k=1} = \yb \cdot \Cb + (1- \yb) \cdot \Mb \in \RR^{K \times n}$. We define the masked value as the maximum of the support area (which is 1 for all latents). 
The ratio of the measured variables $\rho$ varies from 10\% (only one latent is measured) to 100\% (all latents are measured). The average $\rho$ is set to $50\%$. After obtaining the masked latent variables, we retrieve $K$ corresponding images from the dataset, based on the index searching scheme provided by~\citet{von2021self}. 
\paragraph{Results.} 
We evaluate our method separately on each object class in \emph{PartialCausal3DIdent} and show the results in Table~\ref{tab:c3di_res}. Although the performance fluctuates slightly across classes, our method achieves a high MCC over $80\%$ for all classes, which verifies that our approach is empirically applicable even on highly nonlinear high dimensional data.
We provide all details and  ablation studies on $\delta$ in App.~\ref{app:experiments_causal3DIdent}.

\begin{table}[t]
    \begin{center}
    \begin{small}
    \caption{MCC on \emph{PartialCausal3DIdent} over each object class, with masking distance $\delta=10$ for all latent variables.} %
    \label{tab:c3di_res}
    \begin{sc}
    \begin{tabular}{lll}
    \toprule
    \textbf{Object class id}  &  \textbf{0} & \textbf{1} \\
    \textbf{MCC} (mean $\pm$ std) &  $0.842 \pm 0.018$ & $0.804 \pm 0.023$\\
    \midrule
    \textbf{Object class id} & \textbf{2} & \textbf{3} \\
    \textbf{MCC} (mean $\pm$ std) &  $0.828 \pm 0.014$ & $0.820 \pm 0.009$\\
    \midrule
    \textbf{Object class id} & \textbf{4} & \textbf{5} \\
    \textbf{MCC} (mean $\pm$ std) &  $0.837 \pm 0.020$ & $0.821 \pm 0.033$ \\
    \midrule
    \textbf{Object class id} & \textbf{6} & \textbf{avg. mean} \\
    \textbf{MCC} (mean $\pm$ std) &  $0.858 \pm 0.005$ & $0.832 \pm 0.016$\\
    \bottomrule
    \end{tabular}
    \end{sc}
    \end{small}
    \end{center}
\end{table}

\section{ Related Work}
\label{sec: relwork-long}
Most closely related to our work are recent identifiability studies, which also explicitly %
learn causal representations in a partially observable setting. \Citet{yao2023multi} consider learning from tuples of simultaneously observed views, which depend on different fixed (potentially overlapping) subsets of latents with modality-specific mixing functions, and prove identifiability results for different blocks of shared content variables~\citep{von2021self}. Compared to our setting, such 
paired multi-view data may be harder to obtain.
\citet{sturma2023unpaired} study an unpaired, multi-domain setup, in which observations from each domain depend on a different fixed subset of latents, and show identifiability of the causal representation and graph in the fully linear case.
This setting resembles our results for the piecewise linear case in Thm.~\ref{thm: disentanglement piecewise}, where we assume we have the group information for each observation, which can be seen as a single domain. On the other hand, for the linear case in Thm.~\ref{thm: linear disentangle}, we do not need the group information, hence we also allow for mixtures of data from multiple domains.

Other works in an i.i.d.\ setting can be viewed as modelling partial observability \textit{implicitly} by restricting the graph connecting latent and observed variables and establishing identifiability for linear~\citep{adams2021identification, cai2019triad,silva2006learning, xie2020generalized, xie2022identification} or discrete~\citep{kivva2021learning} settings.
In our work, we consider the case in which either the causal model or the mixing are not linear and do not constrain the connectivity between $\Zb$ and~$\Xb$. 

Not considering partial observability, other works on CRL aim to also learn the causal graph from different types (hard/soft, single/multi-node) of interventions in linear~\citep{squires2023linear,bing2023identifying}, partially~\citep{buchholz2023learning,ahuja2023interventional,ahuja2023multi,zhang2023identifiability}, or fully nonlinear~\citep{von2023nonparametric,varici2023score} settings.
We focus instead on only recovering the latent causal variables without access to interventional data.

Other works have also explored the piecewise linear setting for identifiability, including with the assumption of Gaussian causal variables.
In particular, Thm.~\ref{thm: disentanglement piecewise} resembles one of the identifiability results from \citet{kivva2022identifiability} which assumes $\Zb$ is a mixture of \textit{non-degenerate} Gaussians and $\fb$ is a piecewise linear function. We note that, in Thm.~\ref{thm: disentanglement piecewise}, $\Zb$ is also a mixture of Gaussians, where the ``cluster index'' corresponds to $\Yb$. However, this mixture contains components which are degenerate, in the sense that their covariances might be singular (this occur when $\Yb_i = 0$ for some $i$). This prevents us from applying the result of \citet{kivva2022identifiability} directly to our setting. On the other hand, although we allow for degenerate components, our result assumes knowledge of the groups $g$, unlike \citet{kivva2022identifiability}.

Prior work has also leveraged sparsity in representation learning, for example via a Spike and Slab prior~\citep{tonolini2020variational,moran2022identifiable}, assuming structural sparsity on the support of Jacobian of nonlinear mixing function~\citep{zheng2022identifiability,zheng2023generalizing}, exploiting the sparsity constraint on the linear mixing function~\citep{ng2023identifiability}, or by relating the learnt representation to multiple tasks, each depending only on a small subset of latents~\citep{lachapelle2023synergies,fumero2023leveraging}. Other work that is closely related to ours is work by~\citet{lachapelle2022disentanglement,lachapelle2024nonparametric}, who have proposed a sparsity principle for identifiable CRL in interventional and temporal settings, motivated by the sparse mechanism shift hypothesis~\citep{scholkopf2021toward,perry2022causal}.

Among these works leveraging sparsity, \citet{lachapelle2023synergies} introduce a  sparsity principle that inspired our work, but for the multi-task setting, where each task depends on a subset of variables. More precisely, they assume that the connections $W$ between tasks $Y$ and representations $f(X)$ are sparse, i.e. $Y=W f(X)$ and $W$ is a column-sparse matrix (some of the columns are filled with zeros). In our setting, the sparsity assumption is on the representation itself, i.e. each causal variable has a positive probability of being masked. While the proof strategies are similar, our result also had to address the piecewise linear case. 
Other work that is closely related to ours is work by~\citet{lachapelle2022disentanglement,lachapelle2024nonparametric}, who have proposed a sparsity principle for identifiable CRL in interventional and temporal settings, motivated by the sparse mechanism shift hypothesis~\citep{scholkopf2021toward,perry2022causal}.

Our work is also related to \textit{sparse component analysis}~\citep{gribonval2006survey} and \textit{sparse dictionary learning}~\citep{Mairal_Bach_Ponce2009}. These unsupervised representation learning methods assume linear mixing functions $\fb$ and learn a sparse representations of the input $\Xb$, similar to our $\gb(\Xb)$. The identifiability of dictionary learning has been studied, e.g. by \citet{hu2023global}, in the finite sample regime. The main distinction with our work is that we focus on identifiability with nonlinear mixing.

Similar to our piecewise linear theorem, \citet{liu2022identifying} also assumes that the underlying SCM is linear Gaussian. As opposed to our work, it leverages the fact that the coefficients (or weights) of the causal relations are varying across environments, while in our case we leverage the partial observability patterns and assume that the underlying SCM is the same for all data. Similarly, \citet{liu2024identifiable} extends the results from \citet{liu2022identifying} to polynomial causal models with exponential family noise variables. These works are related to ours in leveraging changes across environments, but provide a different type of results.

\section{Conclusions and Limitations}
\label{sec: Discussion}

In this work, we focused on learning causal representations in the \emph{unpaired partial observability} setting, i.e., when only an instance-dependent subset of causal variables are captured in the measurements. We first proved the identifiability with linear mixing functions $\fb$ under a sparsity constraint. We then presented an example to illustrate why extending the results to nonlinear $\fb$ is not possible without additional assumptions. We proved identifiability when $\fb$ is piecewise linear, both the causal variables and the learned representations are Gaussian, and we know the group of each sample. 

While our experiments validate our theoretical results, there are still several limitations. From the theoretical point of view, knowing the group of each sample might not always be possible, so extending our results beyond this limitation is an exciting direction. Additionally, piecewise linear functions are a limited class and there might be other classes of nonlinear functions for which our identifiability results could be extended. In particular, our results hinge on the linearity of the composition $g$ and $f$ on $\mathcal{Z}$,
which can be implied by other types of constraints on $f$, $g$ and the causal variables.
Finally, our Gaussianity constraint is empirically difficult to satisfy, as shown by the gap between the performances of our method and the oracle. This warrants further investigation in other ways to encourage Gaussianity of the learned representations.

\clearpage

\section*{Acknowledgments}
This work was initiated at the Second Bellairs Workshop on Causality held at the Bellairs Research Institute, January 6--13, 2022; we thank all workshop participants for providing a stimulating research environment.
The research of DX and SM was supported by the Air Force Office of Scientific Research under award number FA8655-22-1-7155. Any opinions, findings, and conclusions or recommendations expressed in this material are those of the author(s) and do not necessarily reflect the views of the United States Air Force.
We also thank SURF for the support in using the Dutch National Supercomputer Snellius. 
DY was supported by an Amazon fellowship, the International Max Planck Research School for Intelligent Systems (IMPRS-IS), and the ISTA graduate school. Work was done outside of Amazon.
SL was supported by an IVADO
excellence PhD scholarship and by Samsung Electronics Co., Ldt.
JvK acknowledges support from the German Federal
Ministry of Education and Research (BMBF) through the Tübingen AI Center (FKZ: 01IS18039B).

\section*{Impact Statement}

This paper presents conceptual work whose goal is to advance the field of Machine Learning, and specifically Causal Representation Learning. There are many potential societal consequences of our work, none of which, we feel must be specifically highlighted here.

\bibliography{icml2024}
\bibliographystyle{plainnat}

\onecolumn
\newpage
\appendix
\newpage
\appendix

\section{Relation between partial observability and do interventions}\label{app:relation_partial_obs_interventions}

\begin{figure}[h]
    \centering
    \includegraphics[width=0.8\textwidth]{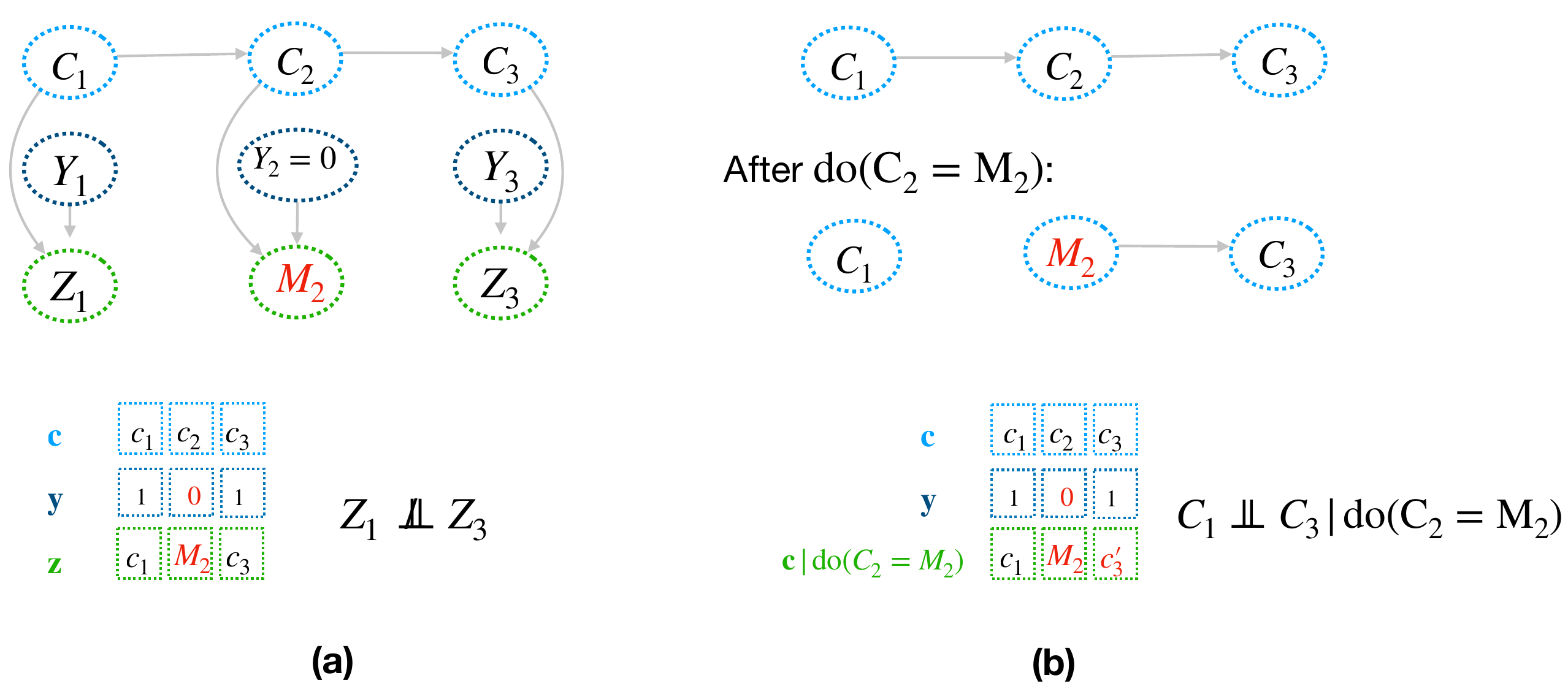}
    \caption{Comparison between (a) masking on $C_2$, and (b) do intervention on $C_2$. In the second case, there is an effect on $C_3$, and the intervention cuts the link and hence the dependence between $C_1$ and $C_3$.}
    \label{fig:int_vs_po}
\end{figure}

In our setting we consider two types of related variables: the causal variables $\Cb$ and the masked causal variables $\Zb$.
The \emph{causal variables} $\Cb=(C_1,...,C_n)$ represent the underlying causal system. 
The \emph{masked causal variables} $\Zb=(Z_1,...,Z_n) \in \Zcal$ 
are a combination of the causal variables and the masks $\Yb$, and they are the latent inputs to the mixing function $\fb$ that we are trying to recover. In particular, they are  the Hadamard product of the causal variables with the  binary mask variable, i.e., $\Zb=\Yb \odot \Cb$. For sample $i \in [ N ]$ and any causal variable $j \in [n]$, if the mask value $y^i_j$ is $1$, then the causal variable $c^i_j$ is measured and $z^i_j$ is $c^i_j$. Instead, if $y^i_j = 0$, then the causal variable is unmeasured and $z^i_j$ takes a fixed masked value $M_j$. We show an example in Fig.~\ref{fig:int_vs_po}a, where the binary mask $Y_2$ for causal variable $C_2$ is 0.

A superficially similar operation is a \emph{do-intervention} \citep{pearl2009causality}, in which we fix the value of a random variable to a specific value. For example, one could consider an intervention $\mathrm{do}(z^i_j = M_j)$, which forces the value of the variable $z^i_j$ to $M_j$. As shown in Fig.~\ref{fig:int_vs_po}, there are several differences between the masked variables and the intervened variables.

Despite having the same fixed value for the masked or intervened variable, these two operations have different effects. In particular, masking variables does not influence any downstream variables, as an intervention would. 
For example, consider the case of three variables $C_1 \rightarrow C_2 \rightarrow C_3$ in Fig.~\ref{fig:int_vs_po}a. In this case, if we mask $C_2$, then there will be no effect on the value of $C_3$, and $C_1$ and $C_3$ will still be dependent.
If we performed an intervention on $C_2$, as in Fig.~\ref{fig:int_vs_po}b, then there would be a change in the value of $C_3$, and $C_1$ and $C_3$ would be independent. %

\section{Proofs}
\label{app:proofs}

\subsection{Proof of the results for linear mixing functions (Theorem~\ref{thm: linear disentangle})}
\label{app:proofs:thm_linear}

We first introduce the definition of \emph{dependent inputs}, which intuitively is the set of variables on which a reconstruction of a given variable depends.

\begin{definition} \label{def:Ni_def} 
Let $\vb: \Zcal \rightarrow \mathbb{R}^n$ be a function with variables $\zb=(z_1,...,z_n) \in\Zcal$. 
For all $i \in [n]$ consider $N_i$ to be the set of all variables on which $\vb_i$ depends, which we will call \emph{dependent inputs}. Formally, we define the set of dependent inputs $N_i \subseteq [n]$ as
    \begin{align}
        N_i := \{j \in [n] \mid  \exists (z_j, \zb^0_{-j}), (z'_j, \zb^0_{-j})\in \Zcal \text{ where } z_j \neq z'_j \text{ s.t. } \vb_i(\zb^0_{-j}, z_j) \neq \vb_i(\zb^0_{-j}, z'_j)
        \}\,, 
    \end{align}
    where $\zb^0_{-j} $  is any $n-1$ dimensional vector that represents all of the components of a vector $\zb \in \Zcal$ except for the index $j$.
\end{definition}
This definition intuitively represents the indices of $\zb$ for which the function $\vb_i$ is not constant, or in other words, the inputs on which $\vb_i$ depends.
Note that if there is only one $z_j$ such that $(z_j, \zb^0_{-j}) \in \Zcal$ for an appropriate $\zb^0_{-j}$, then any function defined on $\Zcal$ is constant for that index, and hence $N_i$ does not contain $j$.

We now show a lemma that we will use to prove the theorem showing that for any diffeomorphism and any variable, there always exists a permutation that ensures that a variable is in the set of its dependent inputs. Intuitively, this ensures that for all variables, there always exists a permutation, such that the reconstruction of a given variable depends on the variable itself.

\begin{lemma}[Existence of permutation $\pi$ s.t. $i \in N_{\pi(i)}$] \label{lemma:Dv_perm}  
Let $\vb: \Zcal \rightarrow \mathbb{R}^n$
be a diffeomorphism onto its image with variables $\zb=(z_1,...,z_n) \in \Zcal$. 
Assume there exists a point $\zb^0 \in \Zcal$ and a value $\epsilon>0$, such that $\forall i \in [n]$, $\eb_i \odot [-\epsilon,\epsilon)^{n}+\zb^0 \subset \Zcal$, where $\eb_i$ is the standard basis for space $\RR^n$ for the $i$-th dimension, i.e. a $n$-dimensional vector in which all dimensions except $i$ are 0, and the $i$-th dimension is 1.
Then there exists a permutation $\pi: [n] \rightarrow [n]$ such that $i \in N_{\pi(i)}$ for all $i$, where $N_i$ is defined as in Def.~\ref{def:Ni_def}. 

\end{lemma}

\begin{proof}
Since $\vb$ is a diffeomorphism, its Jacobian $D\vb = \{\frac{\partial\vb _{i}}{\partial z_{j}}\}_{i, j \in [n]}$ is invertible everywhere, so it is invertible at $\zb^0 \in \Zcal$. Since $D\vb(\zb^0) $ is invertible, we have that its determinant is non-zero, i.e.
\begin{align}
    \det(D\vb(\zb^0)) := \sum_{\pi\in \mathfrak{S}_n} \text{sign}(\pi) \prod_{i=1}^n D\vb(\zb^0)_{\pi(i), i} \neq 0 \, , 
\end{align}
where $\mathfrak{S}_n$ is the set of $n$-permutations. This equation implies that at least one term of the sum is non-zero, and that for that term, all of the terms in the product are non-zero, meaning:
\begin{align}
    \exists \pi\in \mathfrak{S}_n, \forall i \in [n], D\vb(\zb^0)_{\pi(i), i} \neq 0 \; .
\end{align}
This means that, for all $i \in [n]$, $\frac{\partial\vb _{\pi(i)}}{\partial z_{i}}(\zb^0) \not = 0$, which implies that $\vb_{\pi(i)}$ is not constant for $z_{i}$ in $\zb^0$. Then by definition of $N_i$ in Def.~\ref{def:Ni_def}, $i \in N_{\pi(i)}$.

\end{proof}

As an intermediate step, we first prove an important lemma that shows that by enforcing sparsity of the transformed variables,  
the corresponding transformation is an element-wise linear function.
Intuitively, these transformed variables will be the reconstructed  masked latent variables $\Zb$.

\begin{lemma}[Element-wise Identifiability for Linear Transformation] Assume that the masked latent variables $\Zb$ with support $\Zcal$ follow the data generating process in Sec.~\ref{sec: Problem set up} and Ass.~\ref{assump: sufficient support} holds. Let the function $\vb: \Zcal \rightarrow \RR^n$ be invertible and linear on $\Zcal$
, and
\begin{align}
\mathbb{E}\norm{\vb(\Zb)}_0 &\leq \mathbb{E}\norm{\Zb}_0.
\label{equ: sparsity_constraint0}
\end{align}
Then $\vb$ is a permutation composed with an element-wise invertible linear transformation on $\Zcal$.
\label{lemma: element_wise v}
\end{lemma}
\begin{proof}

    We reuse the definition of the support indices $\Sb:=\{i \in [n]: Z_i \neq 0\}$ and analyze each side of inequality~\eqref{equ: sparsity_constraint0}, starting with its right-hand side.
    \begin{align}
        \mathbb{E}||\Zb||_0 &= \mathbb{E}\sum^n_{i=1} \mathbbm{1}(\Zb_i \not= 0) \\
        &= \sum_{\sb \in \mathcal{S}} p(\sb) \mathbb{E}\left[\sum^n_{i=1} \mathbbm{1}(\Zb_i \not= 0) \mid \Sb = \sb\right] \\
        &= \sum_{\sb \in \mathcal{S}} p(\sb) \sum^n_{i=1}\mathbb{E}[\mathbbm{1}(\Zb_i \not= 0) \mid \Sb = \sb]\\
        &= \sum_{\sb \in \mathcal{S}} p(\sb) \sum^n_{i=1}\mathbbm{1}( i\in \sb)\,
        \label{eq:rhs_ineq}
    \end{align}
    Now analyzing the left hand side of~\eqref{equ: sparsity_constraint0}, starting with similar steps as previously we get
    \begin{align}
        \mathbb{E}||\vb(\Zb)||_0 &= \sum_{\sb \in \mathcal{S}} p(\sb) \sum^n_{i=1}\mathbb{E}[\mathbbm{1}(\vb_i(\Zb) \not= 0) \mid \Sb = \sb]\\
        &= \sum_{\sb \in \mathcal{S}} p(\sb) \sum^n_{i=1}\mathbb{P}_{\Zb\mid \Sb=\sb}[\vb_i(\Zb) \not= 0] \label{eq:left_ineq}\\
        &= \sum_{\sb \in \mathcal{S}} p(\sb) \sum^n_{i=1}\left(1 - \mathbb{P}_{\Zb\mid \Sb=\sb}[\vb_i(\Zb) = 0]\right) \,. \label{eq:lhs_ineq}
    \end{align}

    For 
    $\vb$ to be a permutation composed with an element-wise invertible linear transformation on $\Zcal$ %
    , it is enough to show there exists a permutation $\pi: [n] \rightarrow [n]$ such that, for every $i$, $N_i = \{\pi(i)\}$. To achieve this, we are going to first show that 
\begin{align}\label{eq:equality_seb}
    \mathbb{P}_{\Zb\mid \Sb=\sb}[\vb_i(\Zb) = 0] = \alpha_i\mathbbm{1}(N_i \cap \sb = \emptyset) \,,
\end{align}
where $\alpha_i \in \{0,1\}$. Since $\vb_i$ is linear on $\Zcal$, 
 we have that $\vb_i(\Zb) = \wb^i \cdot \Zb+\cb_i$ for some $\wb^i \in \mathbb{R}^{n}$. Furthermore, we show that in this case $N_i = \{j \in [n] \mid \wb^i_j \not=0\}$. In one direction, if $\wb^i_j = 0$, then $\vb_i$ is  constant in dimension $j$, so $\wb^i_j$ should be non-zero for $j$ to be included in $N_i$. For the other direction, if $\wb^i_j \neq 0$ there are two cases in which we could have that $j \not \in N_i $: (i) if $\wb^i_j \Zb_j$ always cancels out with another $\wb^i_k \Zb_k$, where both $\wb^i_j, \wb^i_k \neq 0$, (ii) if the variable $\Zb_j$ can only take a single value.
 We show that neither of these cases can happen because of Ass.~\ref{assump: sufficient support}. In particular, the first case cannot happen because two masks for two different variables cannot be the same and satisfy this assumption. The second case also cannot happen because of Ass.~\ref{assump: sufficient support}, since there needs to be a support index set in which $i$ is masked and one in which it is not masked.
 Thus,
 $$\vb_i(\Zb) = \wb^i \cdot \Zb +\cb_i = \wb^i_{N_i} \cdot \Zb_{N_i}+\cb_i\, .$$

 \textbf{Case 1:} Suppose $N_i \cap \sb = \emptyset$. Then, 
 $$\mathbb{P}_{\Zb\mid \Sb=\sb}[\vb_i(\Zb) = 0] = \mathbb{P}_{\Zb\mid \Sb=\sb}[\wb^i_{N_i} \cdot \Zb_{N_i}+\cb_i = 0] = \mathbb{P}_{\Zb\mid \Sb=\sb}[\wb^i_{N_i} \cdot \bm 0 +\cb_i = 0] = \mathbb{P}_{\Zb\mid \Sb=\sb}[\cb_i = 0] = \alpha_i\,.$$
Note that the event $\mathbb{P}_{\Zb\mid \Sb=\sb}[\cb_i = 0]$ is deterministically either true of false, hence
$\alpha_i \in \{0,1\}$.

 \textbf{Case 2:} Suppose $N_i \cap \sb \not= \emptyset$. Thus, $\wb^i_{\sb} \not= \bm 0$. Thus,
 $$\mathbb{P}_{\Zb\mid \Sb=\sb}[\vb_i(\Zb) = 0] = \mathbb{P}_{\Zb\mid \Sb=\sb}[\wb^i \cdot \Zb +\cb_i = 0] = \mathbb{P}_{\Zb\mid \Sb=\sb}[\wb^i_\sb \cdot \Zb_\sb +\cb_i = 0] \,.$$

 Note that the event $\{\Zb_\sb \mid \wb^i_\sb \cdot \Zb_\sb +\cb_i = 0\}$ corresponds to the kernel of the linear map $\wb^i_\sb$. We can thus infer its dimensionality via the rank-nullity theorem~\citep{friedberg2014linear} which states that 
 $\text{rank}(\wb_{\sb}^i) + \text{dim}(\text{Ker}(\wb_{\sb}^i)) = \text{dim}(\text{Dom}(\wb_{\sb}^i))$, where Ker() is nullity and Dom() is domain, which here implies that $1 + \text{dim}(\text{Ker}(\wb_{\sb}^i)) = |\sb|$. We thus have $\text{dim}(\{\Zb_\sb \mid \wb^i_\sb \cdot \Zb_\sb +\cb_i= 0\}) = |\sb| - 1$. Since $\mathbb{P}_{\Zb\mid \Sb=\sb}$ has a density w.r.t. to the Lebesgue measure, we have that $\mathbb{P}_{\Zb\mid \Sb=\sb}[\wb^i_\sb \cdot \Zb_\sb +\cb_i = 0] = 0$ (since a density w.r.t. to Lebesgue cannot concentrate mass on a lower-dimensional linear subspace).

 We thus have proved that indeed, $\mathbb{P}_{\Zb\mid \Sb=\sb}[\vb_i(\Zb) = 0] = \alpha_i\mathbbm{1}(N_i \cap \sb = \emptyset)$.

Putting \eqref{equ: sparsity_constraint0}, \eqref{eq:rhs_ineq}, \eqref{eq:lhs_ineq} and \eqref{eq:equality_seb} together, we obtain
    \begin{align}
        \sum_{\sb \in \mathcal{S}} p(\sb) \sum^n_{i=1} [(1 -\alpha_i)+  \alpha_i\mathbbm{1}(N_i \cap \sb \neq \emptyset)] \leq \sum_{\sb \in \mathcal{S}} p(\sb) \sum^n_{i=1}\mathbbm{1}( i\in \sb)
    \end{align}
We can now use Lemma~\ref{lemma:Dv_perm}, because Ass.~\ref{assump: sufficient support} implies the existence of the $\zb_0$ point, which in this case is $(0,...,0)$.

By using this lemma, we can show that there exists a permutation $\pi$ such that, for all  $i \in [n]$, $i \in N_{\pi(i)}$.
 We now permute the terms on the l.h.s. according to $\pi$ and reorganize the terms as:
    \begin{align}
        &\sum_{\sb \in \mathcal{S}} p(\sb) \sum^n_{i=1} [(1 -\alpha_{\pi(i)})+  \alpha_{\pi(i)} \mathbbm{1}(N_{\pi(i)} \cap \sb \neq \emptyset)] \leq \sum_{\sb \in \mathcal{S}} p(\sb) \sum^n_{i=1}\mathbbm{1}( i\in 
        \sb) \nonumber \\
        &\sum_{\sb \in \mathcal{S}} p(\sb) \sum^n_{i=1} [(1 -\alpha_{\pi(i)})+  \alpha_{\pi(i)}\mathbbm{1}(N_{\pi(i)} \cap \sb \neq \emptyset) - \mathbbm{1}( i\in \sb)] \leq 0
        \label{eq:leq_zero}
    \end{align}
    Note how, for all $i$, $(1 -\alpha_{\pi(i)})+  \alpha_{\pi(i)}\mathbbm{1}(N_{\pi(i)} \cap \sb \neq \emptyset) - \mathbbm{1}( i\in \sb) \geq 0$, since whenever $i \in \sb$, we must have $N_{\pi(i)} \cap \sb \not= \emptyset$, because we chose a permutation such that $i \in N_{\pi(i)}$. Note also
    that this is true irrespective of the value of $\alpha_{\pi(i)}\in\{0,1\}$. 
    
    We then can notice that if $i \not \in \sb$, then $\alpha_{\pi(i)}=0$. The function can have either value $0$ or $1$, but in any case not negative. Hence the inequality in \eqref{eq:leq_zero} is actually an equality and hence for all $\sb \in \mathcal{S}$ and all $i \in [n]$,
    \begin{align}
        (1 -\alpha_{\pi(i)})+  \alpha_{\pi(i)}\mathbbm{1}(N_{\pi(i)} \cap \sb \neq \emptyset) - \mathbbm{1}( i\in \sb)=0
        \,. 
        \label{equ: mid}
    \end{align}
The first thing we conclude is that if $i \not\in \sb$,  then $\alpha_{\pi(i)}=1$, since otherwise
Equ.\eqref{equ: mid} is violated. Under Ass.~\ref{assump: sufficient support}, we have that, for all $i \in [n]$, there
exists an $\sb \in \Scal $ such that $i \not\in \sb$. We thus conclude that $\alpha_i=1$ for all $i \in [n]$, which allows us to write
\begin{align}
        \mathbbm{1}(N_{\pi(i)} \cap \sb \not= \emptyset) &= \mathbbm{1}( i\in \sb) \\
        \mathbbm{1}(N_{\pi(i)} \cap \sb = \emptyset) &= \mathbbm{1}( i\not\in \sb)\,. \label{eq:cool_eq}
    \end{align}

    Importantly, this means
    \begin{align}
        \forall i \in [n], \forall \sb\in \mathcal{S},\ i\not\in \sb &\implies N_{\pi(i)} \cap \sb = \emptyset 
        \implies N_{\pi(i)} \subseteq \sb^c\,,
    \end{align}
    which can be rewritten as 
    \begin{align}
        \forall i \in [n], N_{\pi(i)} \subseteq \bigcap_{\sb \in \mathcal{S} \mid i \not\in \sb} \sb^c\,. \label{eq:final_eq}
    \end{align}
    We now rewrite Assumption~\ref{assump: sufficient support} below 
    and take the complement on both sides:
  \begin{align}
        \forall i\in [n], \bigcup_{\sb \in \mathcal{S} \mid i \not\in \sb} \sb &= [n] \setminus \{i\}\\
        \bigcap_{\sb \in \mathcal{S} \mid i \not\in \sb} \sb^c &= \{i\} \label{eq:final_eq2}
    \end{align} 
    Combining \eqref{eq:final_eq} with \eqref{eq:final_eq2} implies that $N_{\pi(i)} = \{i\}$ for all $i$, which concludes the proof.
\end{proof}

We now prove identifiability up to permutation and element-wise linear transformations for the case of a linear mixing function, given the assumption of sufficient support index variability.

\thmlinear*
\begin{proof}
    Since $\Xb = \fb(\Zb)$, we can rewrite Equation \eqref{eq:zero_reconstruction} (perfect reconstruction) as
    \begin{align}
        \mathbb{E}||\fb(\Zb) - \hat\fb(\gb(\fb(\Zb)))||^2_2 = 0\,.
    \end{align}
This means $\fb$ and $\hat\fb\circ\gb\circ\fb$ are equal $\mathbb{P}_\Zb$-almost everywhere.
Both of these functions are continuous, $\fb$ by assumption and  $\hat\fb\circ\gb\circ\fb$ because $\hat\fb$ is continuous, and $\fb, \gb$ are linear.
Since they are continuous and equal $\mathbb{P}_\Zb$-almost everywhere, this means that they must be equal over the support of $\Zb$, $\mathcal{Z}$,
i.e.,
    \begin{align}
        \fb(\zb) = \hat \fb \circ \gb \circ \fb (\zb)\,, \forall \zb \in \mathcal{Z}\,.
    \end{align}
This can be easily shown by considering any point $\zb' \in \mathcal{Z}$ on which $\fb$ and $\hat\fb\circ\gb\circ\fb$ are different, i.e. $\hat\fb\circ\gb\circ\fb(\zb') \neq \fb(\zb')$, This would imply that $(\fb-\hat\fb\circ\gb\circ\fb)$, which is also a continuous function, is non-zero in $\zb'$, and in its neighbourhood. This would contradict the assumption that $\fb$ and $\hat\fb\circ\gb\circ\fb$ are the same almost everywhere. We can now apply the inverse%
on both sides to obtain
    \begin{align}
        \hat\fb^{-1} \circ \fb(\zb) = \underbrace{\gb \circ \fb}_{\vb :=} (\zb)\,, \forall \zb \in \mathcal{Z}\,.
    \label{equ: piecewise repeat}
    \end{align}
    Since both $\fb$ is an injective linear function from $\Zcal$ to $\Xcal$ and $\gb$ is an invertible linear function from $\Xcal$ onto its image, then, $\hat\fb^{-1} \circ \fb$ is an invertible linear function on $\Zcal$. As $\vb$ is equal to $\hat\fb^{-1} \circ \fb$ on $\Zcal$, then, we have $\vb$ is also an invertible linear function on $\Zcal$. %
    By Lemma~\ref{lemma: element_wise v}, we can derive $\vb$ is a permutation composed with an element-wise linear transformation on $\Zcal$. %
\end{proof}

\subsection{Proof of the results for the piecewise linear functions( Theorem~\ref{thm: disentanglement piecewise})} 
\label{app:proof element-wise piecewise linear}

For piecewise linear mixing functions, we first prove an intermediate result (Thm~\ref{adap_thm:identif-inv-affine}) for a weaker form of identifiability that does not imply a disentanglement, but an affine correspondence between the ground truth and the learned variables.

\begin{definition}[Identifiability up to affine transformation \citep{khemakhem2020variational,lachapelle2022disentanglement}]
 \label{def: linear ident}
The ground truth representation vector $\Zb$ ($n$-dimensional random vector) is  \emph{identified up to affine transformations} by a learned representation vector $\hat{\Zb}$ (also $n$-dimensional random vector) when there exists an invertible \emph{linear transformation} $\hb$ such that, $\hat{\Zb}=\hb(\Zb)$ almost surely.%
\end{definition}

We then introduce Lemma~\ref{lemma: affine given mask}, which proves that, when $\Yb$ is provided, we can identify the distribution of latent variables by linear transformations that may vary across $\Yb$. It is crucial to emphasize that in this work, having $\mathbf{Y} = \mathbf{y}$ does not imply knowledge of the exact value of $\mathbf{y}$; instead, we simply need the information on \emph{grouping}, i.e. the partitioning of the dataset based on mask values. Finally, we use these intermediate results to conclude the prove of the  theorem.

\subsubsection{Linear Identifiability for (De)-MVNs with Piecewise Affine $\fb$}

In what follows, we let $\equiv$  denote equality in distribution.

\begin{restatable}[Linear Identifiability for (De)-MVNs with Piecewise Affine $\fb$]{theorem}{thmpcaffine}
\label{adap_thm:identif-inv-affine}
Assume $\fb, \hat{\fb}:\mathbb{R}^n \rightarrow \mathbb{R}^d$ are injective and piecewise affine. We assume $\Zb$ and $\hat{\Zb}$ follow a (degenerate) multivariate normal distribution.
If $\fb(\Zb) \equiv \hat{\fb}(\hat{\Zb})$, then there exists an invertible affine transformation $\hb:\RR^n \rightarrow \mathbb{R}^n$ such that $\hb(\Zb) \equiv \hat{\Zb}$
(Def.~\ref{def: linear ident}).%
\end{restatable}

Intuitively, Thm~\ref{adap_thm:identif-inv-affine} states that if we can align two (De)-
MVN-distributed random vectors to the same distribution through piecewise affine transformations, then their distributions can be interchanged via an affine transformation. This finding is initially inspired by \citet{kivva2022identifiability}, who established the linear identifiability for latent variables following \emph{non-denegerate} Multivariate Normal (MVN) distributions. Our primary focus revolves around addressing two key issues to extend the findings from MVN to de-MVN: i) determining the identifiability of de-MVN distributions, which do not have a probability density function, and ii) constructing an affine function when the domain changes from $\RR^n$ to $\Zcal$.

\label{app:proofs:thm_pc_affine}

We start by proving an useful lemma about (degenerate) multivariate normal distributions.
\begin{lemma}
\label{lemma: close affine}
(Degenerate) Multivariate Normals, or (De)MVNs,  are close under affine transformation. More formally, if $\Zb \sim N(\mub,\Sigmab)$ with $\mu \in \mathbb{R}$ and $|\Sigmab| \geq 0$, is a potentially degenerate multivariate normal variable, then $\Ab\Zb$, where $\Ab\in \mathbb{R}^{n\times n}$ is also a potentially degenerate multivariate normal variable.
\end{lemma}
\begin{proof}
Let $\hat{\Zb}=\Ab\Zb$, then
$P_{\hat{\Zb}}=\Ab N(\mub,\Sigmab)=N(\Ab\mu,\Ab \Sigmab \Ab^T)$
where the determinant of the covariance, $|\Ab \Sigmab \Ab^T| \geq 0$.
Therefore, $\hat{\Zb}$ is a potentially degenerate multivariate normal variable.
\end{proof}

We now summarize the results on the identifiability of non-degenerate multivariate normal variables by \citet{kivva2022identifiability}. We report an adapted version of Theorem C.3 by \citet{kivva2022identifiability}.
\begin{theorem}[Identifiability of non-degenerate MVNs \citep{kivva2022identifiability}]
\label{thm:local-gmm-iden}
Consider a pair of non-degenerate MVNs in $\mathbb{R}^n$. If
\begin{equation}
    P=N(\mub, \Sigmab)\quad \text{and}\quad P'= N(\mub', \Sigmab'),
\end{equation}
and there exists a ball $B(\xb_0, \delta)$, where $\xb_0\in \mathbb{R}^n$ and $\delta>0$, such that $P$ and $P'$ induce the same measure on $B(\xb_0, \delta)$, then $P \equiv P'$.
\end{theorem}

The original proof follows from the identity theorem for real analytic functions. 
We extend this result to the case of potentially degenerate multivariate normal variables, which we call \emph{(De)-MVNs}. We first propose an intermediate result for the case in which only one of the variables is a (degenerate) multivariate normal, while the other variable is a non-degenerate multivariate normal. We then use this result to prove the general case in which both variables are potentially degenerate MVNs.

\begin{lemma}[Identifiability of a (De)-MVNs and a non-degenerate MVN]\label{adap_thm:local-full-de-iden}
Consider a pair of random vectors $\Xb$, $\Xb'$ in $\mathbb{R}^n$ distributed as
\begin{equation}
    \Xb \sim N(\mub, \Sigmab)\quad \text{and}\quad \Xb'\sim N(\mub', \Sigmab'),
\end{equation}
for appropriate values of $\mub, \mub'$ and where the determinant $|\Sigmab|\geq 0$ and the determinant $|\Sigmab'|>0$. In other words, $\Xb$ is a potentially degenerate MVN, while $\Xb'$ is a non-degenerate MVN. 

If there exists a ball $B(\xb_0, \delta) \subseteq \mathbb{R}^n$, where $\xb_0\in \mathbb{R}^n$ and $\delta>0$, such that $\Xb$ and $\Xb'$ follow the same distribution on $B(\xb_0, \delta)$, then $\Xb \equiv \Xb'$, i.e.,$(\mub, \Sigmab) = (\mub', \Sigmab')$.
\end{lemma}
\begin{proof}
Let the rank of $\Sigmab$ be $k \leq n$ and consider the spectral decomposition of $\Sigmab$:
\begin{equation}
    \Sigmab=QDQ^T,
\end{equation}
where $Q$ is an orthogonal $n\times n$ matrix and $D$ a the diagonal matrix. If $n=k$ we consider D to have $k$ diagonal entries $(\sigma_1^2, \sigma_2^2, \dots, \sigma_k^2)$ where $\sigma_i$ for $i\in [k]$ are the eigenvalues. Otherwise, if $k <n$, then $D$ has $n$ diagonal entries $(\sigma_1^2, \sigma_2^2, \dots, \sigma_k^2, 0, \dots, 0)$ where $\sigma_i$ for $i\in [k]$ are the eigenvalues.

Let $\Yb=Q^T \Xb$ and $\Yb'=Q^T \Xb'$. Since $Q$ is an orthogonal matrix, this means that 
\begin{align}
    \Yb &\sim N (Q^T \mub, Q^T QD Q^T Q) = N(Q^T \mu, D)\\
    \Yb' &\sim N (Q^T \mub', Q^T \Sigma' Q)
\end{align}

Since we know $\Xb \equiv \Xb'$ in $B(\xb_0, \delta)$, then we can derive that $\Yb \equiv \Yb'$ in $B(Q^T \xb_0, \tilde{\delta})$ for an appropriate $\tilde{\delta} > 0$. We project $B(Q^T \xb_0, \tilde{\delta})$ into two subspaces, $B(Q^T \xb_0, \tilde{\delta})_{1:k}$ and $B(Q^T \xb_0, \tilde{\delta})_{k+1:n}$. The first captures the first $k$ dimensions of the ball, and the second the last $(n-k)$ dimensions.

We can pick the first $k$ dimensions of $\Yb$ and $\Yb'$, and denote them as $\Yb_{1:k}$ and $\Yb'_{1:k}$ respectively. 
The first $k$ dimensions of both variables are still the same, i.e., $\Yb_{1:k}\equiv \Yb'_{1:k}$ in $B(Q^T \xb_0, \tilde{\delta})_{1:k}$.
We can show that $\Yb_{1:k}$ is a non-degenerate multivariate normal, because its covariance matrix $D_{1:k,1:k}$ is full rank. Since both $\Yb_{1:k}$ and $\Yb'_{1:k}$ are non-degenerate multivariate normals, by Theorem~\ref{thm:local-gmm-iden} by \citep{kivva2022identifiability} we have $\Yb_{1:k}\equiv \Yb'_{1:k}$.

We will now prove by contradiction that $\Yb$ is also a non-degenerate MVN, i.e., that $k=n$. 
We consider the other $(n-k)$ dimensions of $\Yb$ and $\Yb'$.
The covariance matrix of $\Yb_{k+1:n}$ is $D_{k+1:n,k+1:n}$, which is a zero matrix.
However, since determinant $|\Sigmab'|>0$, the variance of any component of $\Yb'_{k+1:n}$ cannot be $0$. Since $\Yb_{k+1:n} \equiv \Yb'_{k+1:n}$ in the ball $B(Q^T \xb_0, \tilde{\delta})_{k+1:n}$, their covariance matrices should be the same. We now come to a contradiction, because one is supposed to be a zero matrix, while the other one is supposed to be full rank. 
We therefore derive that $k=n$, and hence $\Yb_{1:k}\equiv \Yb'_{1:k}$ implies $\Yb \equiv \Yb'$.
We can now exploit that $\Xb=Q\Yb$ and $\Xb'=Q\Yb'$, due to the orthogonality of $Q$, and conclude that $\Xb \equiv \Xb'$.
\end{proof}

\begin{lemma}[Identifiability of (De)-MVNs]\label{adap_thm:local-gmm-iden}
Consider a pair of random vectors $\Xb$, $\Xb'$ in $\mathbb{R}^n$ distributed as
\begin{equation}
    \Xb \sim N(\mub, \Sigmab)\quad \text{and}\quad \Xb'\sim N(\mub', \Sigmab'),
\end{equation}
for appropriate values of $\mub, \Sigmab, \mub', \Sigmab'$, including also singular $\Sigmab$ and $\Sigmab'$. If there exists a ball $B(\xb_0, \delta) \subseteq \mathbb{R}^n$, where $\xb_0\in \mathcal{X}$, $\delta>0$ and $\mathcal{X}$ is support of $\Xb$, such that $\Xb$ and $\Xb'$ follow the same distribution on $B(\xb_0, \delta)$, then $\Xb \equiv \Xb'$, i.e.,$(\mub, \Sigmab) = (\mub', \Sigmab')$.
\end{lemma}

\begin{proof}
Let the rank of $\Sigmab$ be $k \leq n$ and consider the spectral decomposition of $\Sigmab$:
\begin{equation}
    \Sigmab=QDQ^T,
\end{equation}
where $Q$ is an orthogonal $n\times n$ matrix and D a the diagonal matrix. If the rank $k < n$, i.e. $\Xb$ is a degenerate multivariate normal, we consider D to have $n$ diagonal entries $\sigma_1^2$, $\sigma_2^2$, ..., $\sigma_k^2$, 0, ..., 0, where $\sigma_i$ for $i\in [k]$ are the eigenvalues.

Let $\Yb=Q^T \Xb$ and $\Yb'=Q^T \Xb'$. This means that 
\begin{align}
    \Yb &\sim N (Q^T \mub, Q^T QD Q^T Q) = N(Q^T \mub, D)\label{eq: dist_Y}\\
    \Yb' &\sim N (Q^T \mub', Q^T \Sigma' Q)
\end{align}
Since we know $\Xb \equiv \Xb'$ in $B(\xb_0, \delta)$, then we can derive that $\Yb \equiv \Yb'$ in $B(Q^T \xb_0, \tilde{\delta})$ for an appropriate $\tilde{\delta} > 0$.  We project $B(Q^T \xb_0, \tilde{\delta})$ into two subspaces, $B(Q^T \xb_0, \tilde{\delta})_{1:k}$ and $B(Q^T \xb_0, \tilde{\delta})_{k+1:n}$. The first captures the first $k$ dimensions of the ball, and the second the last $(n-k)$ dimensions.

We can pick the first $k$ dimensions of $\Yb$ and $\Yb'$, and denote them as $\Yb_{1:k}$ and $\Yb'_{1:k}$ respectively. 
The first $k$ dimensions of both variables are still the same, i.e., $\Yb_{1:k}\equiv \Yb'_{1:k}$ in $B(Q^T \xb_0, \tilde{\delta})_{1:k}$.
We can show that $\Yb_{1:k}$ is a non-degenerate multivariate normal, because its covariance matrix $D_{1:k,1:k}$ is full rank. So by Lemma~\ref{adap_thm:local-full-de-iden} we have $\Yb_{1:k}\equiv \Yb'_{1:k}$, i.e. $(Q^T \Sigma' Q)_{1:k,1:k}=D_{1:k,1:k}$.

For the other $(n-k)$ dimensions of $\Yb$ and $\Yb'$, i.e., $\Yb_{k+1:n}$ and $\Yb'_{k+1:n}$, we can also show that $\Yb_{k+1:n}\equiv \Yb'_{k+1:n}$ in $B(Q^T \xb_0, \tilde{\delta})_{k+1:n}$.
For $\Yb_{k+1:n}$, since $\xb_0$ is contained in $B(\xb_0, \delta)$, we can derive that $Q^T \xb_0$ is contained in $B(Q^T \xb_0, \tilde{\delta})$. Since the covariance matrix of $\Yb_{k+1:n}$ is $D_{k+1:n, k+1:n}$, which is a zero matrix, the distribution of $\Yb_{k+1:n}$ is a point mass with all of the probability on a single value $(Q^T \xb_0)_{k+1:n}$. From \eqref{eq: dist_Y}, we know that $\Yb_{k+1:n} \sim N((Q^T \mub)_{k+1:n}, D_{k+1:n, k+1:n})=N((Q^T \mub)_{k+1:n}, \mathbf{0})$, so we can derive $(Q^T \mub)_{k+1:n}=(Q^T \xb_0)_{k+1:n}$.

Since $\Yb_{k+1:n}\equiv \Yb'_{k+1:n}$ in $B(Q^T \xb_0, \tilde{\delta})_{k+1:n}$, and $\Yb_{k+1:n}$ is a point mass on $(Q^T \mub)_{k+1:n} \in B(Q^T \xb_0, \tilde{\delta})_{k+1:n}$, then we can derive that $\Yb'_{k+1:n}$ should be a point mass on the same point $(Q^T \mub)_{k+1:n}$. Therefore, $(Q^T \Sigma' Q)_{k+1:n,k+1:n}$is a zero matrix, which is equal to $D_{k+1:n,k+1:n}$,  and $(Q^T \mub)_{k+1:n}=(Q^T \mub')_{k+1:n}$. This means that $\Yb_{k+1:n}\equiv \Yb'_{k+1:n}$.

We can now exploit that $\Xb=Q\Yb$ and $\Xb'=Q\Yb'$ due to the orthogonality of $Q$, and we can write:
\begin{align}
    \Xb=Q\Yb=Q ({\Yb_{1:k}}^T,{\Yb_{k+1:n}}^T)\\
    \Xb'=Q\Yb'=Q ({\Yb'_{1:k}}^T,{\Yb'_{k+1:n}}^T),
\end{align}
which together with $\Yb_{1:k}\equiv \Yb'_{1:k}$ and $\Yb_{k+1:n}\equiv \Yb'_{k+1:n}$ implies that $\Xb\equiv \Xb'$.
\end{proof}

\thmpcaffine*

\begin{proof} 
By the definition of (De) MVN, we can write $\Zb = \Ab \boldsymbol \varepsilon + \bb$, where $\boldsymbol \varepsilon \sim N(\mathbf{0},\Ib)$ with dimension $d_{\boldsymbol \varepsilon} \leq n$ and support $\RR^{d_{\boldsymbol \varepsilon}}$,  and $\Ab \in \RR^{n \times d_{\boldsymbol \varepsilon}}$ has full column rank. 
The support of $\Zb$, $\Zcal\subseteq \RR^n$, is  $\Zcal= colsp(\Ab)+\bb$, where $colsp(\Ab)$ is the column space of matrix $\Ab$.

Since $f (\Zb)$ and $\hat{f}(\hat{\Zb})$ are equally distributed, they have the same image space. Therefore, we can define $\hb_0 := \hat{\fb}^{-1}\circ \fb$ on $\Zcal$, %
which is a piecewise affine function.
Since we assumed that $\fb(\Zb)$ is equally distributed to $\hat{\fb}(\hat{\Zb})$, then $\hat{\Zb} \equiv \hat{\fb}^{-1}\circ \fb(\Zb) = \hb_0(\Zb)$.
We can now define another function that takes as domain the whole space $\RR^n$:
\begin{align}
 \widetilde{\hb}(\xb):=\hb_0(\Ab(\Ab^T \Ab)^{-1}\Ab^T(\xb-\bb)+\bb),   
\end{align} where $\xb \in \RR^n$. This function is the same as $\hb_0$ on $\Zcal$ 
We first proof that $\widetilde{\hb}$ and $ \hb_0$ agree on $\Zcal$. In particular, $\forall \zb_0 \in \Zcal$, $\exists \boldsymbol \varepsilon_0 \in \RR^{d_{\boldsymbol \varepsilon}}$, s.t. $\zb_0=\Ab\boldsymbol \varepsilon_0+\bb$. Thus, by definition of $\widetilde{\hb}$, we have $\forall \zb_0 \in \Zcal$, \begin{align}
     \widetilde{\hb}(\zb_0) &= 
     \hb_0(\Ab (\Ab^T \Ab)^{-1}\Ab^T(\zb_0-\bb) + \bb)\\
     &= 
     \hb_0(\Ab (\Ab^T \Ab)^{-1}\Ab^T(\Ab\boldsymbol \varepsilon_0+\bb-\bb) + \bb)\\
     &=
     \hb_0(\Ab\underbrace{(\Ab^T \Ab)^{-1}\Ab^T\Ab}_{=\Ib}\boldsymbol \varepsilon_0 +\bb)\\
     &= \hb_0(\Ab\boldsymbol \varepsilon_0+\bb)
     \label{equ: compo h_tilde}\\
     &= \hb_0(\zb_0).
\end{align}

So we have $\hb_0$ and $\widetilde{\hb}$ agree on $\Zcal$. This implies that $\hat{\Zb} \equiv \widetilde{\hb}(\Zb)$.

Then, we will complete the proof by two steps:
\begin{itemize}
    \item[i] There exists a $\zb_0 \in \Zcal$ and $\delta >0$ s.t. $\widetilde{\hb}$ is affine on the ball $\Bb(\zb_0, \delta) \subset \RR^n$ (potentially not completely contained in $\Zcal)$. %
    
    \item[ii] We can define an affine function $\hb$ on $\Zcal$ such that $\widetilde{\hb}$ and $\hb$ agree on the ball $B$. Then we show $\hb(\Zb)\equiv \hat{\Zb}$.
\end{itemize}

\emph{(i)}. We now first show that $\exists \zb_0 \in \Zcal$, s.t. $\widetilde{\hb}$ is differentiable at point $\zb_0$.

As shown in Eq.~\ref{equ: compo h_tilde}, $\widetilde{\hb}(\zb_0)$ can be written as the composition of the functions $\hb_0(\Ab\boldsymbol \varepsilon_0 +\bb)$ and $(\Ab^T \Ab)^{-1}\Ab^T(\zb_0-\bb)$. The function $\hb_0(\Ab\boldsymbol \varepsilon_0 +\bb )$ is piecewise affine, because it is a  composition of (piecewise) affine maps. Hence, there must exists a point $\boldsymbol\epsilon_0 \in \RR^{d_\epsilon}$ s.t. $\hb_0(\Ab\boldsymbol \varepsilon +\bb )$ is differentiable at that point. Moreover, we have that $(\Ab^T \Ab)^{-1}\Ab^T(\xb-\bb)$ is differentiable at $\zb_0 := \Ab\boldsymbol \varepsilon_0 + \bb$, because it is an affine function, and $(\Ab^T \Ab)^{-1}\Ab^T(\xb-\bb)$ evaluated at $\zb_0$ yields $\boldsymbol \varepsilon_0$, so the composition $\widetilde{\hb}$ is differentiable at $\zb_0$. %
Since $\widetilde{\hb}$ is piecewise affine and differentiable in $\zb_0$, then we can construct a ball $\Bb(\zb_0, \delta)$ (not necessarily completely contained in $\Zcal$) which contains one single linear piece of $\widetilde{\hb}$.

\emph{(ii)} Let $\hb:\RR^n \rightarrow \RR^n$ %
be an invertible affine function such that $\hb$ coincides with $\widetilde{\hb}$ on $\Bb(\zb_0,\delta)$. This means $\hb(\Zb)$ and $\widetilde{\hb}(\Zb)$ coincide on $\Bb(\zb_0,\delta) \cap \Zcal$. Since we have shown that $\widetilde{\hb}(\Zb) \equiv \hat{\Zb}$ are equal in distribution on $\Zcal$, then $\hb(\Zb) = \widetilde{\hb}(\Zb) \equiv \hat{\Zb}$ on $\Bb(\zb_0,\delta)\cap \Zcal$.

Since we assume $\Zb$ is a (De-)MVN, by Lemma~\ref{lemma: close affine}, $\hb(\Zb)$ is a (De-)MVN as well, because $\hb$ is affine. Moreover, we know $\hat{\Zb}$ is a (De-)MVN by assumption. We leverage the fact that they are equal on the intersection of the ball $\Bb(\zb_0,\delta)$ and the support of $\Zcal$ and use Lemma~\ref{adap_thm:local-gmm-iden} to prove that $\hb(\Zb)\equiv \hat{\Zb}$.
\end{proof}

\subsubsection{Linear Identifiability given $\Yb=\yb$ for Piecewise Linear $\fb$} \label{app:proof affine given masks}

We now show that
given the information of the binary mask $\Yb$, we can identify the latent factors $\Zb$ up to an affine transformation (Def.~\ref{def: linear ident}). 
It is crucial to emphasize that in this paper, having $\mathbf{Y} = \mathbf{y}$ does not imply knowledge of the exact value of $\mathbf{y}$; instead, it simply needs the information on \emph{grouping}, i.e. the partitioning of the dataset based on mask values.

\begin{restatable}[Linear Identifiability given $\Yb=\yb$ for Piecewise Linear $\fb$]{lemma}{lemmapcaffine}
\label{lemma: affine given mask}
Assume the observation $\Xb = \fb(\Zb)$ follows the data-generating process in Sec.~\ref{sec: Problem set up}, Ass.~\ref{assump: gaussian dist} and\ref{assump:independent masks} hold, and $\fb: \Zcal \to \Xcal$ is an injective piecewise linear function. Let $\gb: \Xcal \rightarrow \mathbb{R}^n$ be a %
continuous piecewise linear function and $\hat\fb: \mathbb{R}^n \rightarrow \mathbb{R}^d$ be an injective piecewise linear function. %
If both following conditions hold,
\begin{align}
    \mathbb{E}\norm{\Xb - \hat\fb(\gb(\Xb))}^2_2 &= 0 \,,
    \label{eq:zero_reconstruction} \text{and}\\ 
    \gb(\Xb)~|~(\Yb=\yb) &\sim N(\mub_{\yb},\Sigmab_{\yb}),
\end{align}
for some $ \mub_{\yb} \in \mathbb{R}^{n}, \Sigmab_{\yb} \in \mathbb{R}^{n \times n}$,
then $\Zb\mid (\Yb = \yb)$ is identified by $\hat\fb^{-1}(\Xb)\mid (\Yb = \yb)$ up to affine transformation, i.e., there exists an invertible affine function $\hb_{\Yb}: \Zcal_{\Yb} \rightarrow \RR^n$, such that $\hb_{\yb}(\Zb) \mid (\Yb=\yb) \equiv \gb(\fb(\Zb)) \mid (\Yb =\yb)$. 
\end{restatable}

\begin{proof}
We start by introducing some additional notation.

\begin{definition} 
\label{def:Z_y}
Let $\vb: \Zcal \rightarrow \mathbb{R}^n$ be a function with variables $\zb=(z_1,...,z_n) \in\Zcal$. Let $\yb \in [0,1]^n$ be a $n$-dimensional binary mask. We define the \emph{sub-support space} as 
\begin{align}
\Zcal_{\yb}=\{\zb\in \Zcal: \mathbbm{1}(z_i=0)\geq \mathbbm{1}(\yb_i=0), \quad \forall i\in[n] \}.
\end{align}
Intuitively, this means if $\yb_i=0$, this must cause $\zb_i=0$.

Alternatively, let $\sb \subseteq [n]$ be the index set of nonzero elements in $\yb$, we can define
\begin{align}
\Zcal_{\sb}=\{\zb\in \Zcal: \mathbbm{1}(z_i=0)\geq \mathbbm{1}(i \not\in \sb), \quad \forall i\in[n] \}.
\end{align}
\end{definition}

From the perfect reconstruction constraint \eqref{eq:zero_reconstruction1}, we can derive
\begin{align}
     \mathbb{E}||\Xb - \hat\fb(\gb(\Xb))||^2_2 &= 0 \\
     \mathbb{E}||\fb(\Zb) - \hat\fb(\gb(\fb(\Zb))))||^2_2 &= 0 \\
    \mathbb{E}\left\{ \mathbb{E}\left[||\fb(\Zb) - \hat\fb(\gb(\fb(\Zb)))||^2_2 \mid \Yb\right] \right\} &= 0 \\
    \mathbb{E}\left[||\fb(\Zb) - \hat\fb(\gb(\fb(\Zb)))||^2_2 \mid \Yb\right]&=0 \quad \mathbb{P}_{\Yb}\mathrm{-a.e.}
    \label{eq:condition_expectation_zero}
\end{align}
by first substituting $\Xb = \fb(\Zb)$, then applying the law of total expectation and finally using the fact that the sum of squares is a positive function. 
Finally $\Yb$ is a discrete random variable, in this case $\mathbb{P}_{\Yb}$-almost everywhere means everywhere on its support.
We now denote $\vb:=\gb \circ \fb : \RR^n\rightarrow \mathbb{R}^n$.
Then, following \eqref{eq:condition_expectation_zero}, we have for any value $\yb \in \mathcal{Y}$ we have
\begin{align}
    \mathbb{E}\left[||\fb(\Zb) - \hat\fb(\vb(\Zb))||^2_2 \mid \Yb = \yb\right]=0,
\end{align}
This means that for the data that satisfy $\Yb=\yb$, $\fb(\Zb)$ and $\hat\fb(\vb(\Zb))$ are equal $\mathbb{P}_{\Zb|\Yb}$-almost everywhere, which implies $\fb(\Zb)$ and $\hat\fb(\vb(\Zb))$ are equally distributed.
Since by assumption $\Zb$ and $\vb(\Zb)=\gb(\Xb)$ are potentially degenerate MVNs, by Theorem~\ref{adap_thm:identif-inv-affine}, there exists an invertible affine transformation %
$\hb_{\yb}:\RR^n\rightarrow \mathbb{R}^n$ such that 
\begin{align}
\hb_{\yb}(\Zb) \equiv \vb(\Zb) \, .
\label{equ h_Y}
\end{align}
This proves that for data coming from the same mask $\Yb=\yb$ (potentially unknown), we can identify the masked causal variables mixed through a piecewise linear function up to a linear transformation.
\end{proof}

\subsubsection{Combining results and concluding the proof of Thm~\ref{thm: disentanglement piecewise}}

Before we prove Theorem~\ref{thm: disentanglement piecewise}, we first report an adapted version of Lemma D.2 from \citet{kivva2022identifiability}, which shows the linearity of the function $\fb$ if both $\Zb$ and $\fb(\Zb)$ follow the same MVN distribution. 

\begin{lemma}(Linearity of $\fb$ for non-degenerate case \citep{kivva2022identifiability})
\label{lemma: linearity of f for non-degenerate}
Let $\Zb \sim N(\mub, \Sigmab)$, where $|\Sigmab|> 0$, i.e. $\Zb$ is a non-degenerate MVN. Assume that $\fb: \RR^n \rightarrow \RR^n$ is a continuous  piecewise affine function such that $\fb(\Zb) \equiv \Zb$, i.e. $\fb(\Zb)$ is equal in distribution to $\Zb$ . Then $\fb$ is affine.%
\end{lemma}

The original Lemma is proved by contradiction. We extend this result to the case of degenerate multivariate normal variables.

\begin{lemma}[Linearity of $\fb$ for (De) MVNs]
\label{lemma: linearity of f for degenerate}
Let $\Zb \sim N(\mub, \Sigmab)$, where $|\Sigmab|\geq 0$, i.e. $\Zb$ is potentially a degenerate multivariate normal. Assume that $\fb: \RR^n \rightarrow \RR^n$ is a continuous  piecewise affine function such that $\fb(\Zb) \equiv \Zb$. Then $\fb$ is affine over $\Zcal$, the support of $\Zb$.
\end{lemma}

\begin{proof}
First, for the simplest case, if $rank(\Sigma)=0$, then, $\Zcal$ is a single point $\{\mub\}$, and $\fb$ is affine over $\Zcal$. %

If $rank(\Sigma)>0$, by the definition of (De) MVN, we know that $\Zb = \Ab \boldsymbol \varepsilon + \bb$, where $\boldsymbol \varepsilon \sim N(\mathbf{0},\Ib)$ with dimension $d_{\boldsymbol \varepsilon}$, and $\Ab \in \RR^{n \times d_{\boldsymbol \varepsilon}}$ has full column rank. This implies that $\Zcal=colsp(A)+\bb$. 

If we substitute $\Zb = \Ab \boldsymbol \varepsilon + \bb$ into $\fb(\Zb) \equiv \Zb$, we get
\begin{align}
      &\fb(\Ab \boldsymbol \varepsilon + \bb) \equiv \Ab \boldsymbol \varepsilon + \bb\\
     &\fb(\Ab \boldsymbol \varepsilon + \bb)-\bb \equiv \Ab \boldsymbol \varepsilon\\
     &(\Ab^{T}\Ab)^{-1}\Ab^{T}\left[\fb(\Ab \boldsymbol \varepsilon + \bb)-\bb\right] \equiv  \boldsymbol \varepsilon
\end{align}
The $\equiv$ symbol denotes that $(\Ab^{T}\Ab)^{-1}\Ab^{T}\left[\fb(\Ab \boldsymbol \varepsilon + \bb)-\bb\right]$ is equal to $\boldsymbol \varepsilon$ in distribution, which means it is not necessarily the same in each point of the support.

By Lemma~\ref{lemma: linearity of f for non-degenerate}, we can derive the function $(\Ab^{T}\Ab)^{-1}\Ab^{T}\left[\fb(\Ab \boldsymbol \varepsilon + \bb)-\bb\right]$ is affine over $\RR^{d_{\boldsymbol \varepsilon}}$. Therefore, there exists $\Mb \in \RR ^{d_{\boldsymbol \varepsilon} \times d_{\boldsymbol \varepsilon}}$, and $\cb \in \RR^{d_{\boldsymbol \varepsilon}}$ such that for every $\boldsymbol{\varepsilon}_0 \in \RR^{d_{\boldsymbol \varepsilon}}$:

\begin{align}
     & (\Ab^{T}\Ab)^{-1}\Ab^{T}\left[\fb(\Ab \boldsymbol \varepsilon_0 + \bb)-\bb\right]=\Mb \boldsymbol \varepsilon_0 + \cb\\
     & \Ab(\Ab^{T}\Ab)^{-1}\Ab^{T}\left[\fb(\Ab \boldsymbol \varepsilon_0 + \bb)-\bb\right]=\Ab(\Mb \boldsymbol \varepsilon_0 + \cb)
\end{align} 

Since $\Ab(\Ab^{T}\Ab)^{-1}\Ab^{T}$ is a projection and its image space is $colsp(A)$, then, it is the identity operator $\Ib$ on $colsp(A)$.

Since $\fb(\Zb)\equiv \Zb$, then, the support of $\fb(\Zb)$ should be the same as the support of $\Zb$, which is $colsp(A)+\bb$. This implies $\fb(\Ab \boldsymbol \varepsilon_0 + \bb)-\bb \in colsp(A)$. Then, we have
\begin{align}
     & \fb(\Ab \boldsymbol \varepsilon_0 + \bb)-\bb=\Ab(\Mb \boldsymbol \varepsilon_0 + \cb)\\
     &\fb(\Ab \boldsymbol \varepsilon_0 + \bb)=\Ab(\Mb \boldsymbol \varepsilon_0 + \cb)+\bb.
     \label{equ: good equ}
\end{align} 
By definition, $\boldsymbol \varepsilon_0=(\Ab^{T}\Ab)^{-1}\Ab^{T}(\zb_0-\bb)$ for every $\zb_0 \in \Zcal$. We substitute this into Equation~\ref{equ: good equ}, and we get
\begin{align}
   \fb(\zb_0)=\Ab [\Mb (\Ab^{T}\Ab)^{-1}\Ab^{T}(\zb_0-\bb)+\cb]+\bb.
\end{align}
Hence, we can conclude $\fb$ is a linear function over $\Zcal$.
\end{proof}

With these results, we can now prove the following Theorem~\ref{thm: disentanglement piecewise}.

\thmpiecewiselinear*

\begin{proof}
Since $\Xb = \fb(\Zb)$, we can rewrite Equation \eqref{eq:zero_reconstruction1} (perfect reconstruction) as
    \begin{align}
        \mathbb{E}||\fb(\Zb) - \hat\fb(\gb(\fb(\Zb)))||^2_2 = 0\,.
    \end{align}
This means $\fb$ and $\hat\fb\circ\gb\circ\fb$ are equal $\mathbb{P}_\Zb$-almost everywhere.

Since $\fb$, $\gb$ and $\hat{\fb}$ are continuous, we can derive $\fb$ and $\hat\fb\circ\gb\circ\fb$ must be equal over the support of $\Zb$, $\mathcal{Z}$,
i.e.,
    \begin{align}
        \fb(\zb) = \hat \fb \circ \gb \circ \fb (\zb)\,, \forall \zb \in \mathcal{Z}\,.
    \end{align}
We denote $\vb:=\gb \circ \fb : \RR^n \rightarrow \mathbb{R}^n$, an invertible continuous piecewise affine function. Take the left inverse of $\hat{\fb}$ on both sides, then we have,
 \begin{align}
        \hat{\fb}^{-1}\circ\fb(\zb) = \vb (\zb)\,, \forall \zb \in \mathcal{Z}\,.
    \end{align}
From Lemma~\ref{lemma: affine given mask}, we know that for all $\yb \in \Ycal$, given the mask $\Yb=\yb$, there exists an invertible affine transformation $\hb_{\yb}:\mathbb{R}^n\rightarrow \mathbb{R}^n$ such that $\vb(\Zb) \equiv \hb_{\yb}(\Zb)$ over $\Zcal_{\yb}$. %

Since we know $\Zb|\Yb=\yb$ follows a (De)MVN and $\hb_{\yb}$ is affine, by Lemma~\ref{lemma: close affine}, we can derive that $\hb_{\yb}(\Zb)$ follows a (De-)MVN distribution as well. Then, we can rewrite $\vb(\Zb) \equiv \hb_{\yb}(\Zb)$ in distribution as 
 \begin{align}
   \hb_{\yb}(\Zb) \equiv \vb \circ {\hb_{\yb}}^{-1} \circ \hb_{\yb}(\Zb).
\end{align}
Since $\hb_{\yb}$ is affine, its inverse is continuous. In addition, $\vb$ is continuous as it is the composition of continuous functions $\fb$ and $\gb$. Therefore, $\vb \circ {\hb_{\yb}}^{-1}$ is continuous piecewise affine. By using Lemma~\ref{lemma: linearity of f for degenerate}, we get that $\vb \circ {\hb_{\yb}}^{-1}$ is affine over $\hb_{\yb}(\Zcal_{\yb})$, i.e. there exists an invertible affine map $\hb_1$ such that 
\begin{align}
   \vb \circ {\hb_{\yb}}^{-1}(\zb)=\hb_1(\zb) \quad \iff \quad \vb(\zb)=\hb_1\circ\hb_{\yb}(\zb) \quad \forall \zb \in \Zcal_{\yb}.
\end{align}

Since $\vb$ is a composition of affine maps, it is also affine on $\Zcal_{\yb}$, $\forall \yb \in \Ycal$.

We define $\Zcal_{\sb}:=\prod_{i=1}^{n} (\RR \cdot \mathbbm{1}\{i \in \sb\})$, %
which is the space of the supports for each dimension given a value of the support index $\sb$. For the dimensions which are included in $\sb_i$, this is  $\RR$, while for the others it is $0$.  By this definition and by the sufficient variability assumption Ass.~\ref{assump: sufficient support},  $\Zcal=\bigcup_{\sb\in \Scal}\Zcal_{\sb}$.%

Since $\Scal$ is a finite set, which implies countable, we can find a way to order the elements in $\Scal$, denoted as $\{\sb_1, ..., \sb_{|\Scal|}\}$. Thus, $\Zcal=\bigcup_{i=1}^{|\Scal|}\Zcal_{\sb_i}$.

While we have already proven that $\vb$ is affine over each subspace $\Zcal_{\sb_i}$, we now show that $\vb$ is a linear function on $\Zcal$, i.e. $\vb(\zb)=\wb\zb, \forall \zb\in\Zcal$, where $\wb \in \RR^{n\times n}$.

\begin{itemize}
\item[i] We first consider two index sets $\sb_1$ and $\sb_2$. Without loss of generality, we assume the index in $\sb_1$ is from 1 to $|\sb_1|$, and the index in $\sb_2$ is from $m$ to $m+|\sb_2|$, where $m \in \{1,2,...,n-|\sb_2|+1\}$. Since $\vb$ is affine on $\Zcal_{\sb_1}$ and $\Zcal_{\sb_2}$ individually, we have 
\begin{align}
   &\vb(\zb)=\zb_1 \ab_1+...+\zb_{|\sb_1|}\ab_{|\sb_1|}+\cb_1 \quad \forall \zb \in \Zcal_{\sb_1}\\
   &\vb(\zb)=\zb_m \bb_m+...+\zb_{|\sb_2|+m}\bb_{|\sb_2|+m}+\cb_2 \quad \forall \zb \in \Zcal_{\sb_2}. 
\end{align}
Since $\Zcal_{\sb_1}\cap \Zcal_{\sb_2}=\{\mathbf{0}\}$, then we can get $\cb_1=\cb_2$.
\begin{itemize}
\item[Case 1.] $\sb_1\cap \sb_2=\emptyset$. Then, we have $\forall \zb \in \Zcal_{\sb_1}\cup\Zcal_{\sb_2}$
\begin{align}
   \vb(\zb)=\zb_1 \ab_1+...+\zb_{|\sb_1|}\ab_{|\sb_1|}+\zb_m \bb_m+...+\zb_{|\sb_2|+m}\bb_{|\sb_2|+m}+\cb_1
\end{align}
\item[Case 2.] $\sb_1\cap \sb_2 \neq \emptyset$. Without loss of generality, we assume $|\sb_2|\leq |\sb_1|$.
\begin{itemize}
\item[(a).] $\sb_2 \subseteq \sb_1$. Then, we can directly get $\forall \zb \in \Zcal_{\sb_1}\cup\Zcal_{\sb_2}$
\begin{align}
   \vb(\zb)=\zb_1 \ab_1+...+\zb_{|\sb_1|}\ab_{|\sb_1|}+\cb_1
\end{align}

\item[(b).]$\sb_2 \not \subseteq \sb_1$. Without loss of generality, we assume $\sb_1\cap\sb_2=\{m,...,m+t\}$, where $t\in \{1,...,|\sb_2|\}$. Then, $\forall \zb \in \Zcal_{\sb_1}\cap\Zcal_{\sb_2}$, we have
\begin{align}
   &\vb(\zb)=\zb_m \ab_m+...+\zb_{m+t}\ab_{m+t}+\cb_1 \\
   &\vb(\zb)=\zb_m \bb_m+...+\zb_{m+t}\bb_{m+t}+\cb_1. 
\end{align}
This implies $\ab_i=\bb_i, i=m,...,m+t$. Therefore, we can derive $\forall \zb \in \Zcal_{\sb_1}\cup\Zcal_{\sb_2}$,
\begin{align}
   \vb(\zb)=\zb_1 \ab_1+...+\zb_{|\sb_1|}\ab_{|\sb_1|}+\zb_{m+t+1} \bb_{m+t+1}+...+\zb_{|\sb_2|+m}\bb_{|\sb_2|+m}+\cb_1.
\end{align}

\end{itemize}

\end{itemize}
By considering both cases, we can now proof that $\vb$ is a linear function on $\Zcal_{\sb_1}\cup\Zcal_{\sb_2}$.

\item[ii] We can iterate this strategy by iteratively adding new $\Zcal_{\sb_i}$ to the union, until $\Zcal_{|\Scal|}$. Finally, we have that $\vb$ is a linear function on $\bigcup_{\sb\in \Scal}\Zcal_{\sb}=\Zcal$.

\end{itemize}

Then, the rest proof immediately follows from Lemma~\ref{lemma: element_wise v}, where we have proven element-wise identifiability for the linear transformation.

\end{proof}

\section{Example: sparsity is not enough for identifiability in the non-linear case.} \label{app:counterexample}
In Theorem~\ref{thm: linear disentangle} we prove that we can achieve element-wise identifiability for an invertible linear mixing function $\fb$ in the Partially Observable Causal Representation Learning setting, assuming sufficient support index variability (Ass.~\ref{assump: sufficient support}) and under the condition of perfect reconstruction and a sparsity principle. 

An obvious extension of this result might be considering the identifiability for non-linear mixing functions. In this section we show a counter-example that describes why this is not possible in general without any further assumption (e.g. assuming both a piecewise linear mixing function and a Gaussian causal model, as shown by our results in Thm.~\ref{thm: disentanglement piecewise}). 

\begin{figure}
    \centering
    \includegraphics[width=0.6\linewidth]{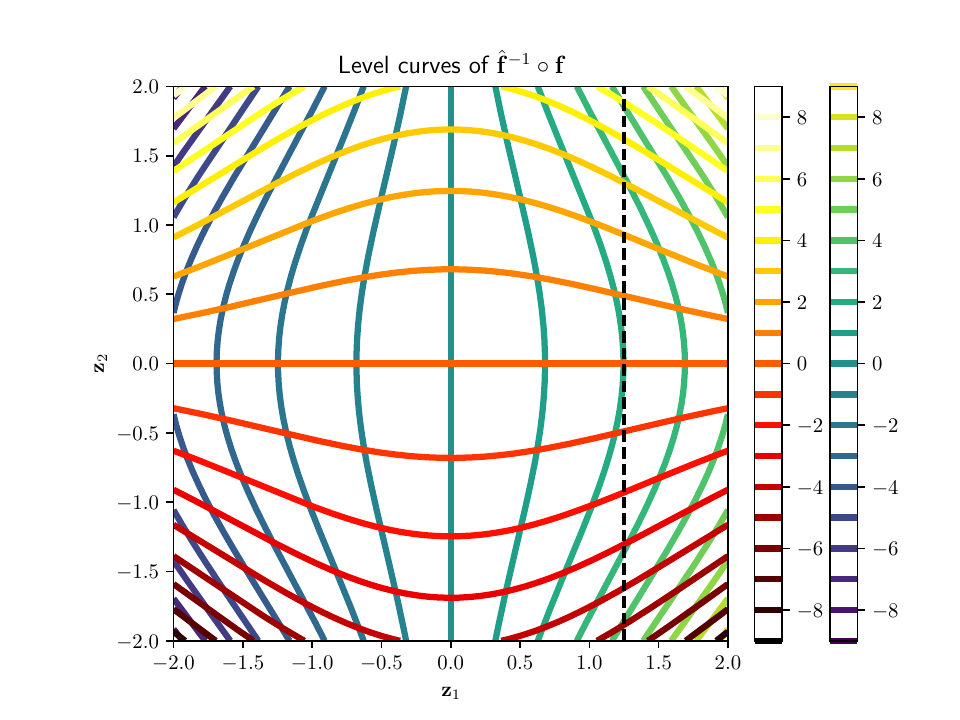}
    \caption{Level curves of the function $\hat\fb^{-1} \circ \fb$ of Example~\ref{ex:counter_example1}.  The cold color scheme corresponds to the level curves of $(\hat\fb^{-1} \circ \fb)_1(\zb)$ while the warm color scheme corresponds to $(\hat\fb^{-1} \circ \fb)_2(\zb)$. The example gives a concrete case where all assumptions of Theorem~\ref{thm: linear disentangle} hold except for the linearity of $\fb$. We can see that $\hat\fb^{-1} \circ \fb$ is not a permutation composed with an element-wise invertible transformation, since along the vertical dashed line, we can see that both components of $\hat\fb^{-1} \circ \fb$ change.   %
    }
    \label{fig:counter_example}
\end{figure}

\begin{example}\label{ex:counter_example1}
Assume we have 2 latent and 2 observed variables, i.e. $n=2$ and $d=2$. Furthermore, assume that the domain of the causal variables $\Ccal = \RR^2$ and that Ass.~\ref{assump: gaussian dist} and \ref{assump:independent masks} hold. For example, consider $\Cb \sim \mathcal{N}(0, \Ib_2)$ %
with an independent mask $\Yb$ with distribution $p(\Yb = \yb) = 1 / 4$ for any $\yb \in \{0,1\}^2$, which trivially satisfies Ass.~\ref{assump: gaussian dist} and \ref{assump:independent masks}. In this case, the support of $\Zb = \Yb \odot \Cb$ is $\Zcal = \RR^2$. Assume further that $\fb:\RR^2 \rightarrow \RR^2$ is defined by
    \begin{align}
        \fb(\zb) := \sinh(\Rb_{\frac{\pi}{4}}\zb) + \sinh(\Rb_{-\frac{\pi}{4}}\zb)\, ,
    \end{align}
    where $\sinh(x) := \frac{e^x - e^{-x}}{2}$ (applied element-wise above) and $\Rb_{\theta}$ is a rotation matrix defined by 
    \begin{align}
        \Rb_\theta := \begin{bmatrix}
            \cos\theta & -\sin\theta \\
            \sin\theta & \cos\theta
        \end{bmatrix} \, .
    \end{align}

Consider $\hat\fb$ and $\gb$ to be the identity function.
We now show that all the assumptions of Theorem~\ref{thm: linear disentangle} except the linearity of $\fb$ are satisfied. First, notice how 
    \begin{align}
        \mathbb{E}|| \Xb - \hat\fb(\gb(\Xb))||^2 = \mathbb{E}|| \Xb - \Xb||^2 = 0 \,.
    \end{align}

Since $\Xb =\fb(\Zb)$ by definition and $\gb(\Xb)= \Xb$, since $\gb$ is the identity function, then $\fb(\Zb) = \gb(\Xb)$ and thus $\mathbb{E}|| \gb(\Xb)||_0 = \mathbb{E}||\fb(\Zb)||_0$. 

We now show that $\mathbb{E}||\fb(\Zb)||_0 \leq \mathbb{E}||\Zb||_0$, which we will then use to prove the sparsity condition on $\mathbb{E}|| \gb(\Xb)||_0$.

We can see that 
    \begin{align}
        \fb(\begin{bmatrix}
            z_1 \\ 0
        \end{bmatrix}) &= \sinh(z_1\begin{bmatrix}
            \cos\frac{\pi}{4} \\
            \sin\frac{\pi}{4}
        \end{bmatrix}) + \sinh(z_1\begin{bmatrix}
            \cos-\frac{\pi}{4} \\
            \sin-\frac{\pi}{4}
        \end{bmatrix}) \\
        &= \sinh(z_1\begin{bmatrix}
            \cos\frac{\pi}{4} \\
            \sin\frac{\pi}{4}
        \end{bmatrix}) + \sinh(z_1\begin{bmatrix}
            \cos\frac{\pi}{4} \\
            -\sin\frac{\pi}{4}
        \end{bmatrix}) \\
        &= \begin{bmatrix}
            2\sinh(z_1 \cos\frac{\pi}{4}) \\
            \sinh(z_1 \sin \frac{\pi}{4}) + \sinh(-z_1\sin\frac{\pi}{4})
        \end{bmatrix}\\
        &= \begin{bmatrix}
            2\sinh(z_1 \cos\frac{\pi}{4}) \\
            \sinh(z_1 \sin \frac{\pi}{4}) - \sinh(z_1\sin\frac{\pi}{4})
        \end{bmatrix} 
        = \begin{bmatrix}
            2\sinh(z_1 \cos\frac{\pi}{4}) \\
            0
        \end{bmatrix}\,,
    \end{align}
    where we use the fact that $\sin$ and $\sinh$ are odd, while $\cos$ is even. An analogous argument shows that
    \begin{align}
        \fb(\begin{bmatrix}
            0 \\ z_2
        \end{bmatrix}) &= \begin{bmatrix}
            0 \\
            2\sinh(z_2\cos\frac{\pi}{4})
        \end{bmatrix}\,.
    \end{align}
We can also easily show that $\fb(0) = 0$.
The above shows that, for all $\zb \in \RR^2$, $||\fb(\zb)||_0 \leq ||\zb||_0$. By taking the expectation on both sides we get the desired result: $\mathbb{E}||\fb(\Zb)||_0 \leq \mathbb{E}||\Zb||_0$ and thus $\mathbb{E}|| \gb(\Xb)||_0 \leq \mathbb{E}||\Zb||_0$. 

However, we can see in in Figure~\ref{fig:counter_example} and in the computation below, $\hat\fb^{-1} \circ \fb = \fb$ is not a permutation composed with an element-wise invertible transformation on $\RR^2$. 
\begin{align}
        \fb(\begin{bmatrix}
            z_1 \\ z_2
        \end{bmatrix}) &= 
        \begin{bmatrix}
            \sinh(\cos\frac{\pi}{4} z_1 - \sin\frac{\pi}{4} z_2) + \sinh(\cos\frac{\pi}{4} z_1 + \sin\frac{\pi}{4} z_2)\\
            \sinh(\sin\frac{\pi}{4} z_1 + \cos\frac{\pi}{4} z_2) - \sinh( \sin\frac{\pi}{4} z_1 - \cos\frac{\pi}{4} z_2)
        \end{bmatrix} 
\end{align}

This counter-example shows that the results of Theorem~\ref{thm: linear disentangle} do not apply in general if $\fb$ is nonlinear, but only with additional assumptions, as shown in Theorem~\ref{thm: disentanglement piecewise}.
\end{example}

\section{Implementation details} \label{app:implementation}
This section provides further details about the experiment implementation in Section~\ref{sec: implementation}. The implementation is built upon the code open-sourced by \citet{lachapelle2022disentanglement} released under Apache 2.0 License; \citet{ahuja2022weakly}; \citet{von2021self} released under MIT License; \citet{zheng2018dags} released under Apache 2.0 License.

\subsection{Hyperparameters}
\begin{table}[h!]
\centering
\begin{tabular}{lllll}
\hline
           & Numerical linear $\fb$ & Numerical p.w. linear $\fb$ & Multiple balls  & PartialCausal3DIdent \\ 
           & Section~\ref{sec:numericalexp}  & Section~\ref{sec:numericalexp}  & Section~\ref{sec:multiballexp}  & Section~\ref{sec:causal3didentexp} \\ 
\hline
$\epsilon$       & 0.001                            & 0.01                                & 0.01            & 0.01                 \\
Optimizer    & ExtraAdam                         & ExtraAdam                            & ExtraAdam       & ExtraAdam           \\
Primal optimizer learning rate   & 1e-4                         & 5e-5                            & 1e-5       & 1e-4           \\
Dual optimizer learning rate    & 1e-4/2                         & 5e-5/2                            & 1e-5/2       & 1e-4/2           \\
Batch size       & 6144                             & 10000                                & 50$\times K$           & 13$\times K$                \\
Group number $K$ & $2^n$                            & latent size $5n$                     & latent size $n$ & 10                   \\
\# Seeds   & 20                           & 20                     & 3 & 3                  \\
\# Iterations   & 30000                           & 20000                     & 10000 & 10000                 \\
\hline
\end{tabular}
\caption{Parameters for experiments results in Sec.~\ref{sec: Experiments} and App.~\ref{app:experiments}.}
\end{table}

\subsection{Sensitivity analysis on $\epsilon$}
\label{app: sensitivity analysis}
\begin{figure*}[t]
    \begin{center}
        \includegraphics[width=0.3\textwidth]{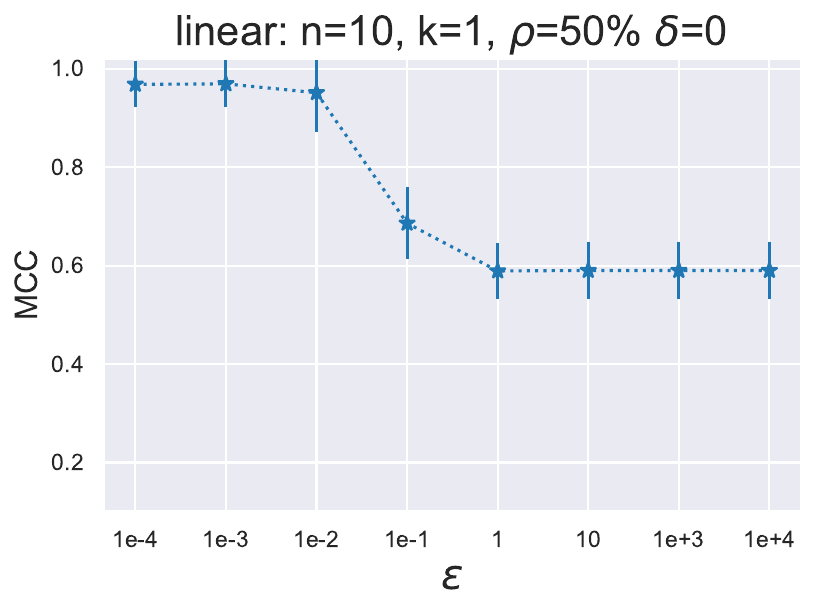}
        \hspace{2cm} 
        \includegraphics[width=0.3\textwidth]{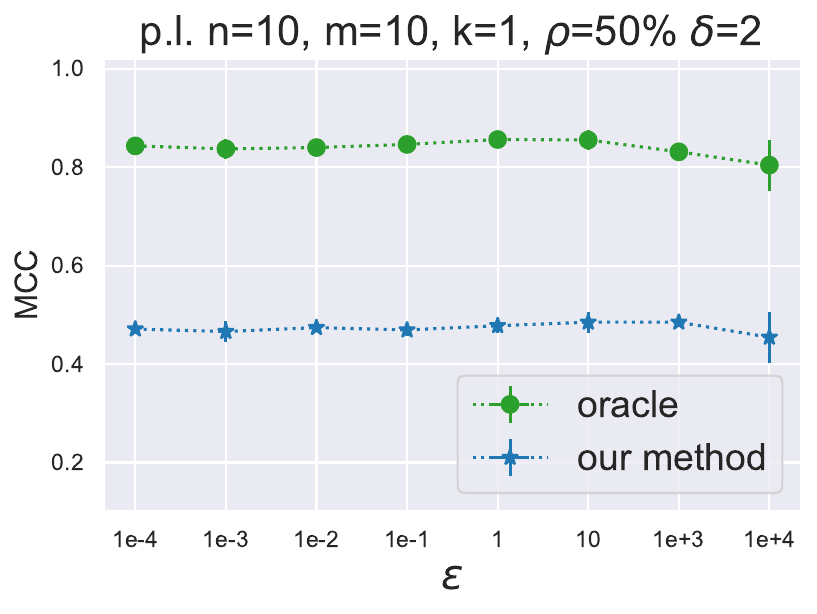}
        \caption{Sensitivity analysis on $\epsilon$ varies from $1e-4$ to $1e4$. The left graph is for the linear case, and right-hand side is for the piecewise linear case. 
        }
        \label{fig: sensitivity}
    \end{center}
\end{figure*}

We conduct a sensitivity analysis on $\epsilon$. As shown in \cref{fig: sensitivity}, we observe that the linear case exhibits heightened sensitivity $\epsilon$, whereas the piecewise linear case does not display such sensitivity. We attribute this difference to the additional group and mask information given in the training phase for piecewise linear case, enhancing the method's robustness to variations in this hyperparameter.
\FloatBarrier

\subsection{Oracle method}
\label{app: oracle}
\begin{figure*}[!hbtp]
    \begin{center}
        \includegraphics[width=0.18\textwidth]{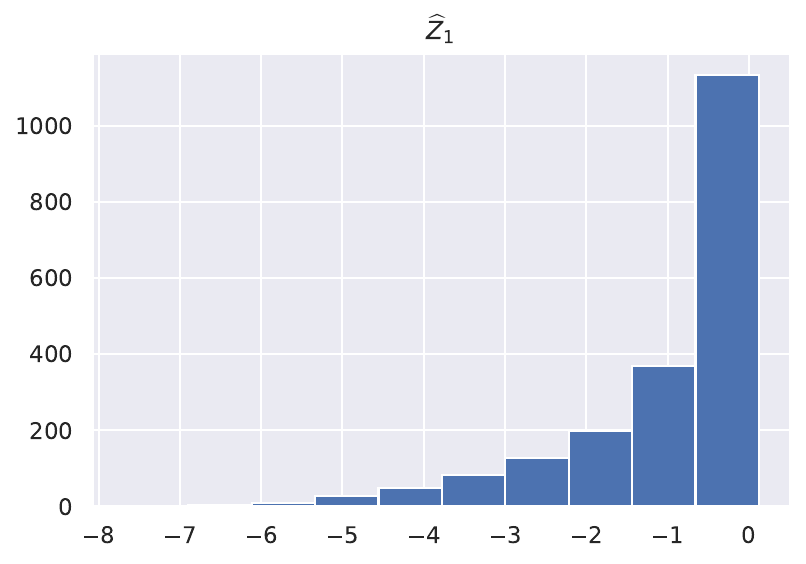}
        \hspace{0cm} 
        \includegraphics[width=0.18\textwidth]{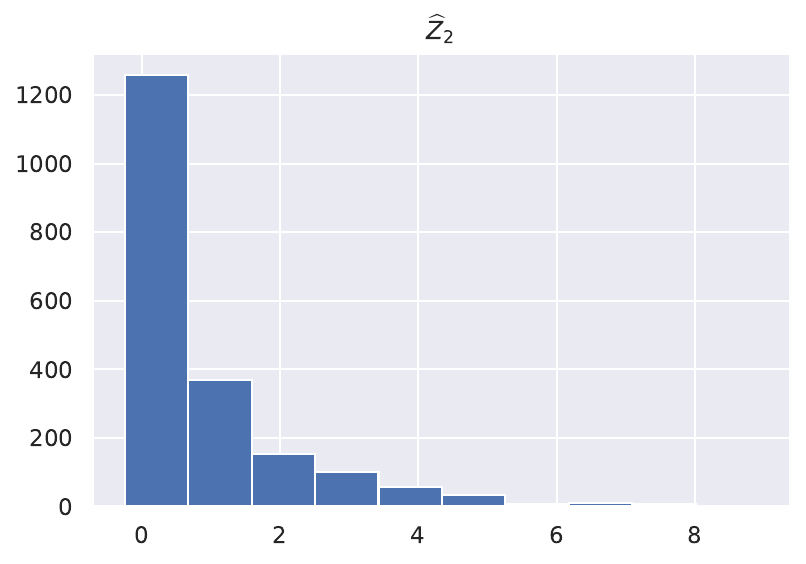}
        \hspace{0cm} 
        \includegraphics[width=0.18\textwidth]{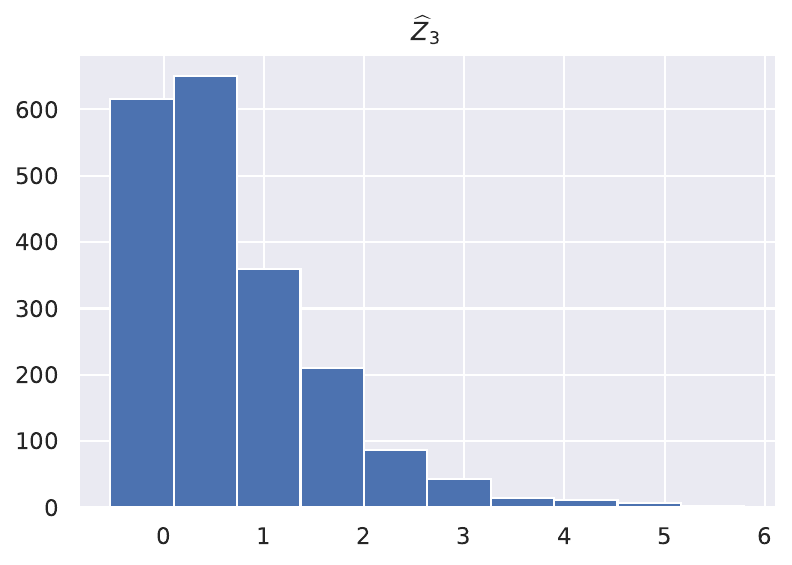}
        \hspace{0cm} 
        \includegraphics[width=0.18\textwidth]{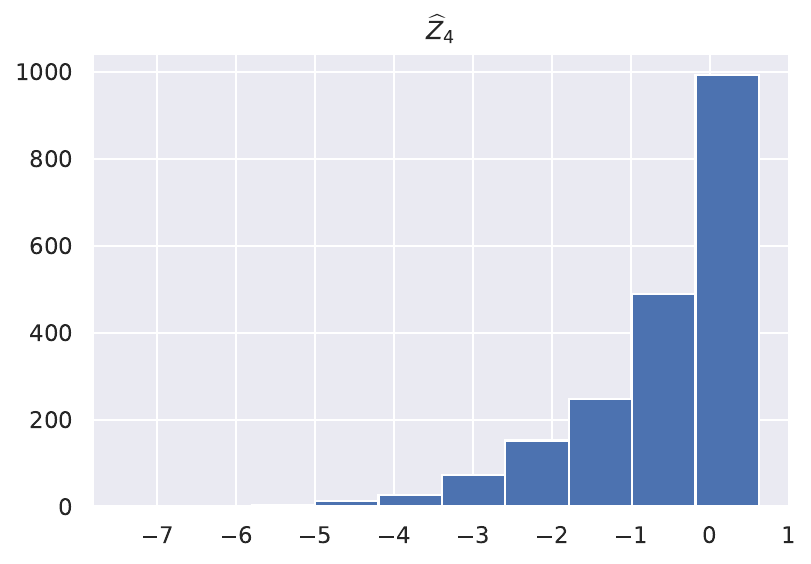}
        \hspace{0cm} 
        \includegraphics[width=0.18\textwidth]{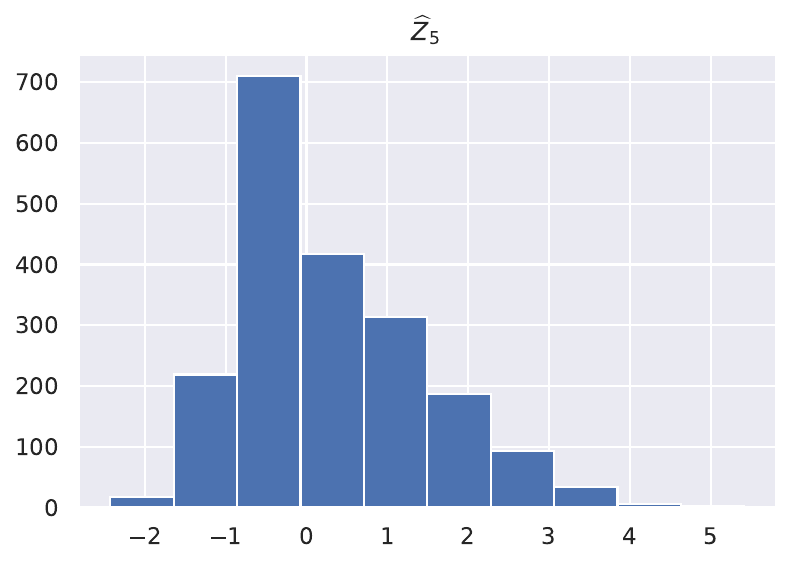}
        \hspace{0cm} 
        \caption{Histogram of estimator for our method for $n=5$, $m=10$, independent $\Zb$, $\rho$=50\%,
        }
        \label{fig:non_gaussian}
    \end{center}
\end{figure*}

In Sec.~\ref{sec: implementation}, to encourage Gaussianity of learned representations, we add a regularization term to force the sample skewness and kurtosis to match the Gaussian distribution. However, as we mention in Sec.~\ref{sec: implementation}, estimated skewness and kurtosis cannot
guarantee Gaussianity. In \cref{fig:non_gaussian}, we empirically show even by adding the two penalty terms into the loss function, the estimator we obtain is still highly non-gaussian.

In the Oracle method, we adopt the assumption of knowing masks in train phases. Instead of directly estimating the unmeasured part as a constant value, we provide less information by replacing $\gb_\psi(\xb_i)$ with a low log standard deviation $-10$ in Eq.~\ref{equ: loss_pw}; for the measured latent variables, we assign $1$ to $\gb_\psi(\xb_i)$. After training, we obtain the encoder function $\gb$. In the test phase, data in distinct groups can be mixed together. There is no longer a requirement for mask-related information. In addition, we find empirically that if we replace $\hat{\mub}_{g}$ with a constant value, e.g., 2, it also enhances the performance for most setups except for the case when $\delta=0$. When $\delta=0$, set the constant value to be 0 can obtain better results.

\FloatBarrier

\subsection{Predefined masks}
\label{app: masks}
In order to satisfy Ass.~\ref{assump: sufficient support}, i.e. for each latent variable $C_i$ $i\in[n]$, all the other latent variables should be measured in at least one of the groups where $C_i$ is unmeasured, the minimum number of distinct masks is equal to the latent size $n$, and the maximum number is up to $2^n$. In this paper, we consider 3 different strategies to generate masks.
\begin{itemize}
\item[i] Consider all $2^n$ possible masks as a mask set $\Yb$. For each data $\cb^i$, $i\in[N]$, we uniformly sample one mask $\yb^i$ from the mask set $\Yb$. This strategy is used in numerical experiments with linear $\fb$ since we do not need group-wise masks.
\item[ii] Define $kn$ possible masks $\{\yb_1, .., \yb_n\}$ with the same ratio of measured variables $\rho$. Define $\yb_g$, $g=[n]$ by this way: randomly set $\rho n$ elements as 1, the rest are 0s. This strategy is used in numerical experiments with piecewise linear $\fb$ ($k=5$) and multiple balls experiment ($k=1$) due to the necessity of having a specific number of samples from the same group in each batch for the computation of sample skewness and kurtosis—something strategy i) is incapable of achieving.
\item[iii] Define $n$ possible masks $\{\yb_1, .., \yb_n\}$ with different ratio of measured variables $\rho_g$, $g=[n]$ from 1 var to 100\%. Define $\yb_g$ this way: set $1$-th to $(g+\rho_g n)$-th element as 1, the rest are 0s. This is employed in the PartialCausal3DIdent experiment. 
\end{itemize}

\section{Full experimental results}
\label{app:experiments}

\subsection{Numerical experiments}
\label{app:experiments_numerical}

\subsubsection{Linear mixing function}
We first show the Pearson Correlation matrix $\mathrm{Corr}^{n\times n}_{\pi}$ with the permutation $\pi$ between ground truth latent variables $\Zb$ and the estimator $\hat{\Zb}=\gb(\Xb)$. One figure represents one ablation study in \cref{tab: num_linear}. We choose one random seed to plot for each setup. Ground truth Pear. Corr. matrix on the left shows the original linear correlation inside $\Zb$, compared with the estimator on the right-hand side.

\begin{figure*}[!htbp]
    \begin{center}
        \includegraphics[width=0.18\textwidth]{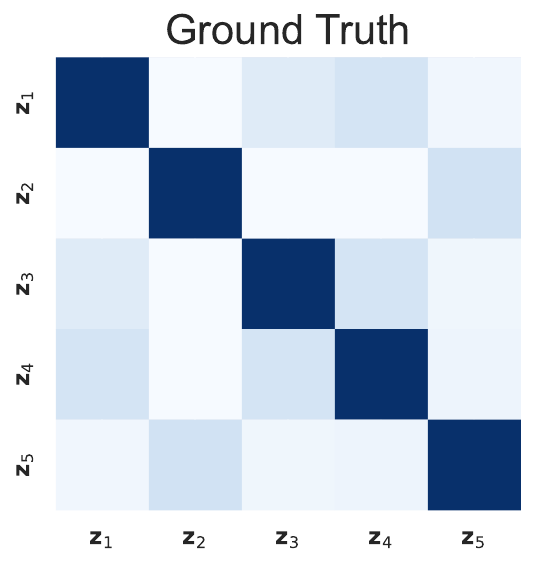}
        \hspace{0cm} 
        \includegraphics[width=0.18\textwidth]{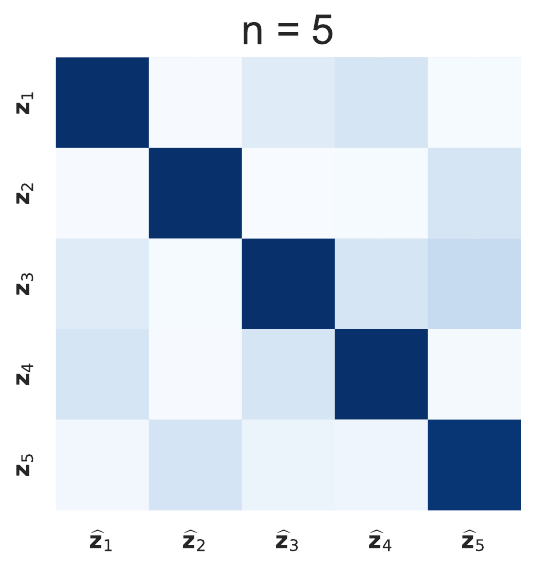}
        \hspace{0.5cm} 
         \includegraphics[width=0.18\textwidth]{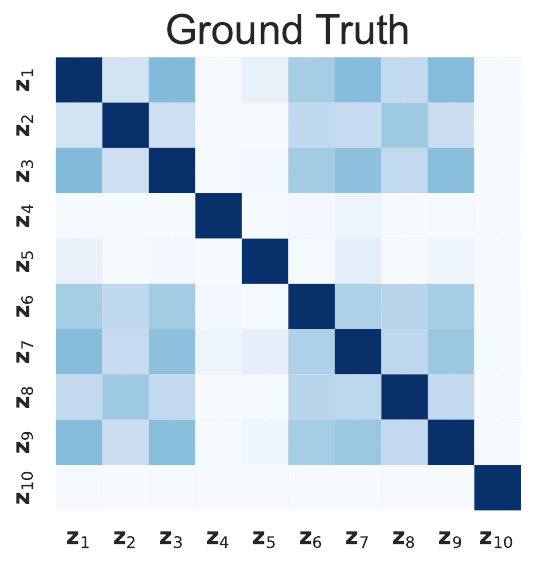}
        \hspace{0cm} 
        \includegraphics[width=0.18\textwidth]{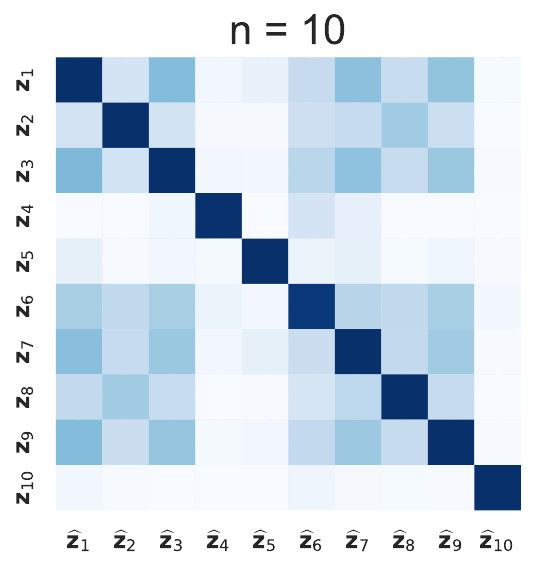}
        \includegraphics[width=0.0375\textwidth]{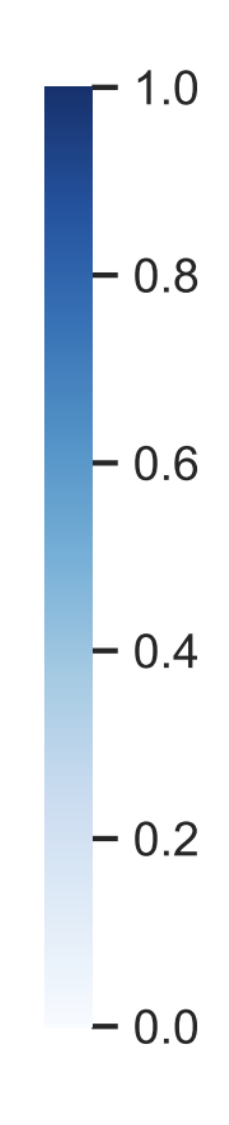}
        \hspace{2cm} 
       \includegraphics[width=0.18\textwidth]{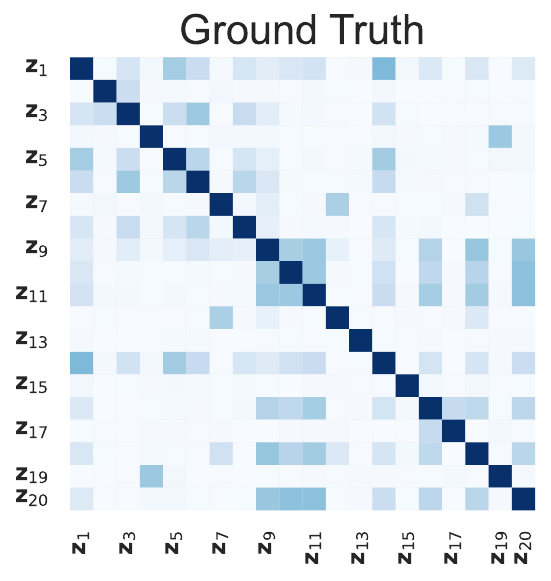}
        \hspace{0cm} 
        \includegraphics[width=0.18\textwidth]{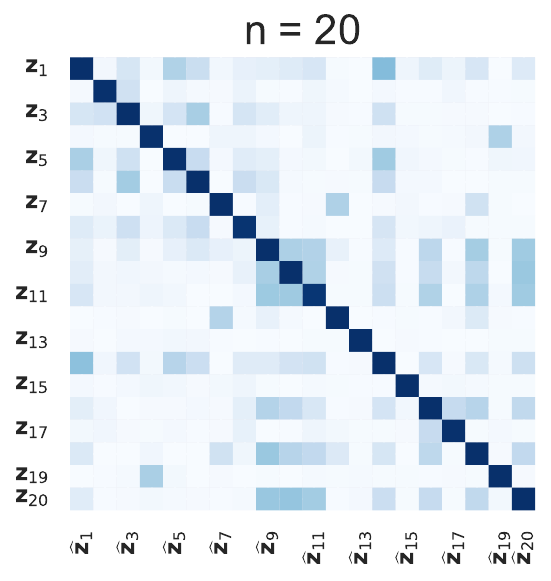}
        \hspace{0.5cm} 
        \includegraphics[width=0.18\textwidth]{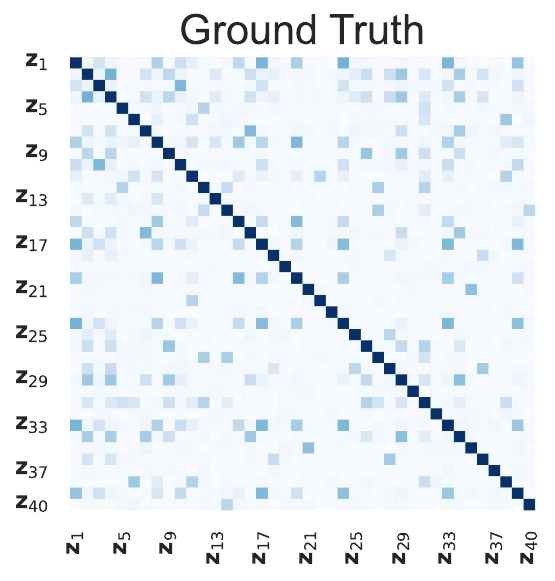}
        \hspace{0cm} 
        \includegraphics[width=0.18\textwidth]{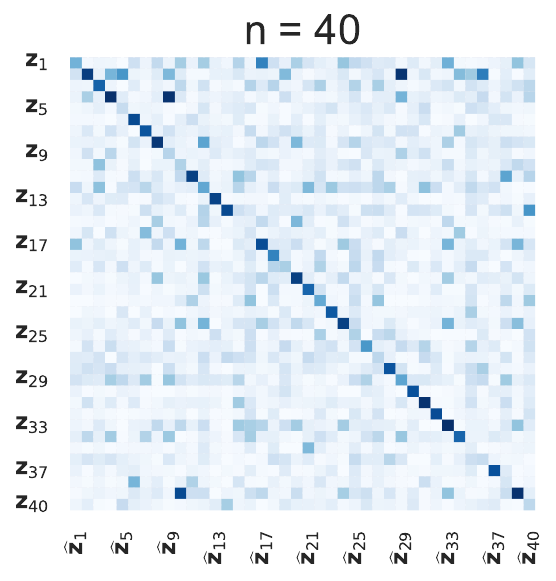}
        \includegraphics[width=0.0375\textwidth]{Figures/cbar.pdf}
        \label{fig:linear_gauss_ablation_n}
        \caption{Linear mixing function with linear causal relation and Gaussian noise: ablation study on increasing the latent dimension $n$ from 5 to 40 and fixing $\delta=0\sigma$, $\rho=50$, $k=1$. For all methods it it harder to learn an increasing number of latent variables.}
        
    \end{center}
\end{figure*}

\begin{figure*}[!htbp]
    \begin{center}
        \includegraphics[width=0.18\textwidth]{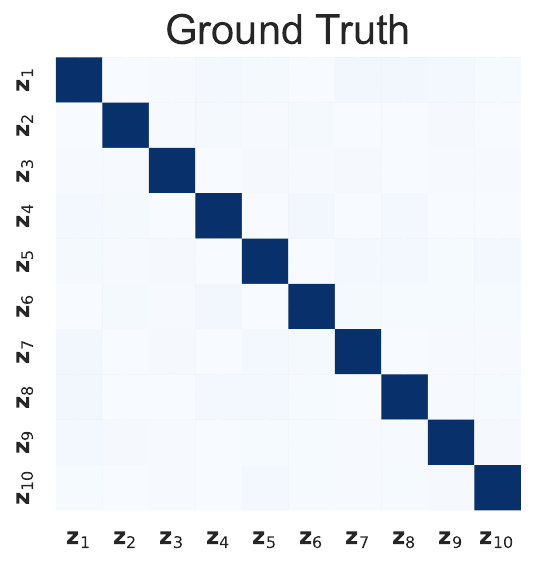}
        \hspace{0cm} 
        \includegraphics[width=0.18\textwidth]{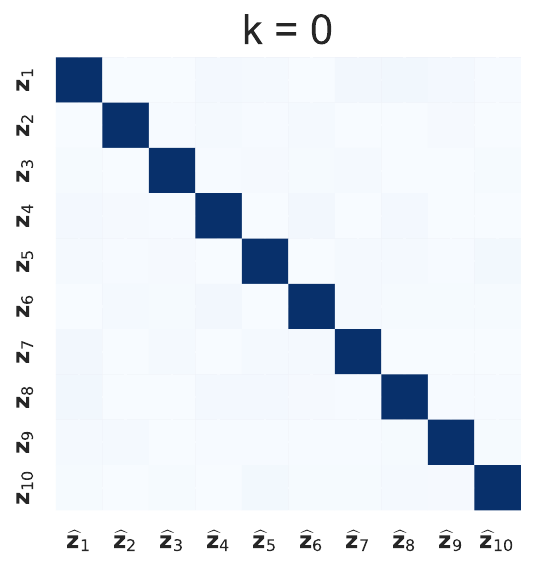}
        \hspace{0.5cm} 
         \includegraphics[width=0.18\textwidth]{Figures/heatmaps_linear_gauss/true/linear1_d0_n10_nn10_M50_G1_rs2_Corr_heatmap.pdf}
        \hspace{0cm} 
        \includegraphics[width=0.18\textwidth]{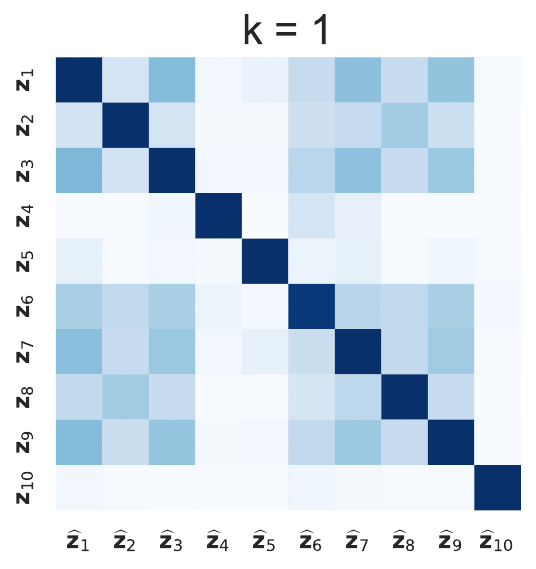}
        \includegraphics[width=0.0375\textwidth]{Figures/cbar.pdf}
        \hspace{2cm} 
      \includegraphics[width=0.18\textwidth]{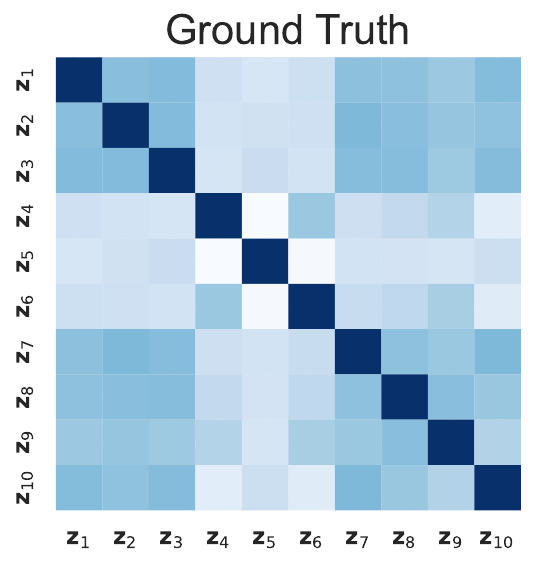}
        \hspace{0cm} 
        \includegraphics[width=0.18\textwidth]{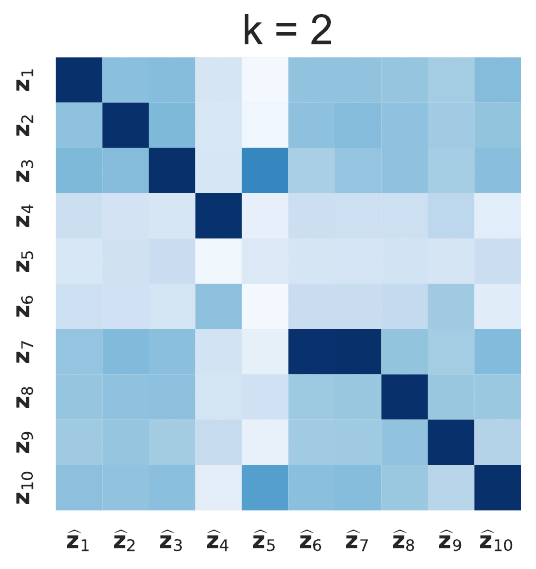}
        \hspace{0.5cm} 
        \includegraphics[width=0.18\textwidth]{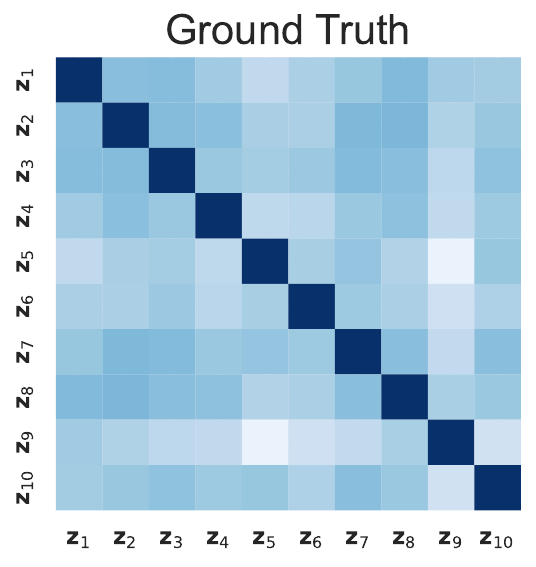}
        \hspace{0cm} 
        \includegraphics[width=0.18\textwidth]{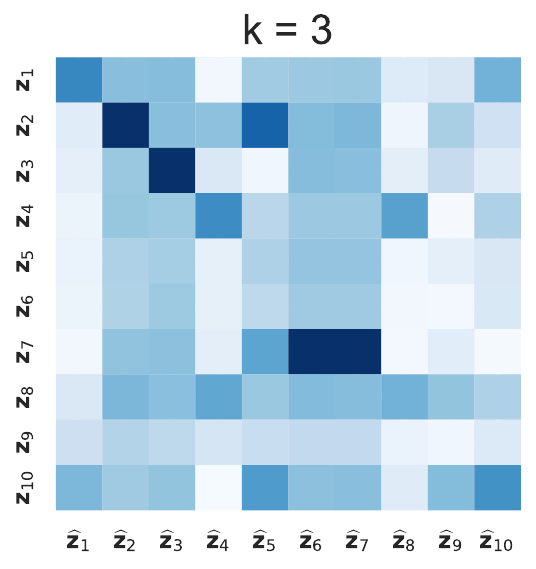}
        \includegraphics[width=0.0375\textwidth]{Figures/cbar.pdf}
        \caption{Linear mixing function with linear causal relation and Gaussian noise: ablation study on increasing the density of causal graphs $k$ from 0 to 3 and fixing $n=10$, $\delta=0\sigma$, $\rho=50$. For all methods it is harder to learn the latent variables in denser graphs.}
        \label{fig:linear_gauss_ablation_k}
    \end{center}
\end{figure*}

\begin{figure*}[!htbp]
    \begin{center}
        \includegraphics[width=0.18\textwidth]{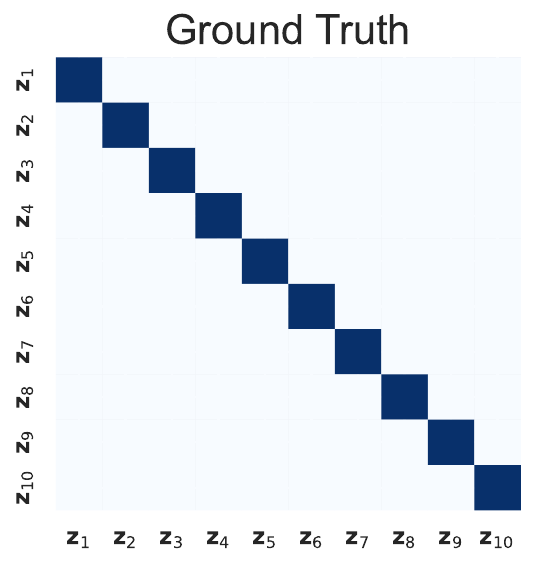}
        \hspace{0cm} 
        \includegraphics[width=0.18\textwidth]{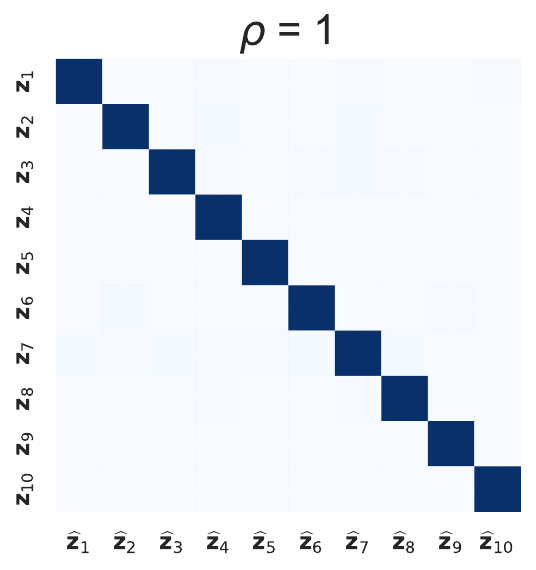}
        \hspace{0.5cm} 
         \includegraphics[width=0.18\textwidth]{Figures/heatmaps_linear_gauss/true/linear1_d0_n10_nn10_M50_G1_rs2_Corr_heatmap.pdf}
        \hspace{0cm} 
        \includegraphics[width=0.18\textwidth]{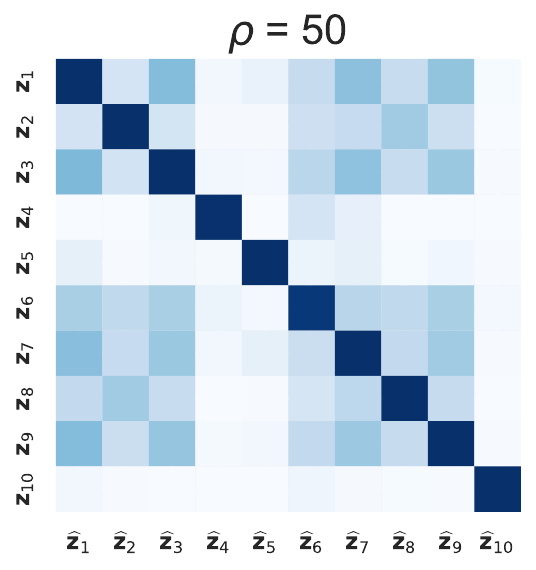}
        \includegraphics[width=0.0375\textwidth]{Figures/cbar.pdf}
        \hspace{2cm} 
      \includegraphics[width=0.18\textwidth]{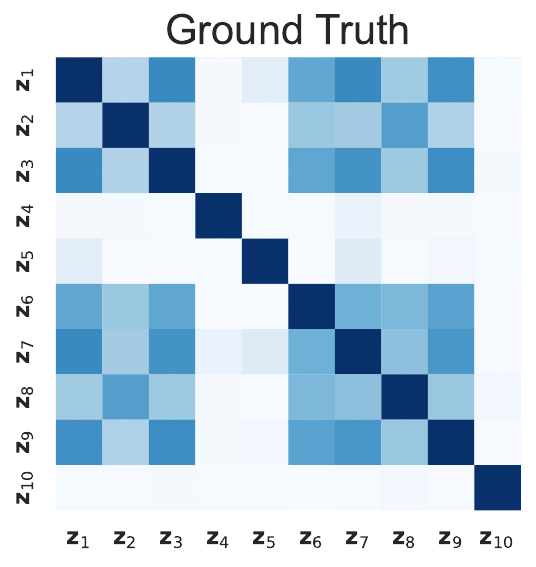}
        \hspace{0cm} 
        \includegraphics[width=0.18\textwidth]{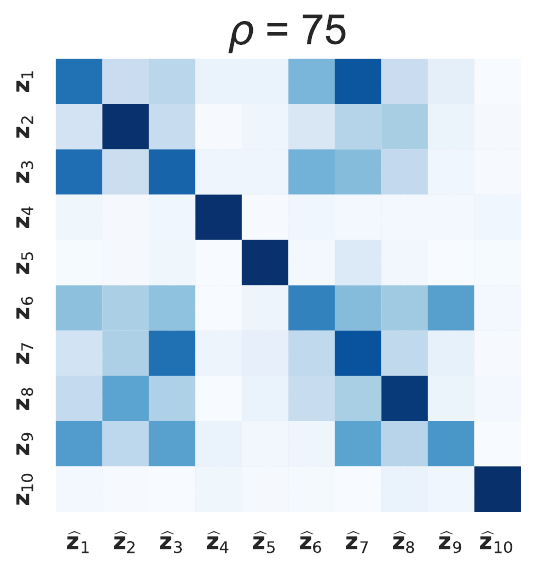}
        \includegraphics[width=0.0375\textwidth]{Figures/cbar.pdf}
        \caption{Linear mixing function with linear causal relation and Gaussian noise: ablation study on increasing the ratio of active (unmasked) variables $\rho$ from 1 variable only to 75\% and fixing $n=10$, $\delta=0\sigma$, $k=1$. Learning the latent variables is harder when there are many active variables.}
        \label{fig:linear_gauss_ablation_rho}
    \end{center}
\end{figure*}

\begin{figure*}[!htbp]
    \begin{center}
        \includegraphics[width=0.18\textwidth]{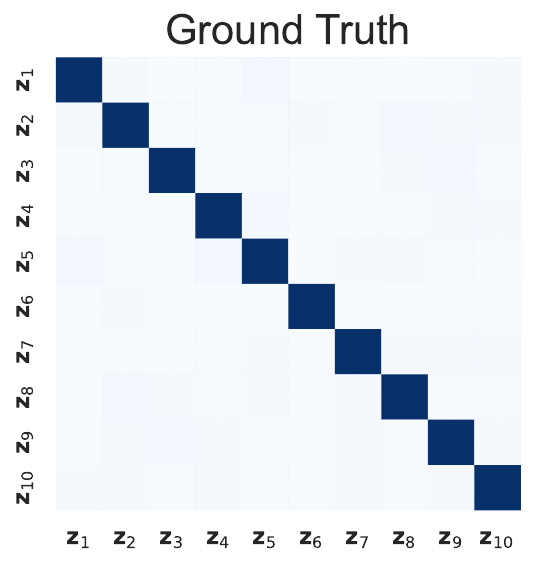}
        \hspace{0cm} 
        \includegraphics[width=0.18\textwidth]{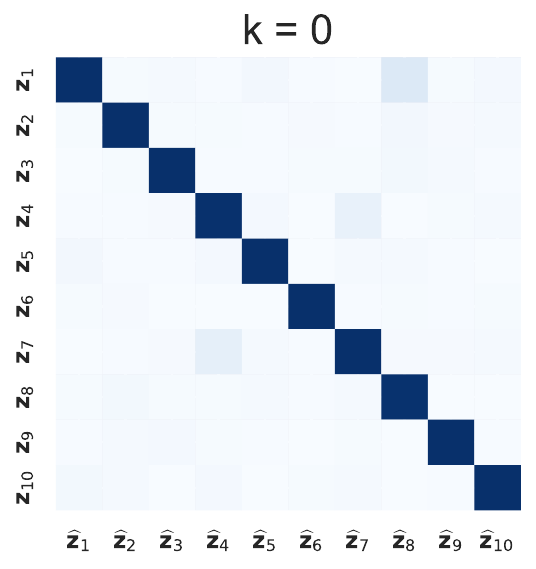}
        \hspace{0.5cm} 
         \includegraphics[width=0.18\textwidth]{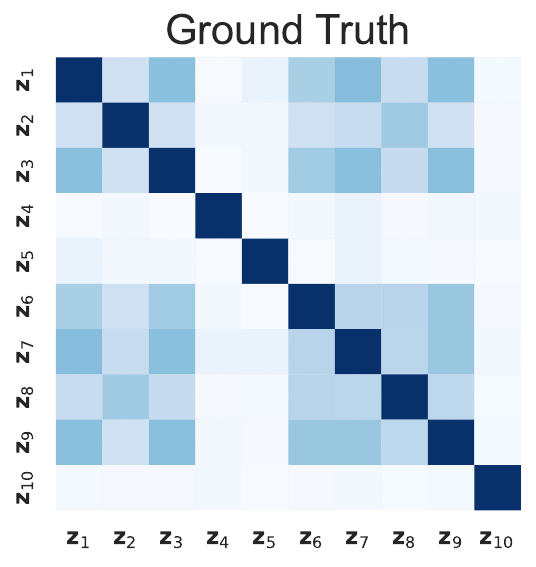}
        \hspace{0cm} 
        \includegraphics[width=0.18\textwidth]{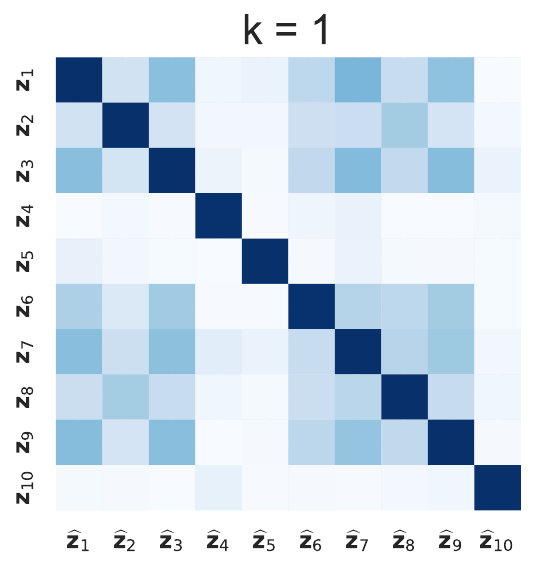}
        \includegraphics[width=0.0375\textwidth]{Figures/cbar.pdf}
        \hspace{2cm} 
      \includegraphics[width=0.18\textwidth]{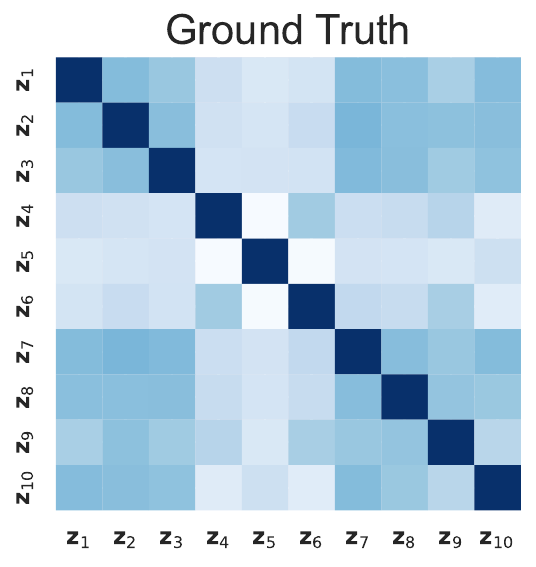}
        \hspace{0cm} 
        \includegraphics[width=0.18\textwidth]{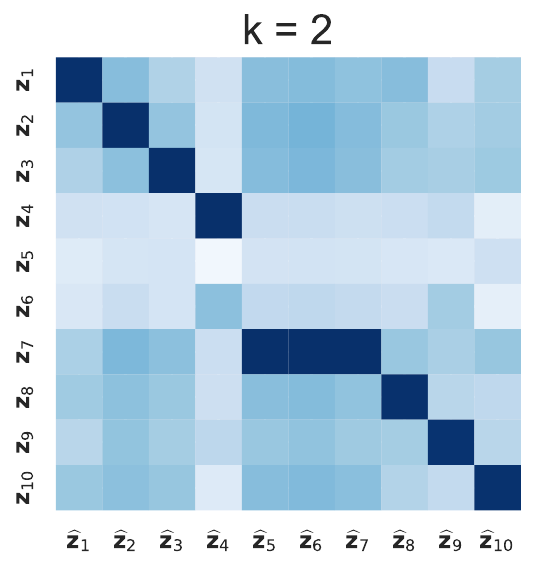}
        \hspace{0.5cm} 
        \includegraphics[width=0.18\textwidth]{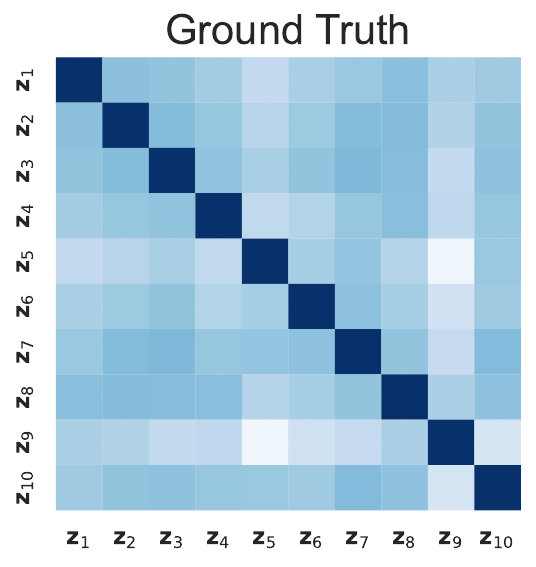}
        \hspace{0cm} 
        \includegraphics[width=0.18\textwidth]{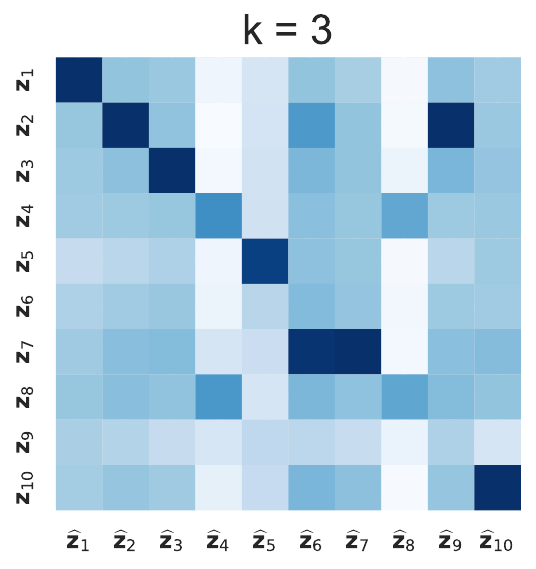}
        \includegraphics[width=0.0375\textwidth]{Figures/cbar.pdf}
        \caption{Linear mixing function with linear causal relation and Exponential noise: ablation study on increasing the density of causal graphs $k$ from 0 to 3 and fixing $n=10$, $\delta=0\sigma$, $\rho=50$. For all methods it is harder to learn the latent variables for denser graphs.}
        \label{fig:linear_exp_ablation_k}
    \end{center}
\end{figure*}

\begin{figure*}[!htbp]
    \begin{center}
         \includegraphics[width=0.18\textwidth]{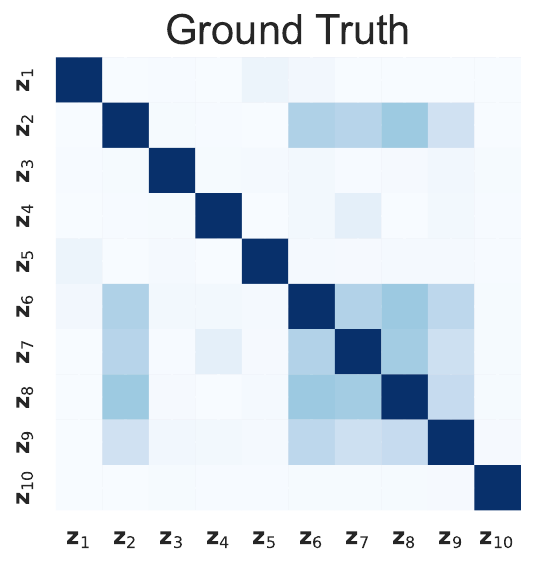}
        \hspace{0cm} 
        \includegraphics[width=0.18\textwidth]{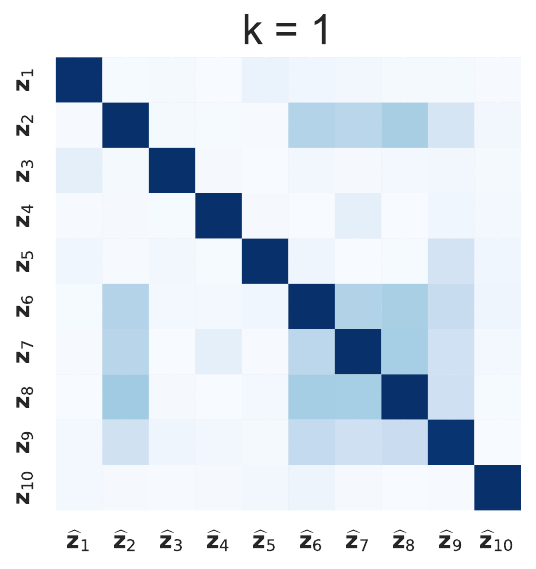}
        \hspace{0.5cm} 
      \includegraphics[width=0.18\textwidth]{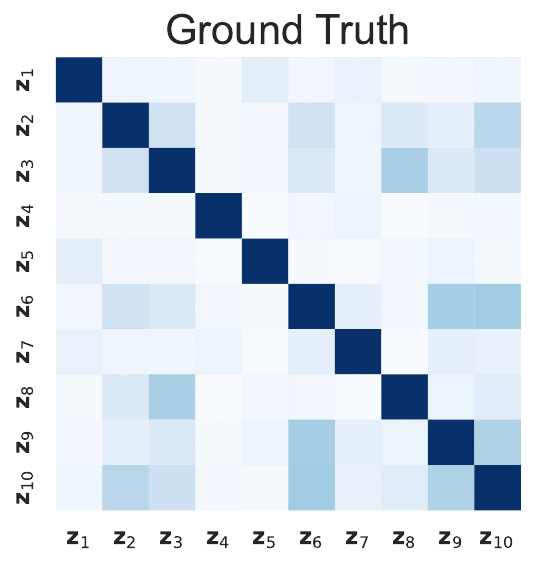}
        \hspace{0cm} 
        \includegraphics[width=0.18\textwidth]{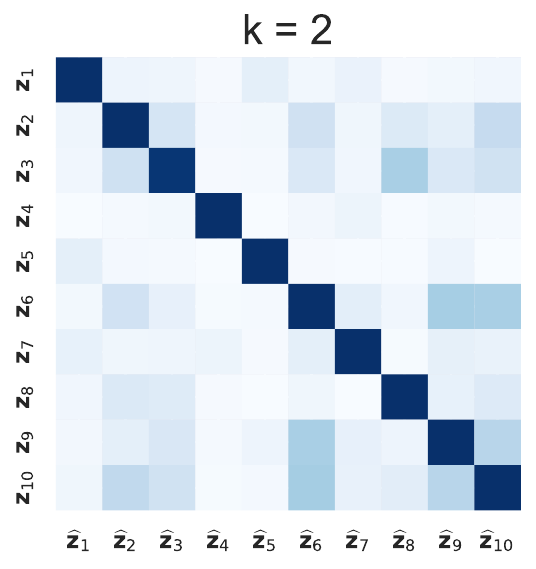}
        \includegraphics[width=0.0375\textwidth]{Figures/cbar.pdf}
        \hspace{2cm} 
        \includegraphics[width=0.18\textwidth]{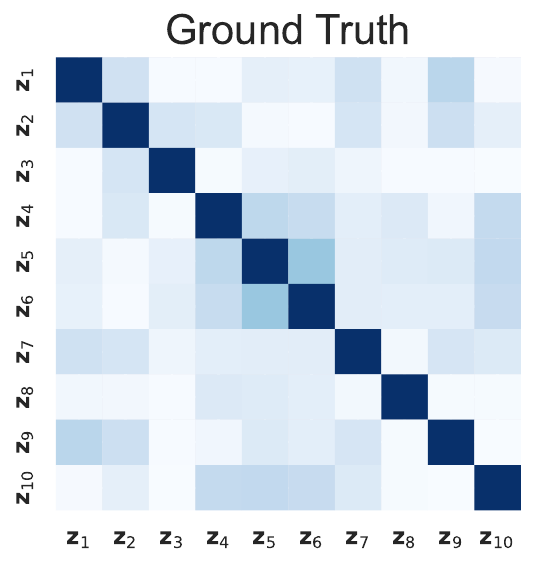}
        \hspace{0cm} 
        \includegraphics[width=0.18\textwidth]{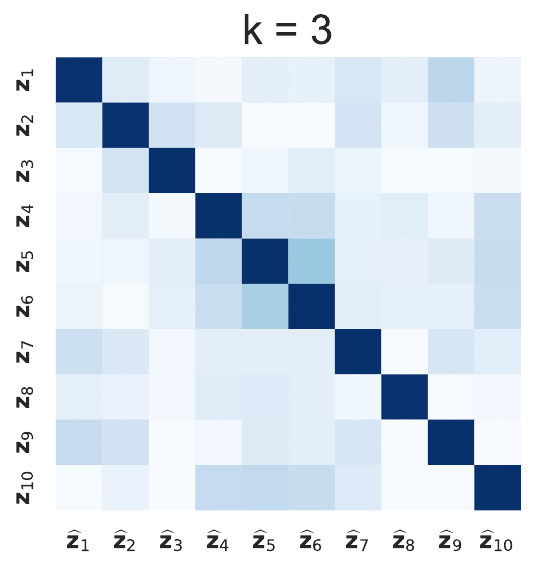}
        \includegraphics[width=0.0375\textwidth]{Figures/cbar.pdf}
        \caption{Linear mixing function with nonlinear causal relation and Gaussian noise: ablation study on increasing the density of causal graphs $k$ from 0 to 3 and fixing $n=10$, $\delta=0\sigma$, $\rho=50$.}
        \label{fig:linear_nonlinear_ablation_k}
    \end{center}
\end{figure*}
\FloatBarrier

To exclude the natural dependency between latent variables that potentially enhance the value of metric MCC, we provide additional experiment results with \emph{independent} latent variables using 20 random seeds in \cref{tab: num_linear_inde}.
\begin{table}[!htbp]
\caption{Results for the numerical experiments in the linear case when $\Zb$ are independent.} %
\label{tab: num_linear_inde}
\begin{center}
\begin{small}
\begin{sc}
\begin{tabular}{|c|c|c|c|}
\hline
$n$ & Causal Graph &  $\rho$ & \textbf{MCC}     \\ \hline
5    & Independent  & 50 \%  &  0.998$\pm$0.002\\
10   &   Independent &  50 \%  & 0.998$\pm$0.001\\
20   &  Independent  & 50 \% & 0.997$\pm$0.002\\
40   &   Independent   & 50 \%  & 0.748$\pm$0.145\\
\hline
10    & Independent  & 1var  &  0.998 $\pm$0.001 \\
10   &   Independent & 50 \%  & 0.998$\pm$0.001\\
10   &  Independent  & 75 \%  & 0.998 $\pm$0.002\\
\hline
\end{tabular}
\end{sc}
\end{small}
\end{center}
\end{table}

\FloatBarrier

\subsubsection{Piecewise linear mixing function}
In \cref{tab: num_piecewise_lowdim} and \cref{tab: num_piecewise}, we provide the numerical results of averaged MCC and standard deviation over 20 random seeds as shown in \cref{fig:num_pw}. In the second block of \cref{tab: num_piecewise_lowdim}, we provide two additional results of our method (training without mask information) when $\Zb$ are independent. In \cref{tab: num_piecewise}, the second right column is for training without mask information, and the most right column is the oracle method, i.e., providing mask information in the training phase. Additionally, to exclude the natural dependency between latent variables that potentially enhance the value of metric MCC, we provide additional experiment results with \emph{independent} latent variables in \cref{tab: num_piecewise_inde}.

\begin{table}[!htbp]
\caption{Results for the numerical experiments in the piecewise linear case for simple settings.}%
\label{tab: num_piecewise_lowdim}
\begin{center}
\begin{small}
\begin{sc}
\begin{tabular}{|c|c|c|c|c|c|}
\hline
$n$ & $m$ & $k$ &  $\rho$ & $\delta$ &\textbf{MCC}  (mean $\pm$ std)          \\ \hline
3   &  3 & 0  & 50 \% & 0 &0.630$\pm$0.064\\
3   &   3  & 0 & 50 \%  & 2 &  0.755$\pm$0.068\\
3   &  3  & 0  & 50 \% & 3 &0.769$\pm$0.075\\
3   &   3  & 0 & 50 \% & 5 &  0.813$\pm$0.070\\
3   &   3  & 0 & 50 \% & 10 &  0.872$\pm$0.071\\
\hline
3   &  3 & 1  & 50 \% & 0 &0.649$\pm$0.059\\
3   &   3  & 1 & 50 \%  & 2 &  0.756$\pm$0.065\\
3   &  3  & 1  & 50 \% & 3 &0.783$\pm$0.082\\
3   &   3  & 1 & 50 \% & 5 &  0.836$\pm$0.091\\
3   &   3  & 1 & 50 \% & 10 &  0.877$\pm$0.074\\
\hline
\end{tabular}
\end{sc}
\end{small}
\end{center}
\end{table}

\begin{table*}[!htbp]
\caption{Results for the numerical experiments in the piecewise linear case for the \emph{dependent} variable case.}

\label{tab: num_piecewise}
\begin{center}
\begin{small}
\begin{sc}
\begin{tabular}{|c|c|c|c|c|c|c|}
\hline
$n$ & $m$ & $k$ &  $\rho$ & $\delta$ &\textbf{MCC}  (mean $\pm$ std) & \textbf{MCC}  (mean $\pm$ std) oracle       \\ \hline
5    &  10  & 1  & 50 \%  & 2& 0.469$\pm$0.050 & 0.898$\pm$0.018 \\
10   &   10  & 1 & 50 \%  & 2 &  0.349$\pm$0.030 & 0.799$\pm$0.020\\
20   &  10  & 1  & 50 \% & 2 &0.280$\pm$0.017 & 0.753$\pm$0.015\\
40   &   10   & 1 & 50 \%  & 2 &0.217$\pm$0.025 & 0.782$\pm$0.012\\
\hline
10   &  10  & 1 & 50 \%  & 2&0.349$\pm$0.030 & 0.799$\pm$0.020\\
10   &  10  & 2 & 50 \%  & 2 &0.337$\pm$0.030 &0.789$\pm$0.021\\
10   &  10   & 3  & 50 \% & 2&0.345$\pm$0.028  &0.785$\pm$0.022
\\ \hline
10   &  3   & 1 & 50 \%  &2 & 0.466$\pm$0.038 & 0.867$\pm$0.007\\
10   &  10   & 1  & 50 \%  & 2 &0.349$\pm$0.030 &0.799$\pm$0.020\\
10   &  20   & 1  & 50 \% &2 & 0.267$\pm$0.031 &0.552$\pm$0.035\\
\hline 
10   &  10   & 1 & 1\textsc{var}  &2 & 0.402$\pm$0.027 &0.908$\pm$0.013\\
10   &  10   & 1 & 50 \%  &2 & 0.349$\pm$0.030  &0.799$\pm$0.020\\
10   &  10   & 1 & 75 \% & 2 &0.352$\pm$0.037 &0.695$\pm$0.026\\
\hline
10   &  10   & 1 & 50 \% &0 & 0.332$\pm$0.040  &0.185$\pm$0.036\\
10   &  10   & 1 & 50 \% &3 & 0.368$\pm$0.028  &0.856$\pm$0.015\\
10   &  10   & 1 & 50 \% & 5 &0.388$\pm$0.029  &0.930$\pm$0.010\\
10   &  10   & 1 & 50 \%  & 10 &0.405$\pm$0.027  &0.980$\pm$0.007\\
\hline
\end{tabular}
\end{sc}
\end{small}
\end{center}
\end{table*}
\FloatBarrier

\begin{table*}[!htbp]
\caption{Results for the numerical experiments in the piecewise linear case for the \emph{independent} variable case.}%
\label{tab: num_piecewise_inde}
\begin{center}
\begin{small}
\begin{sc}
\begin{tabular}{|c|c|c|c|c|c|c|}
\hline
$n$ & $m$ & $k$ &  $\rho$ & $\delta$ &\textbf{MCC}  (mean $\pm$ std) & \textbf{MCC}  (mean $\pm$ std) oracle     \\ \hline
5    &  10  & 0  & 50 \% & 2& 0.462$\pm$0.065 & 0.910$\pm$0.014 \\
10   &   10  & 0 & 50 \%  & 2 &  0.352$\pm$0.028 & 0.807$\pm$0.022\\
20   &  10  & 0  & 50 \% & 2&0.296$\pm$0.018 & 0.732$\pm$0.011\\
40   &   10   & 0 & 50 \%  & 2 &0.246$\pm$0.013 & 0.746$\pm$0.015\\
\hline
10   &  3   & 0 & 50 \%  &2 & 0.523$\pm$0.026&0.865$\pm$0.008\\
10   &  10   & 0  & 50 \% & 2 &0.352$\pm$0.028 & 0.807$\pm$0.022\\
10   &  20   & 0  & 50 \% &2 & 0.269$\pm$0.023 &0.519$\pm$0.028\\
\hline 
10   &  10   & 0 & 1\textsc{var}  &2 & 0.412$\pm$0.027 &0.906$\pm$0.016\\
10   &  10   & 0 & 50 \%  &2 & 0.352$\pm$0.028  &0.807$\pm$0.022\\
10   &  10   & 0 & 75 \% & 2 &0.340$\pm$0.028 &0.719$\pm$0.017\\
\hline
10   &  10   & 0 & 50 \%  &0  & 0.318$\pm$0.026  &0.236$\pm$0.040\\
10   &  10   & 0 & 50 \%  &3 & 0.366$\pm$0.029  &0.852$\pm$0.020\\
10   &  10   & 0 & 50 \% & 5 &0.388$\pm$0.027  &0.926$\pm$0.014\\
\hline
\end{tabular}
\end{sc}
\end{small}
\end{center}
\end{table*}
\FloatBarrier

For each setup in \cref{tab: num_piecewise}, we choose one random seed to plot the heatmap of the Pearson Correlation matrix $\mathrm{Corr}^{n\times n}_{\pi}$ with the permutation $\pi$ . One figure represents one ablation study of one parameter. Ground truth Pear. Corr. matrix on the left shows the original linear correlation inside Z, compared with the estimator on the right-hand side.

\begin{figure*}
    \begin{center}
        \includegraphics[width=0.18\textwidth]{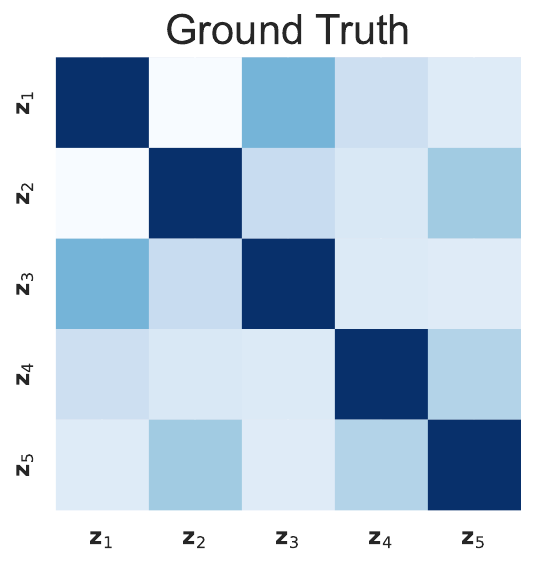}
        \hspace{0cm} 
        \includegraphics[width=0.18\textwidth]{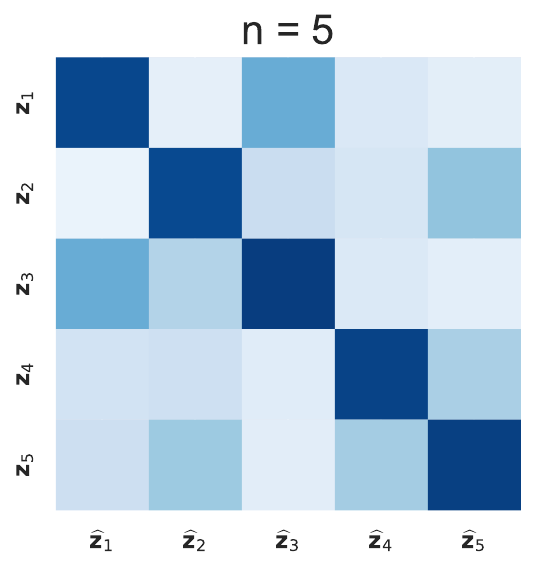}
        \hspace{0.5cm} 
         \includegraphics[width=0.18\textwidth]{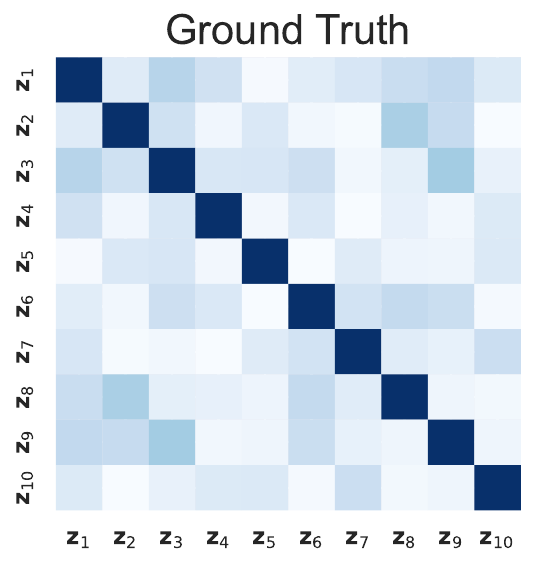}
        \hspace{0cm} 
        \includegraphics[width=0.18\textwidth]{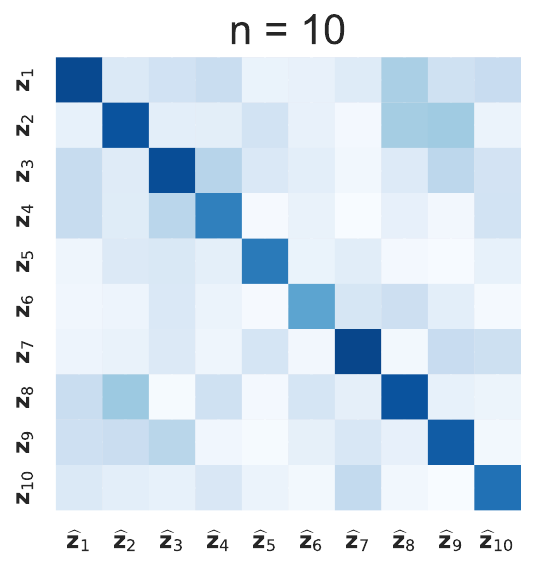}
        \includegraphics[width=0.0375\textwidth]{Figures/cbar.pdf}
        \hspace{0.5cm} 
        \includegraphics[width=0.18\textwidth]{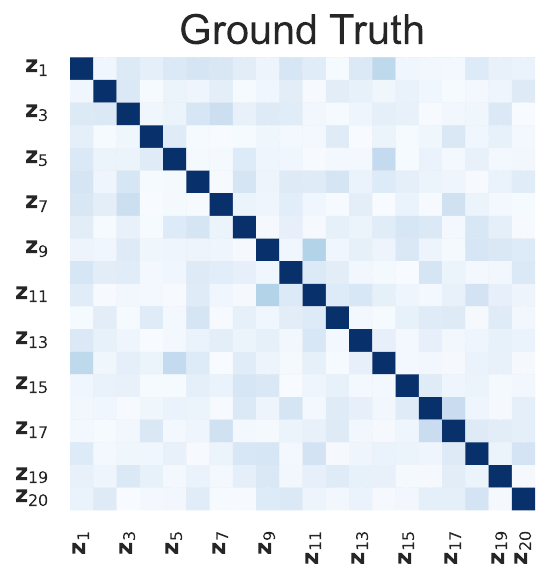}
        \hspace{0cm} 
        \includegraphics[width=0.18\textwidth]{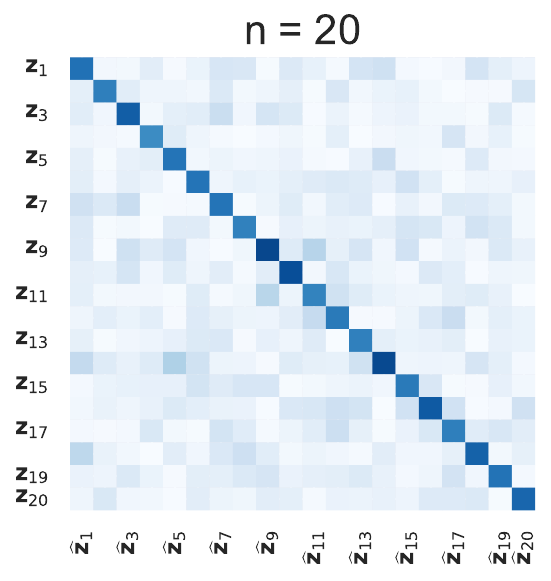}
        \hspace{0.5cm} 
        \includegraphics[width=0.18\textwidth]{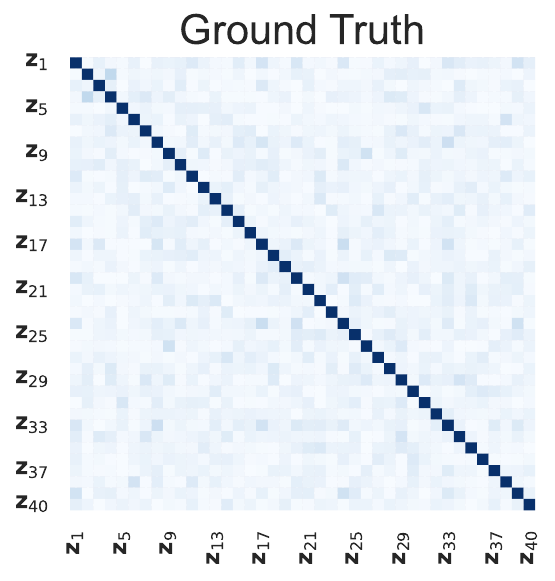}
        \hspace{0cm} 
        \includegraphics[width=0.18\textwidth]{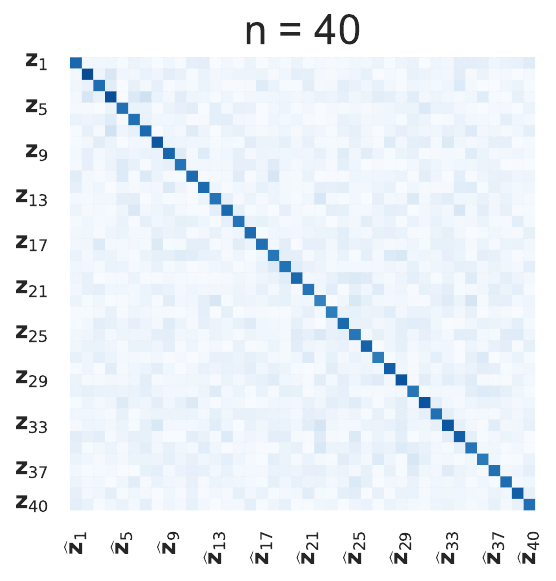}
        \includegraphics[width=0.0375\textwidth]{Figures/cbar.pdf}
        \caption{Piecewise linear mixing function with linear causal relation and Gaussian noise: ablation study on increasing the latent dimension $n$ from 5 to 40 and fixing $\delta=2.0\sigma$, $m=10$, $\rho=50$, $k=1$. The case with higher $n$ is more complicated to learn the latent variables.}
        \label{fig:pw_ablation_n}
    \end{center}
\end{figure*}

\begin{figure*}
    \begin{center}
        \includegraphics[width=0.18\textwidth]{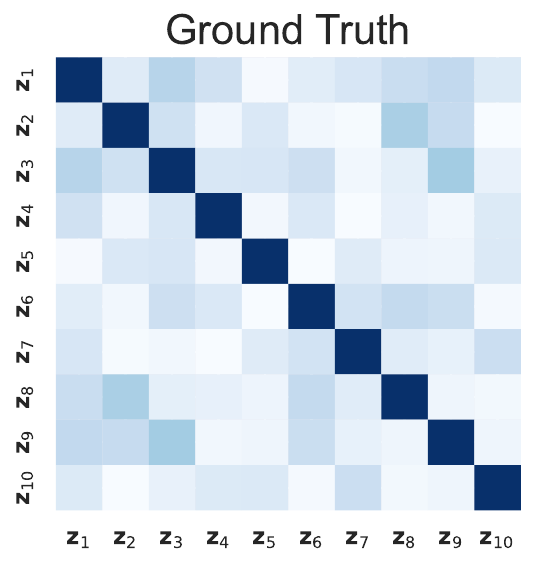}
        \hspace{0cm} 
        \includegraphics[width=0.18\textwidth]{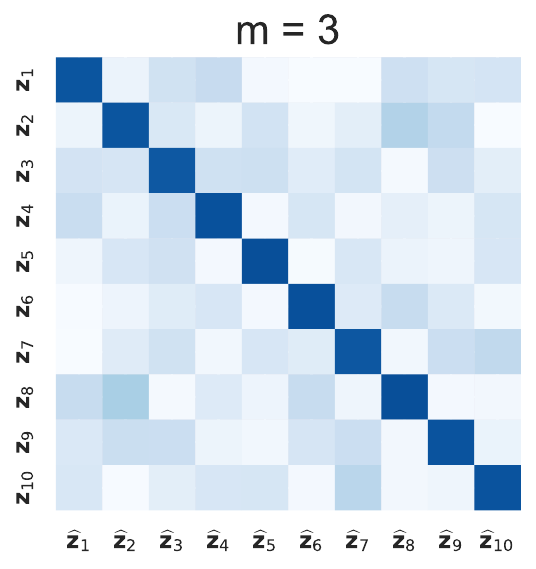}
        \hspace{0.5cm} 
         \includegraphics[width=0.18\textwidth]{Figures/heatmaps_pw/true/pc10_d2.0_n10_nn10_M50_G1_rs2_Corr_heatmap.pdf}
        \hspace{0cm} 
        \includegraphics[width=0.18\textwidth]{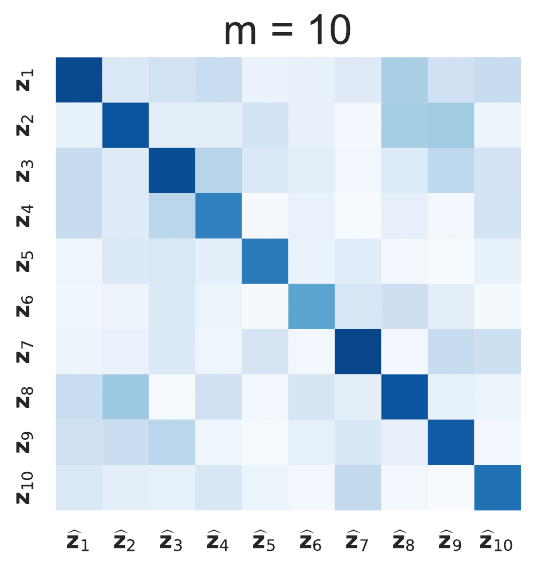}
        \includegraphics[width=0.0375\textwidth]{Figures/cbar.pdf}
        \hspace{0.5cm} 
        \includegraphics[width=0.18\textwidth]{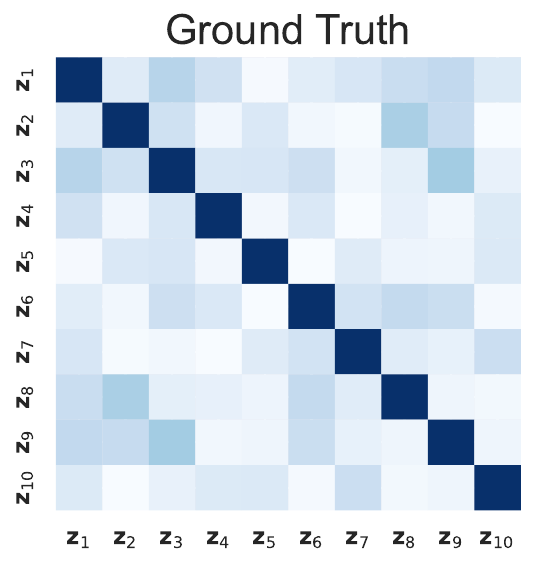}
        \hspace{0cm} 
        \includegraphics[width=0.18\textwidth]{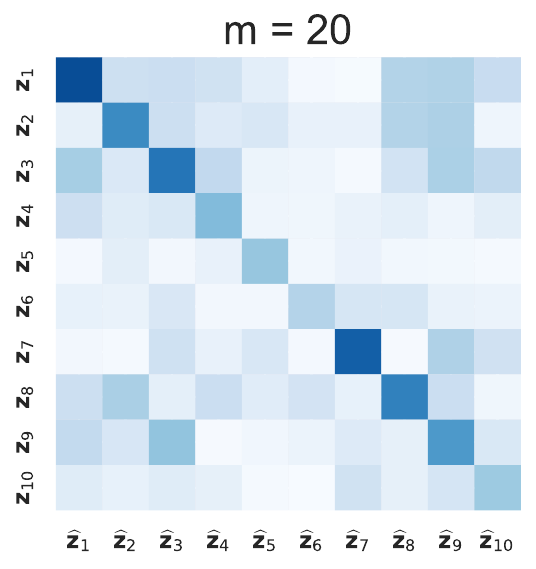}
        \includegraphics[width=0.0375\textwidth]{Figures/cbar.pdf}
        \caption{Piecewise linear mixing function with linear causal relation and Gaussian noise: ablation study on increasing the number of Leaky-ReLU layers $(m-1)$ from 3 to 20 and fixing $n=10$, $\delta=2.0\sigma$, $\rho=50$, $k=1$. The case with larger $m$ is more complicated to learn the latent variables due to a greater extent of nonlinearity.}
        \label{fig:pw_ablation_m}
    \end{center}
\end{figure*}

\begin{figure*}
    \begin{center}
        \includegraphics[width=0.18\textwidth]{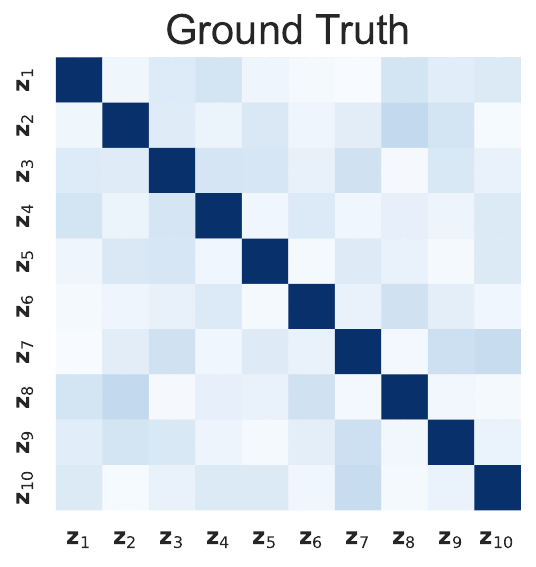}
        \hspace{0cm} 
        \includegraphics[width=0.18\textwidth]{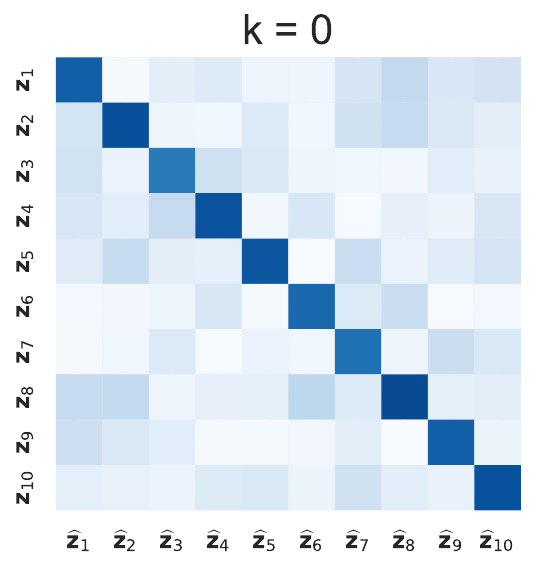}
        \hspace{0.5cm} 
         \includegraphics[width=0.18\textwidth]{Figures/heatmaps_pw/true/pc10_d2.0_n10_nn10_M50_G1_rs2_Corr_heatmap.pdf}
        \hspace{0cm} 
        \includegraphics[width=0.18\textwidth]{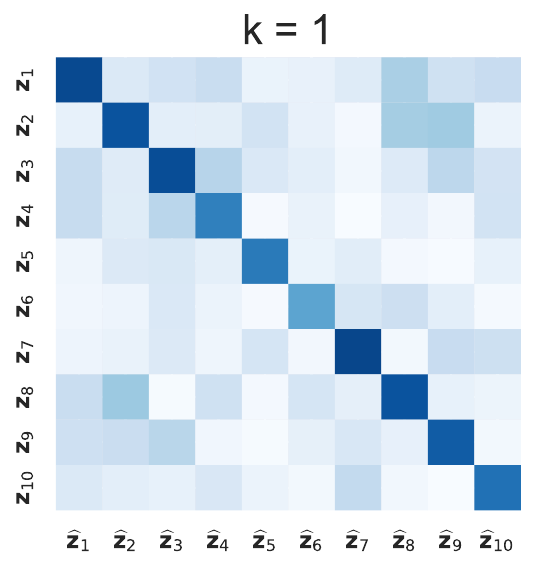}
        \includegraphics[width=0.0375\textwidth]{Figures/cbar.pdf}
        \hspace{0.5cm} 
        \includegraphics[width=0.18\textwidth]{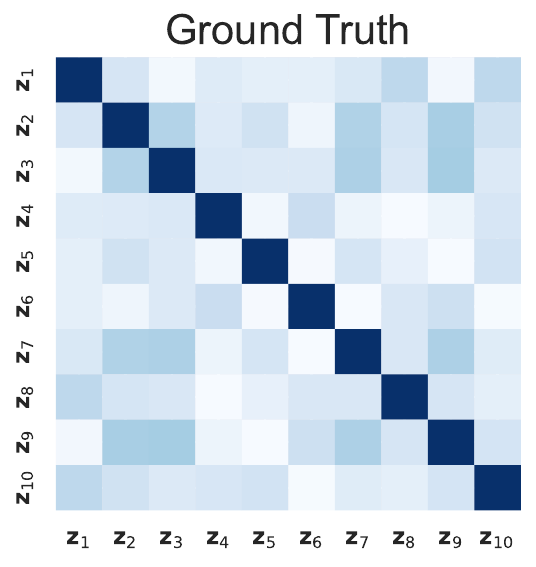}
        \hspace{0cm} 
        \includegraphics[width=0.18\textwidth]{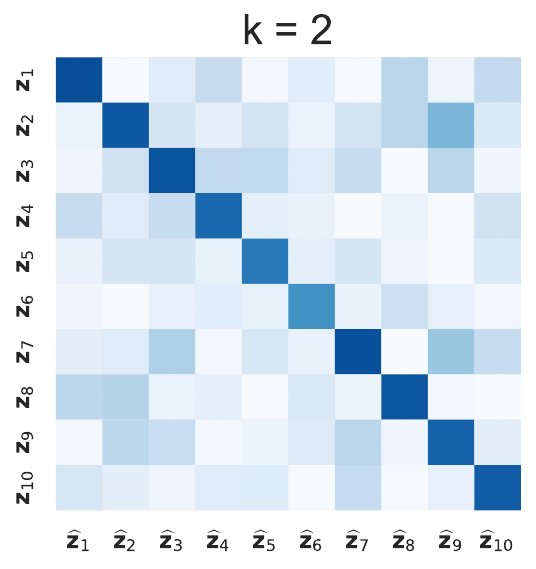}
        \hspace{0.5cm} 
        \includegraphics[width=0.18\textwidth]{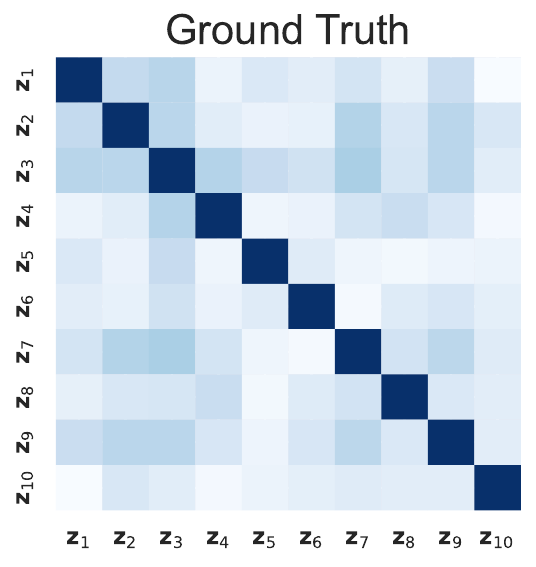}
        \hspace{0cm} 
        \includegraphics[width=0.18\textwidth]{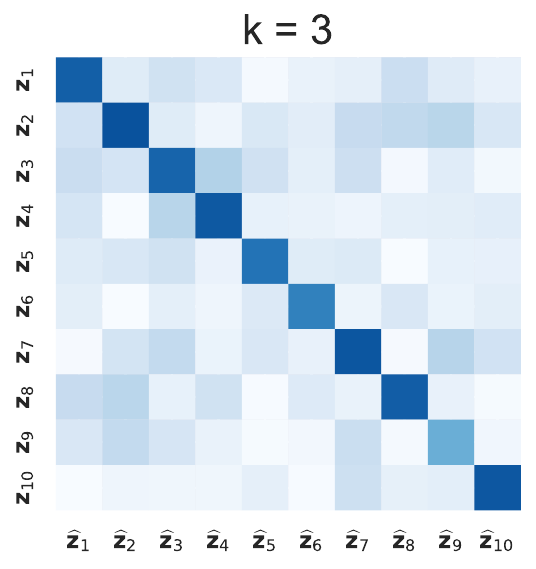}
        \includegraphics[width=0.0375\textwidth]{Figures/cbar.pdf}
        \caption{Piecewise linear mixing function with nonlinear causal relation and Gaussian noise: ablation study on increasing the density of causal graphs $k$ from 0 to 3 and fixing $n=10$, $m=10$, $\delta=2.0\sigma$, $\rho=50$. In the case with a denser graph, it is more complicated to learn the latent variables.}
        \label{fig:pw_ablation_k}
    \end{center}
\end{figure*}

\begin{figure*}
    \begin{center}
        \includegraphics[width=0.18\textwidth]{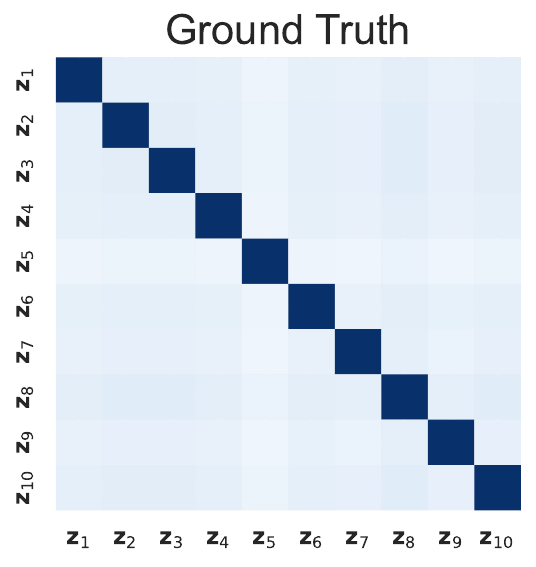}
        \hspace{0cm} 
        \includegraphics[width=0.18\textwidth]{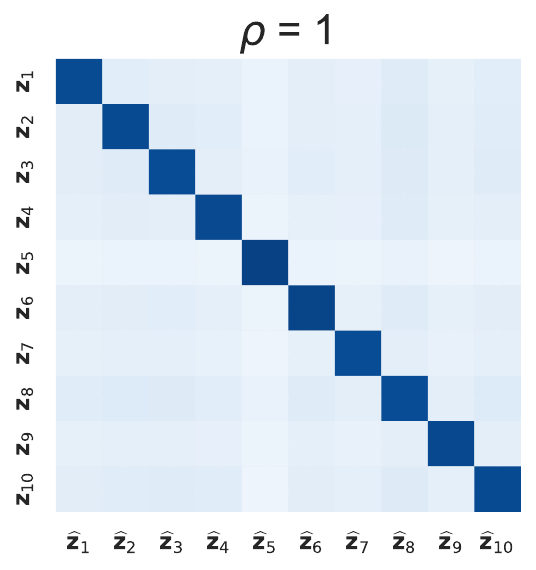}
        \hspace{0.5cm} 
         \includegraphics[width=0.18\textwidth]{Figures/heatmaps_pw/true/pc10_d2.0_n10_nn10_M50_G1_rs2_Corr_heatmap.pdf}
        \hspace{0cm} 
        \includegraphics[width=0.18\textwidth]{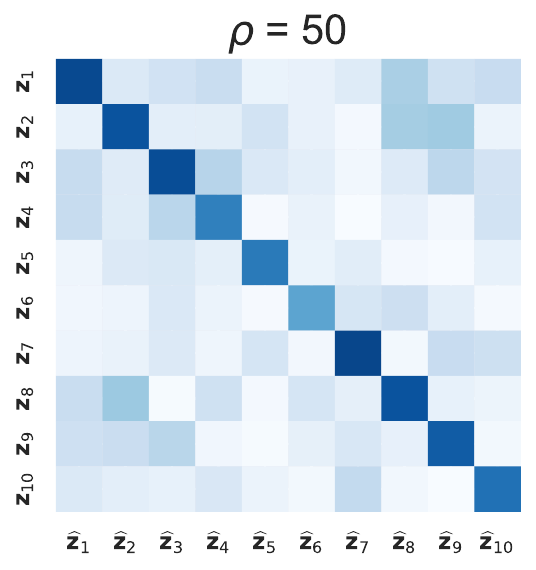}
        \includegraphics[width=0.0375\textwidth]{Figures/cbar.pdf}
        \hspace{0.5cm} 
        \includegraphics[width=0.18\textwidth]{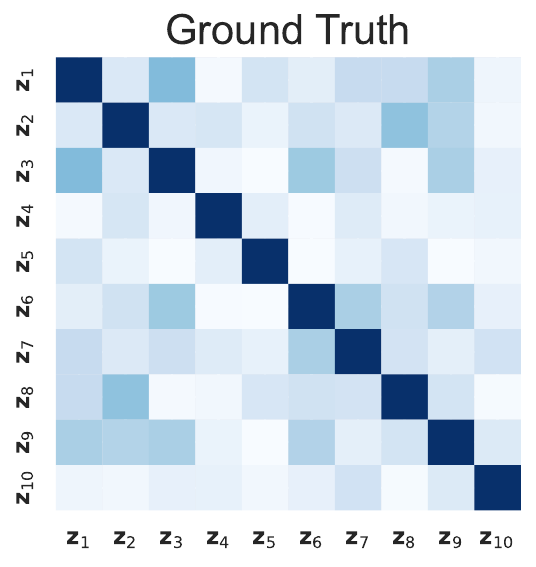}
        \hspace{0cm} 
        \includegraphics[width=0.18\textwidth]{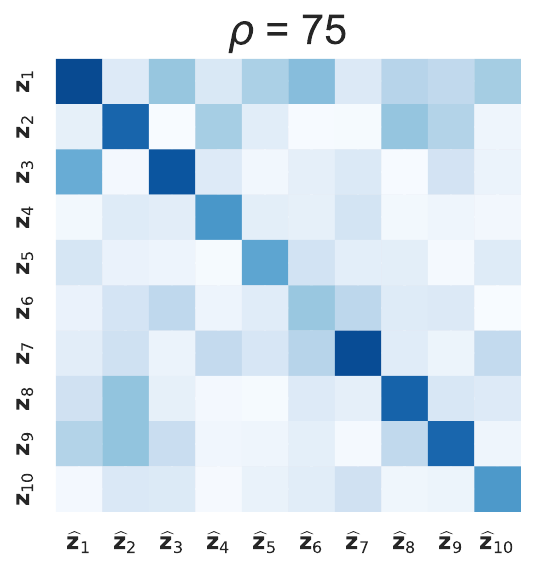}
        \includegraphics[width=0.0375\textwidth]{Figures/cbar.pdf}
        \caption{Piecewise linear mixing function with linear causal relation and Gaussian noise: ablation study on increasing the ratio of active (unmasked) variables $\rho$ from 1 variable only to 75\% and fixing $n=10$, $m=10$, $\delta=2.0\sigma$, $k=1$. Learning the latent variables is more complicated in the case with a larger portion of active variables.}
        \label{fig:pw_ablation_rho}
    \end{center}
\end{figure*}

\begin{figure*}
    \begin{center}
        \includegraphics[width=0.18\textwidth]{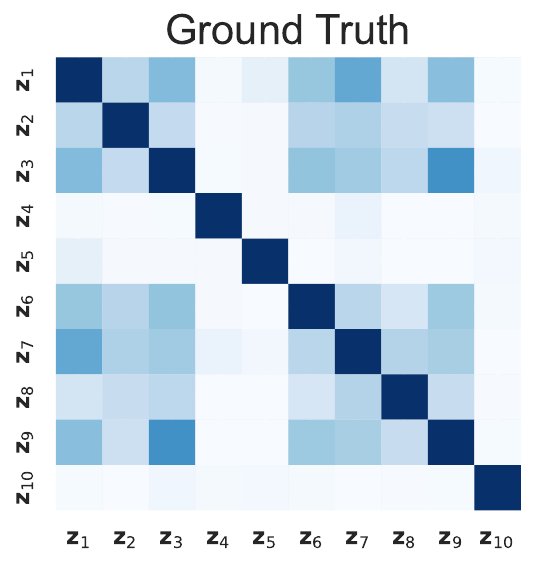}
        \hspace{0cm} 
        \includegraphics[width=0.18\textwidth]{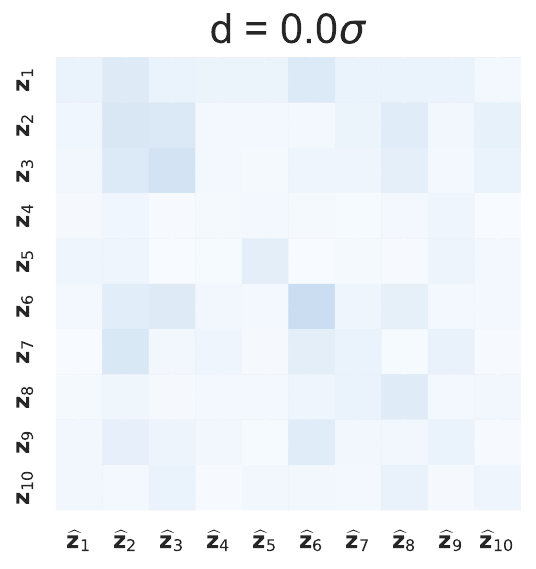}
        \hspace{0.5cm} 
         \includegraphics[width=0.18\textwidth]{Figures/heatmaps_pw/true/pc10_d2.0_n10_nn10_M50_G1_rs2_Corr_heatmap.pdf}
        \hspace{0cm} 
        \includegraphics[width=0.18\textwidth]{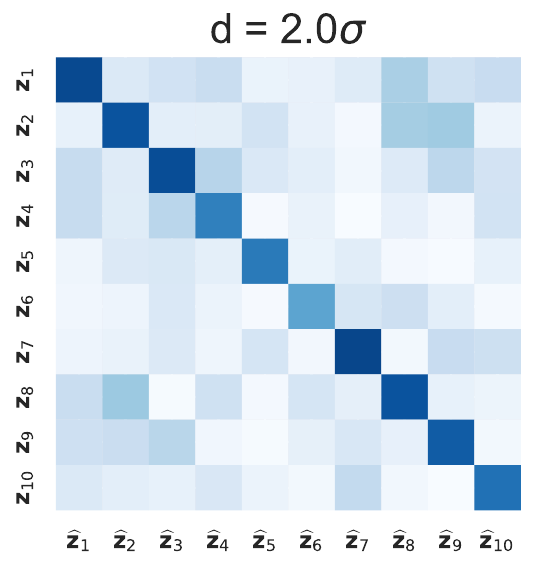}
        \includegraphics[width=0.0375\textwidth]{Figures/cbar.pdf}
        \hspace{0.5cm} 
         \includegraphics[width=0.18\textwidth]{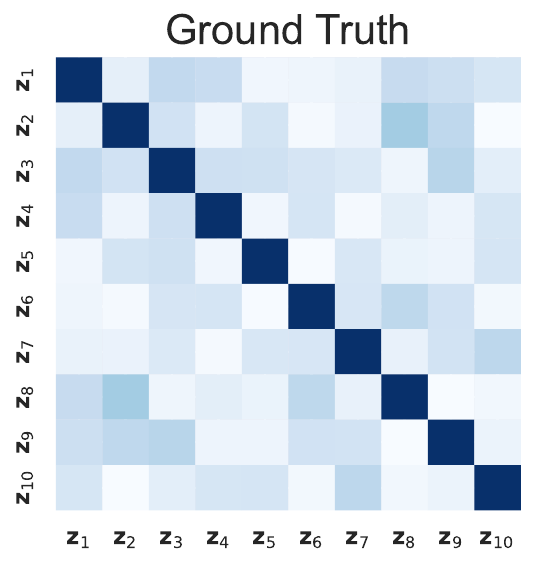}
        \hspace{0cm} 
        \includegraphics[width=0.18\textwidth]{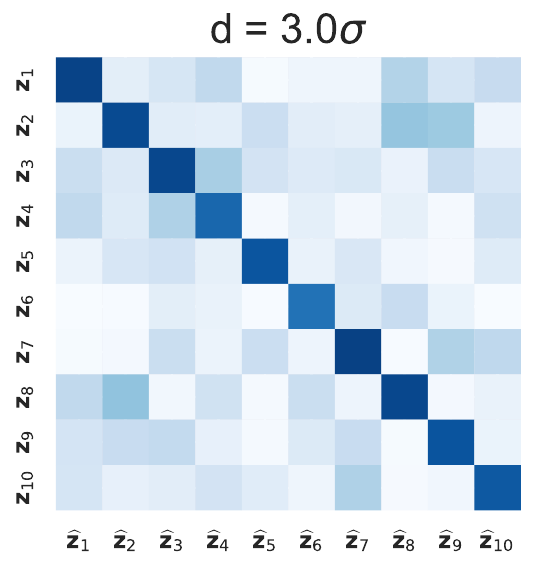}
        \hspace{0.5cm} 
         \includegraphics[width=0.18\textwidth]{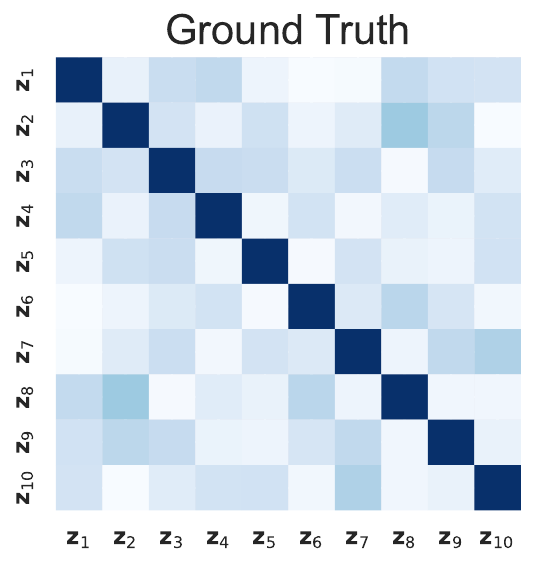}
        \hspace{0cm} 
        \includegraphics[width=0.18\textwidth]{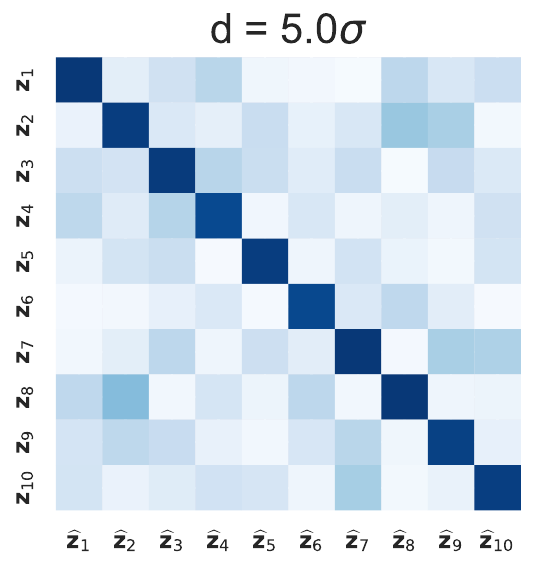}
        \includegraphics[width=0.0375\textwidth]{Figures/cbar.pdf}
        \hspace{0.5cm}
         \includegraphics[width=0.18\textwidth]{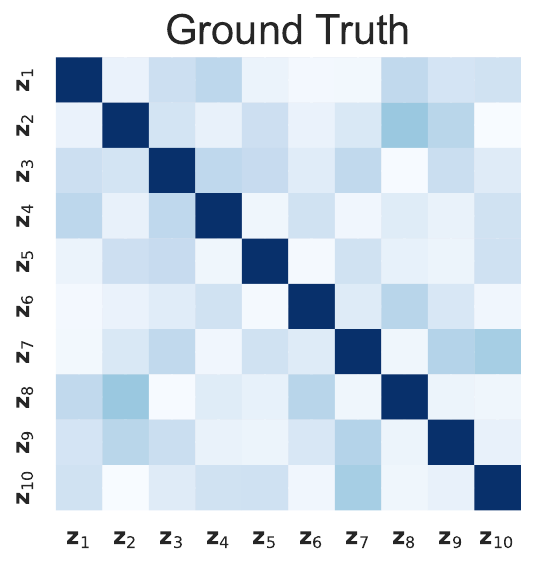}
        \hspace{0cm} 
        \includegraphics[width=0.18\textwidth]{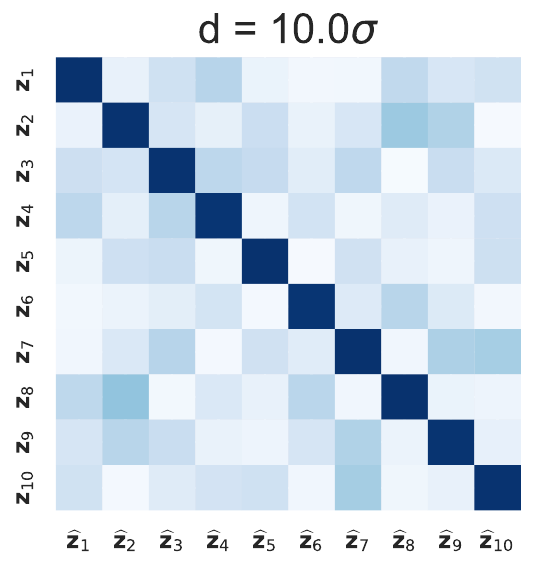}
        \includegraphics[width=0.0375\textwidth]{Figures/cbar.pdf}
        \caption{Piecewise linear mixing function with linear causal relation and Gaussian noise: ablation study on increasing the distance between masked value and mean of latents from $0.0 \sigma$ to $10.0 \sigma$, where $\sigma$ is the standard deviation of latents, and fixing $n=10$, $m=10$, $\rho=50$, $k=1$. Learning the latent variables is more complicated in the case with a smaller distance.}
        \label{fig:pw_ablation_delta}
    \end{center}
\end{figure*}
\FloatBarrier

\subsubsection{Test on independent latent variables}
\label{app: test independent z}
To exclude the effect of the inherent causal relation, which might increase the value of metric MCC, we additionally test our estimated encoder $\gb$ when the unmasked causal variables are independent with each other, and implement the same ablation study as \cref{tab: num_linear} for linear mixing function and \cref{tab: num_piecewise} for piecewise linear mixing function. As we can see, in \cref{tab: num_linear_test_on_indep}, comparing the right two columns, there is no significant difference, which provides evidence that the MCC obtained by our method does not come from the inherent causal relation among latents. 

\begin{table}[!htbp]
\caption{Results of testing on \emph{independent} latents for linear mixing functions with $\delta=0$.
The bold font indicates which parameters are varying in each block of rows. 
}
\label{tab: num_linear_test_on_indep}
\begin{center}
\begin{small}
\begin{sc}
\begin{tabular}{cccccc}
\hline
$n$ & $k$ & SCM &  $\rho$ & MCC (\cref{tab: num_linear}) & MCC (Test on Independent latents)     \\ \hline
\textbf{5}    & 1 &Lin. Gauss  & 50 \%  &  0.997$\pm$0.002 &  0.991$\pm$0.010 \\
\textbf{10}   & 1 &Lin. Gauss &  50 \%   & 0.996$\pm$0.001 & 0.992$\pm$0.004\\
\textbf{20}   &  1 & Lin. Gauss  & 50 \%  & 0.987$\pm$0.029 & 0.985$\pm$0.031\\
\textbf{40}   & 1 &Lin. Gauss   & 50 \%   & 0.714$\pm$0.153 & 0.718$\pm$0.142\\
\hline
10    & \textbf{0} & Indep. Gauss  & 50 \%  &  0.998$\pm$0.001 & 0.998$\pm$0.010\\
10    & \textbf{1} &Lin. Gauss  & 50 \%   &  0.996$\pm$0.001 & 0.992$\pm$0.004\\
10   &  \textbf{2} &Lin. Gauss & 50 \%   &  0.904$\pm$0.113 & 0.893$\pm$0.122\\
10   &  \textbf{3} &Lin. Gauss  & 50 \%   & 0.793$\pm$0.142 & 0.744$\pm$0.179\\
\hline
10    & \textbf{0} & Indep. Exp  & 50 \%   &  0.998$\pm$0.001 & 0.998$\pm$0.001\\
10   &  \textbf{1} & Lin. Exp   & 50 \%   & 0.998$\pm$0.002 & 0.996$\pm$0.002\\
10    & \textbf{2} & Lin. Exp  & 50 \%  &   0.910$\pm$0.108 & 0.895$\pm$0.123\\
10   & \textbf{3} &Lin. Exp & 50 \%   & 0.825$\pm$0.123 & 0.776$\pm$0.156 \\
\hline
10   &  \textbf{1} & Nonlinear  & 50 \%  &  0.997$\pm$0.001 & 0.994$\pm$0.005\\
10   &  \textbf{2} & Nonlinear  & 50 \%  & 0.997$\pm$0.001 & 0.996$\pm$0.003\\
10   &  \textbf{3} & Nonlinear  & 50 \%  & 0.996$\pm$0.001 & 0.994$\pm$0.004\\
\hline
10    & 1 &Lin. Gauss  & \textbf{1\text{var}}  &  0.998 $\pm$0.002 & 0.993$\pm$0.005 \\
10   &  1 &Lin. Gauss & \textbf{50 \%}   & 0.996$\pm$0.001 & 0.992$\pm$0.004 \\
10   &  1 &Lin. Gauss  & \textbf{75 \%}  & 0.877 $\pm$0.096 & 0.925$\pm$0.068\\
\hline
\end{tabular}
\end{sc}
\end{small}
\end{center}
\end{table}
\FloatBarrier

\begin{figure*}[!htbp]
    \begin{center}
        \includegraphics[width=0.18\textwidth]{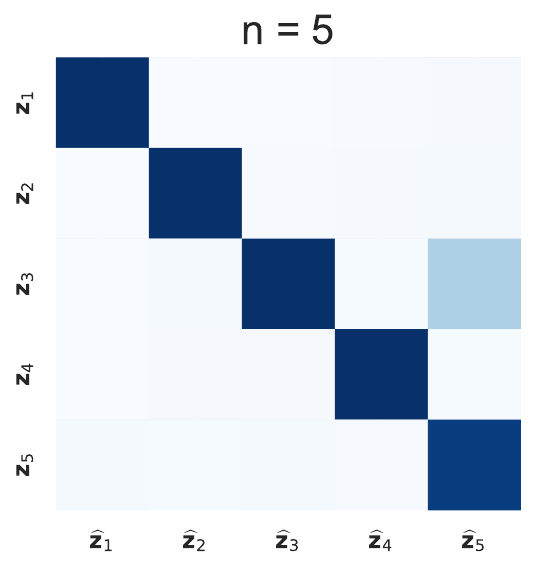}
        \hspace{0.2cm} 
        \includegraphics[width=0.18\textwidth]{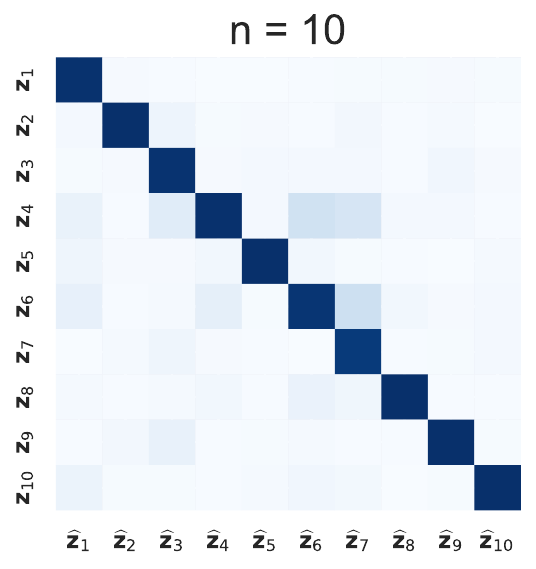}
        \hspace{0.2cm} 
        \includegraphics[width=0.18\textwidth]{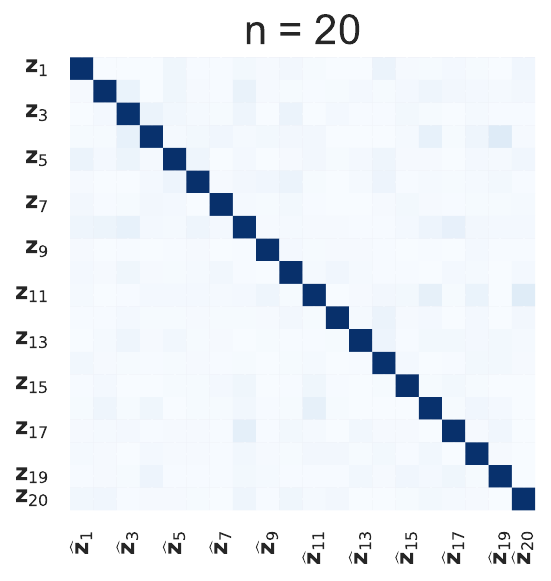}
        \hspace{0.2cm} 
        \includegraphics[width=0.18\textwidth]{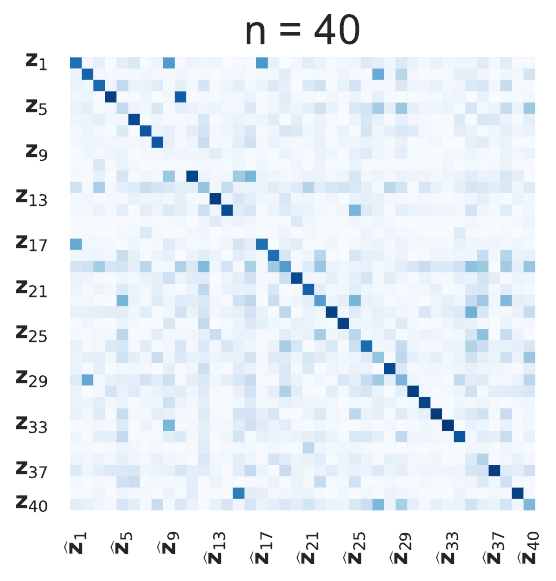}
        \hspace{0cm} 
        \includegraphics[width=0.0375\textwidth]{Figures/cbar.pdf}
         
        \label{fig:linear_gauss_ablation_n_indep}
        \caption{Linear mixing function with linear causal relation and Gaussian noise: ablation study on increasing the latent dimension $n$ from 5 to 40 and fixing $\delta=0\sigma$, $\rho=50$, $k=1$. The case with higher $n$ is more complicated to learn the latent variables.}
        
    \end{center}
\end{figure*}

\begin{figure*}[!htbp]
    \begin{center}
        \includegraphics[width=0.18\textwidth]{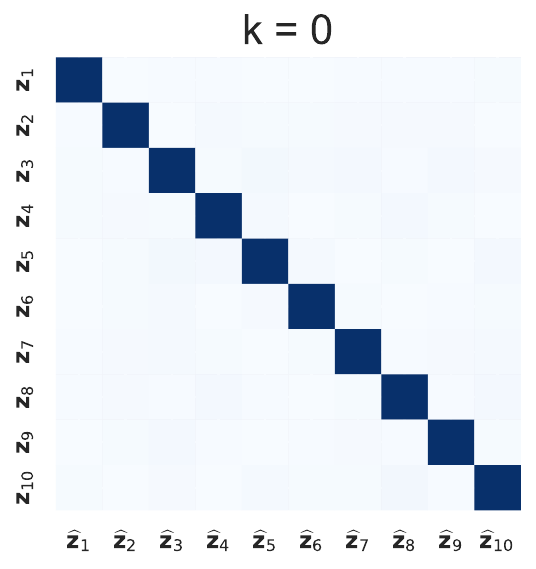}
        \hspace{0.2cm} 
        \includegraphics[width=0.18\textwidth]{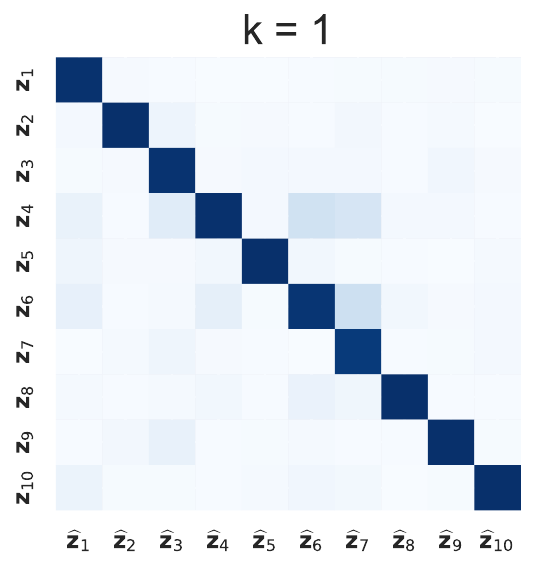}
        \hspace{0.2cm} 
        \includegraphics[width=0.18\textwidth]{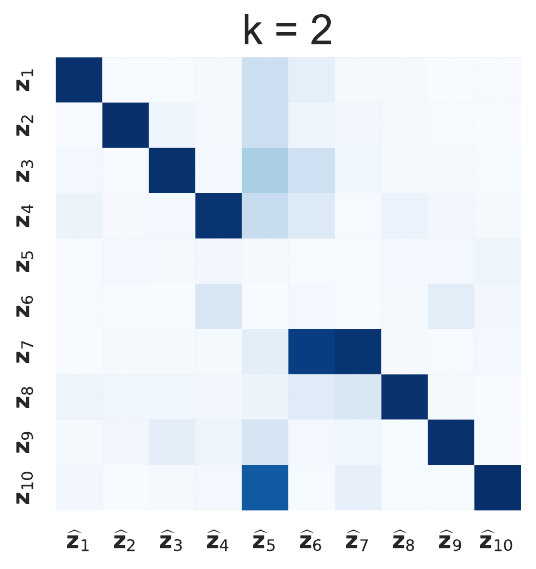}
        \hspace{0.2cm} 
        \includegraphics[width=0.18\textwidth]{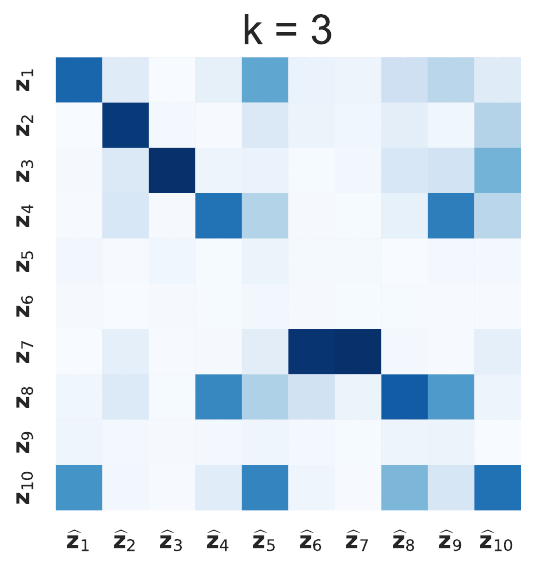}
        \hspace{0.0cm} 
        \includegraphics[width=0.0375\textwidth]{Figures/cbar.pdf}
        \caption{Linear mixing function with linear causal relation and Gaussian noise: ablation study on increasing the density of causal graphs $k$ from 0 to 3 and fixing $n=10$, $\delta=0\sigma$, $\rho=50$. The case with a denser graph is more complicated to learn the latent variables.}
        \label{fig:linear_gauss_ablation_k_indep}
    \end{center}
\end{figure*}

\begin{figure*}[!htbp]
    \begin{center}
        \includegraphics[width=0.18\textwidth]{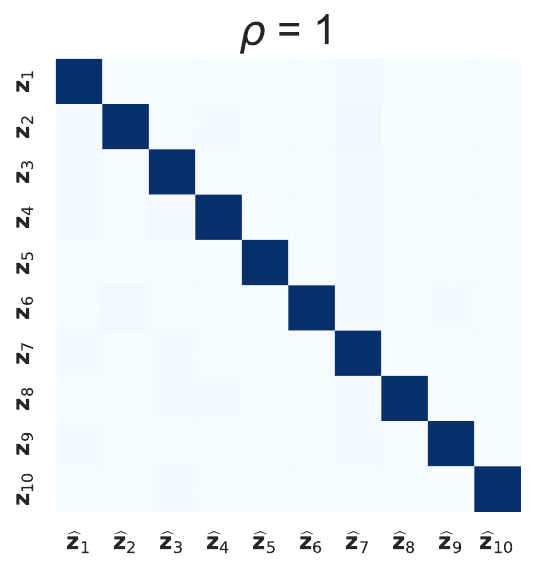}
        \hspace{0.2cm} 
        \includegraphics[width=0.18\textwidth]{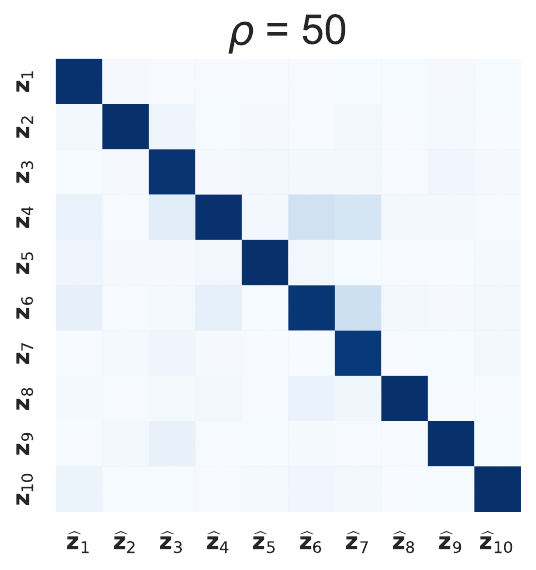}
        \hspace{0.2cm} 
        \includegraphics[width=0.18\textwidth]{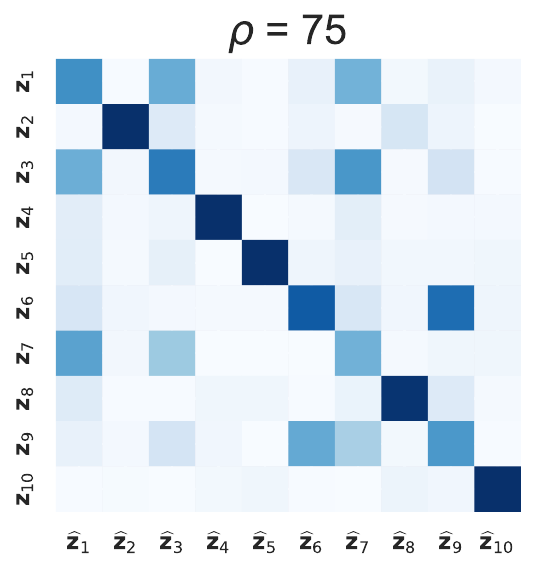}
        \hspace{0.0cm} 
        \includegraphics[width=0.0375\textwidth]{Figures/cbar.pdf}
        \caption{Linear mixing function with linear causal relation and Gaussian noise: ablation study on increasing the ratio of active (unmasked) variables $\rho$ from 1 variable only to 75\% and fixing $n=10$, $\delta=0\sigma$, $k=1$. Learning the latent variables is more complicated in the case with a larger portion of active variables.}
        \label{fig:linear_gauss_ablation_rho_indep}
    \end{center}
\end{figure*}

\begin{figure*}[!htbp]
    \begin{center}
        \includegraphics[width=0.18\textwidth]{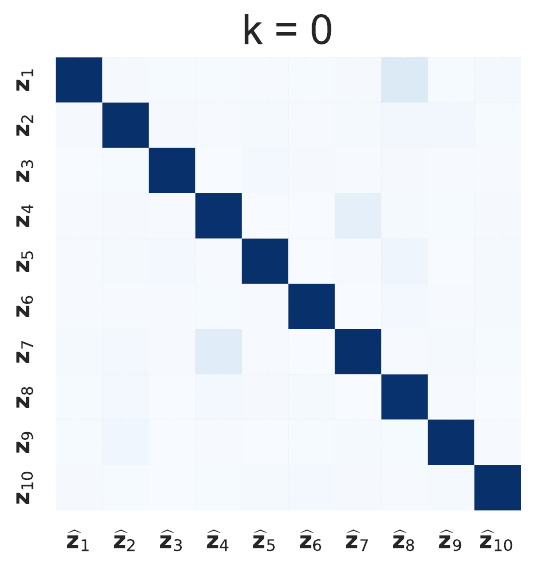}
        \hspace{0.2cm} 
        \includegraphics[width=0.18\textwidth]{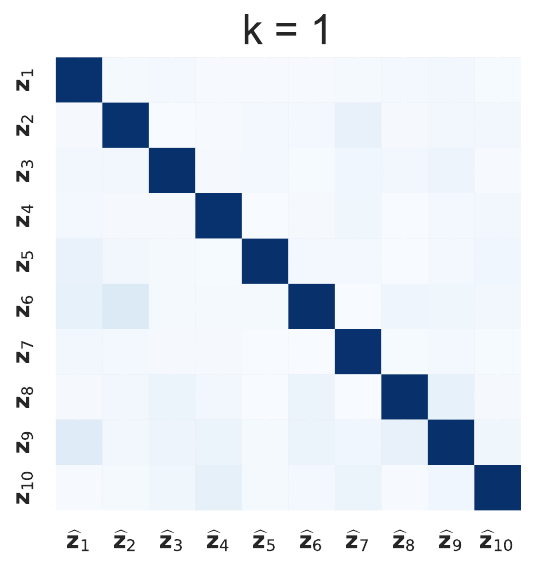}
        \hspace{0.2cm} 
        \includegraphics[width=0.18\textwidth]{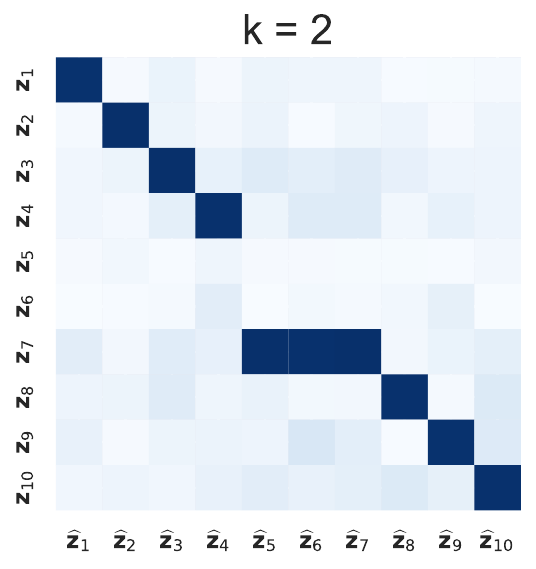}
        \hspace{0.2cm} 
        \includegraphics[width=0.18\textwidth]{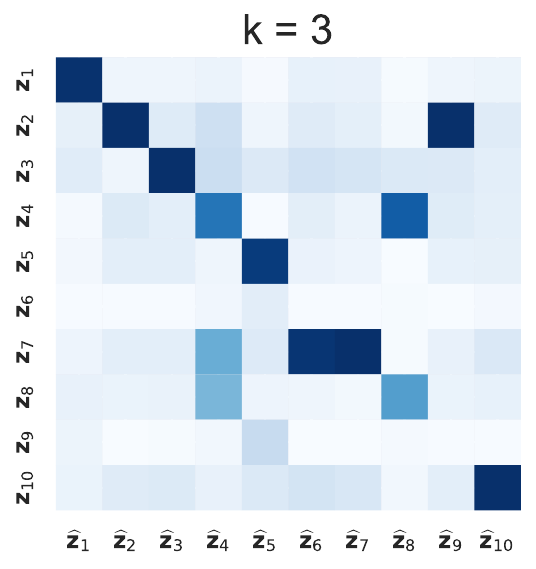}
        \hspace{0.0cm} 
        \includegraphics[width=0.0375\textwidth]{Figures/cbar.pdf}
        \caption{Linear mixing function with linear causal relation and Exponential noise: ablation study on increasing the density of causal graphs $k$ from 0 to 3 and fixing $n=10$, $\delta=0\sigma$, $\rho=50$. The case with a denser graph is more complicated to learn the latent variables.}
        \label{fig:linear_exp_ablation_k_indep}
    \end{center}
\end{figure*}

\begin{figure*}[!htbp]
    \begin{center}
         \includegraphics[width=0.18\textwidth]{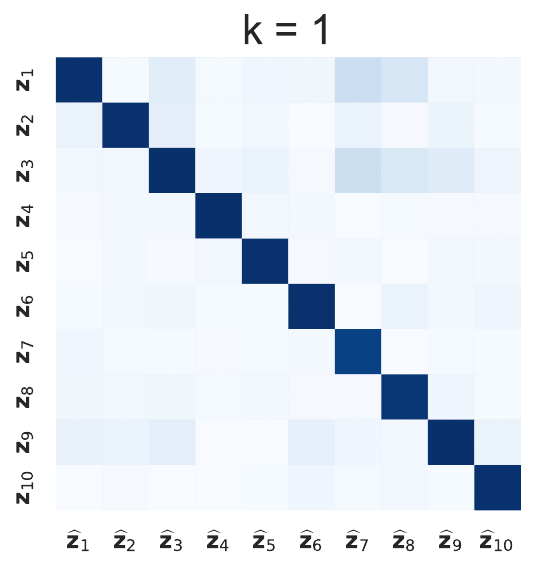}
        \hspace{0.2cm} 
        \includegraphics[width=0.18\textwidth]{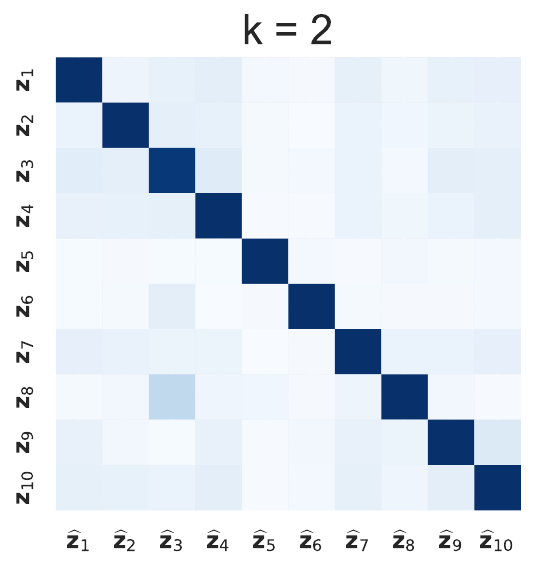}
        \hspace{0.2cm} 
        \includegraphics[width=0.18\textwidth]{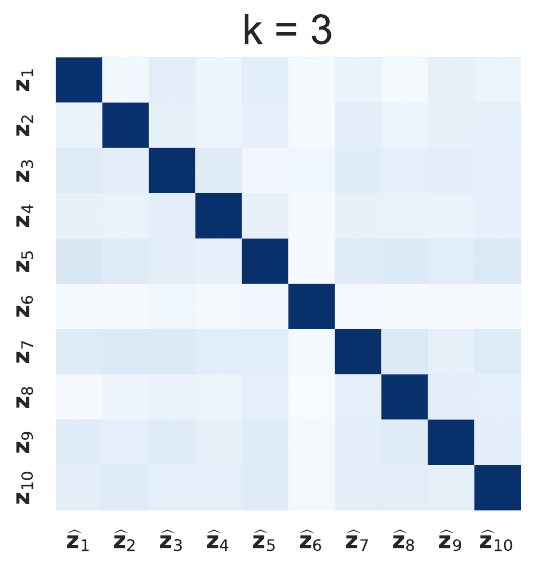}
        \hspace{0.0cm} 
        \includegraphics[width=0.0375\textwidth]{Figures/cbar.pdf}
        \caption{Linear mixing function with nonlinear causal relation and Gaussian noise: ablation study on increasing the density of causal graphs $k$ from 0 to 3 and fixing $n=10$, $\delta=0\sigma$, $\rho=50$. The case with a denser graph is more complicated to learn the latent variables.}
        \label{fig:linear_nonlinear_ablation_k_indep}
    \end{center}
\end{figure*}
\FloatBarrier

\begin{table*}[!htbp]
\caption{Results for the numerical experiments in the piecewise linear case for training on \emph{dependent} variable but testing on \emph{independent} variable case.}

\label{tab: num_piecewise_indep}
\begin{center}
\begin{small}
\begin{sc}
\begin{tabular}{|c|c|c|c|c|c|c|}
\hline
$n$ & $m$ & $k$ &  $\rho$ & $\delta$ & MCC(\cref{tab: num_piecewise}) & MCC(TEST ON INDEPENDENT LATENTS)        \\ \hline
5    &  10  & 1  & 50 \%  & 2 & 0.898$\pm$0.018 & 0.923$\pm$0.021 \\
10   &   10  & 1 & 50 \%  & 2 & 0.799$\pm$0.020& 0.807$\pm$0.023\\
20   &  10  & 1  & 50 \% & 2 & 0.753$\pm$0.015& 0.800$\pm$0.022\\
40   &   10   & 1 & 50 \%  & 2 & 0.782$\pm$0.012& 0.869$\pm$0.013\\
\hline
10   &  10  & 1 & 50 \%  & 2& 0.799$\pm$0.020& 0.807$\pm$0.023\\
10   &  10  & 2 & 50 \%  & 2  &0.789$\pm$0.021& 0.836$\pm$0.036\\
10   &  10   & 3  & 50 \% & 2  &0.785$\pm$0.022& 0.847$\pm$0.036
\\ \hline
10   &  3   & 1 & 50 \%  &2  & 0.867$\pm$0.007& 0.923$\pm$0.034\\
10   &  10   & 1  & 50 \%  & 2  &0.799$\pm$0.020& 0.807$\pm$0.023\\
10   &  20   & 1  & 50 \% &2 & 0.552$\pm$0.035& 0.537$\pm$0.035\\
\hline 
10   &  10   & 1 & 1\textsc{var}  &2  &0.908$\pm$0.013& 0.946$\pm$0.011\\
10   &  10   & 1 & 50 \%  &2  &0.799$\pm$0.020& 0.807$\pm$0.023\\
10   &  10   & 1 & 75 \% & 2  &0.695$\pm$0.026& 0.678$\pm$0.030\\
\hline
10   &  10   & 1 & 50 \% &0   &0.185$\pm$0.036& 0.232$\pm$0.038\\
10   &  10   & 1 & 50 \% &3   &0.856$\pm$0.015& 0.881$\pm$0.018\\
10   &  10   & 1 & 50 \% & 5  &0.930$\pm$0.010& 0.949$\pm$0.009\\
10   &  10   & 1 & 50 \%  & 10  &0.980$\pm$0.007& 0.986$\pm$0.006\\
\hline
\end{tabular}
\end{sc}
\end{small}
\end{center}
\end{table*}
\FloatBarrier

\begin{figure*}[!htbp]
    \begin{center}
        \includegraphics[width=0.18\textwidth]{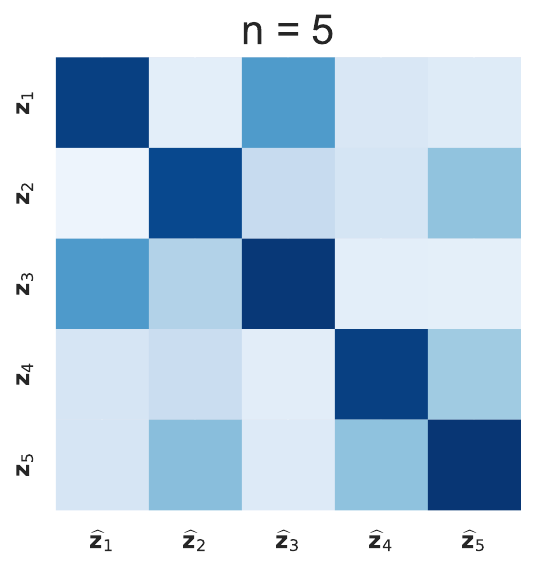}
        \hspace{0.2cm} 
        \includegraphics[width=0.18\textwidth]{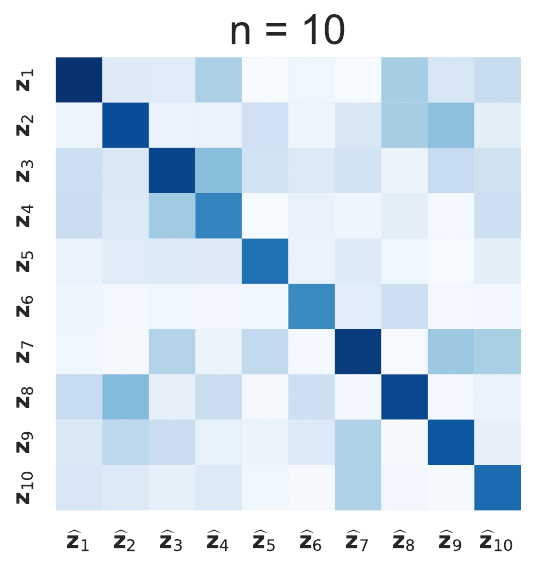}
        \hspace{0.2cm} 
        \includegraphics[width=0.18\textwidth]{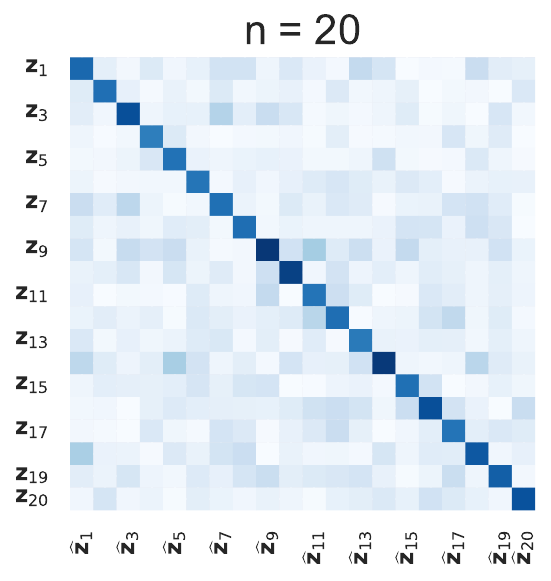}
        \hspace{0.2cm} 
        \includegraphics[width=0.18\textwidth]{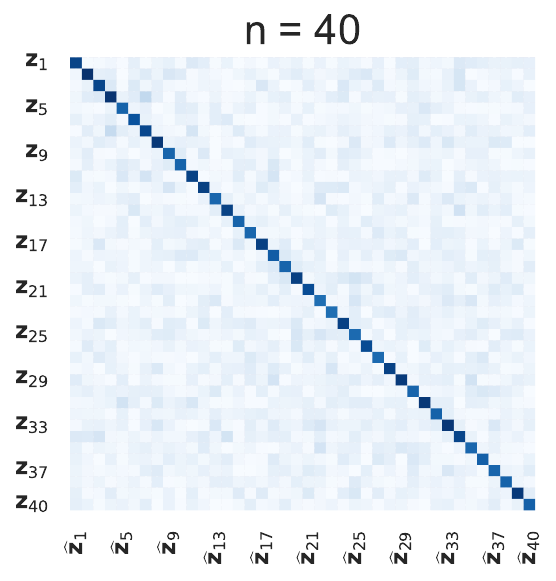}
        \includegraphics[width=0.031\textwidth]{Figures/cbar.pdf}
        \caption{Piecewise linear mixing function with linear causal relation and Gaussian noise: ablation study on increasing the latent dimension $n$ from 5 to 40 and fixing $\delta=2.0\sigma$, $m=10$, $\rho=50$, $k=1$. The case with higher $n$ is more complicated to learn the latent variables.}
        \label{fig:pw_ablation_n_indep}
    \end{center}
\end{figure*}

\begin{figure*}[!htbp]
    \begin{center}
        \includegraphics[width=0.18\textwidth]{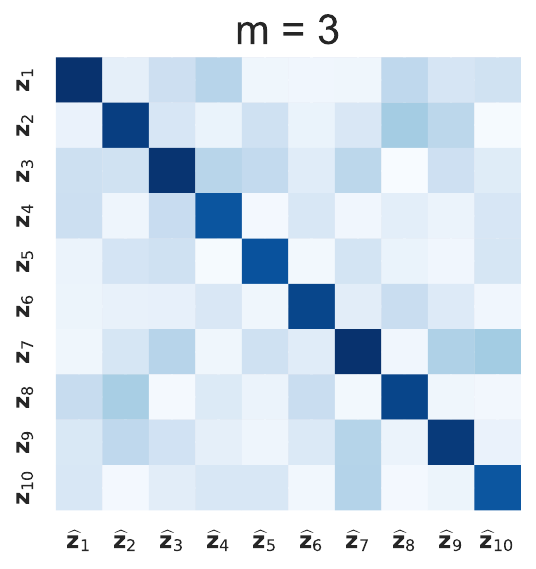}
        \hspace{0.2cm} 
        \includegraphics[width=0.18\textwidth]{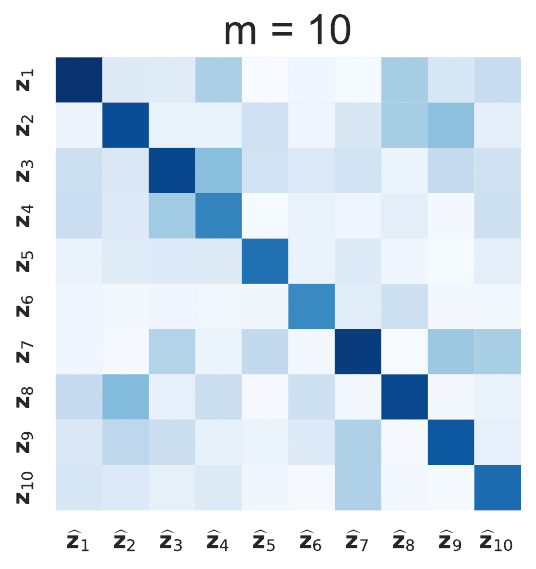}
        \hspace{0.2cm} 
        \includegraphics[width=0.18\textwidth]{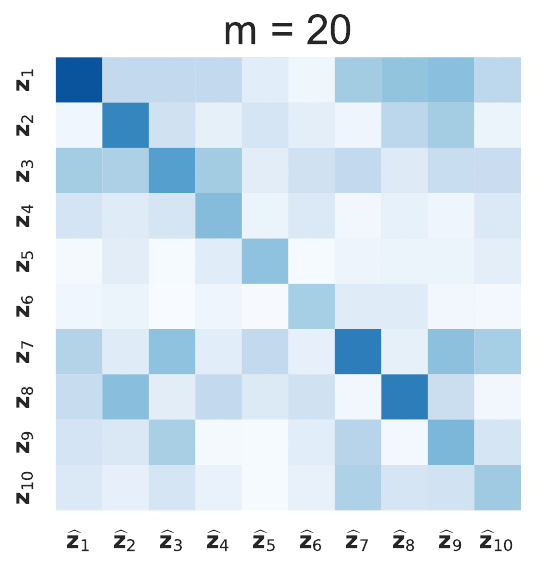}
        \includegraphics[width=0.031\textwidth]
        {Figures/cbar.pdf}
        \caption{Piecewise linear mixing function with linear causal relation and Gaussian noise: ablation study on increasing the number of Leaky-ReLU layers $(m-1)$ from 3 to 20 and fixing $n=10$, $\delta=2.0\sigma$, $\rho=50$, $k=1$. The case with larger $m$ is more complicated to learn the latent variables due to a greater extent of nonlinearity.}
        \label{fig:pw_ablation_m_indep}
    \end{center}
\end{figure*}

\begin{figure*}[!htbp]
    \begin{center}
        \includegraphics[width=0.18\textwidth]{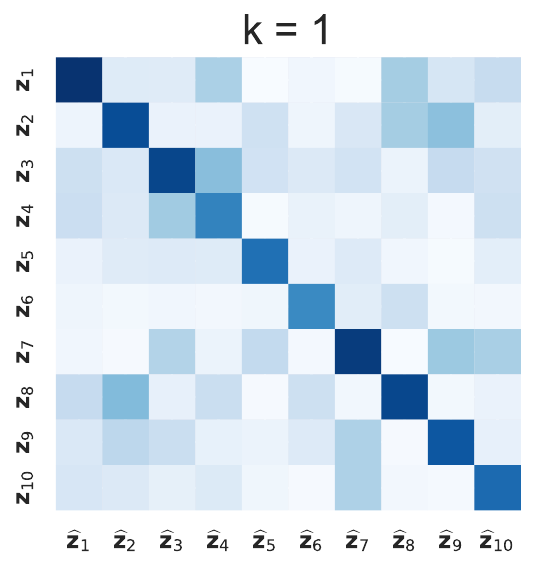}
        \hspace{0.2cm} 
        \includegraphics[width=0.18\textwidth]{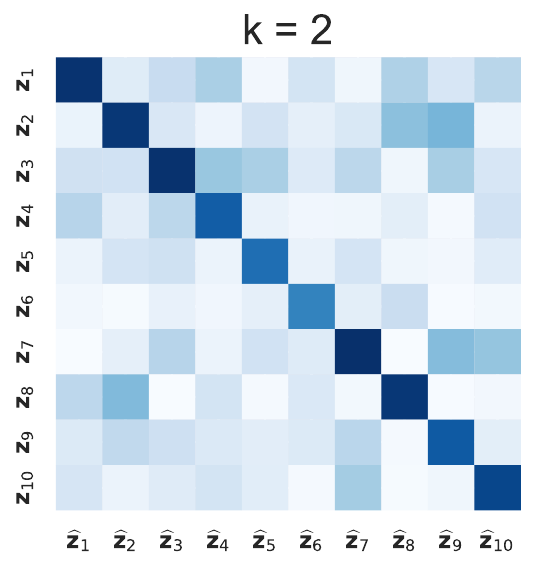}
        \hspace{0.2cm} 
        \includegraphics[width=0.18\textwidth]{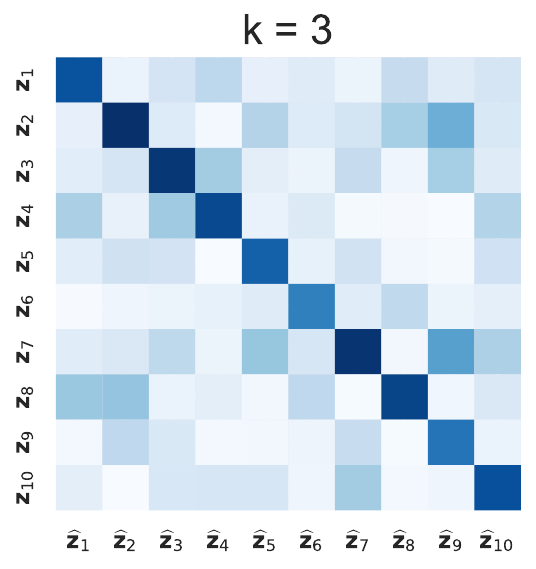}
        \includegraphics[width=0.031\textwidth]{Figures/cbar.pdf}
        \caption{Piecewise linear mixing function with nonlinear causal relation and Gaussian noise: ablation study on increasing the density of causal graphs $k$ from 0 to 3 and fixing $n=10$, $m=10$, $\delta=2.0\sigma$, $\rho=50$. In the case with a denser graph, it is more complicated to learn the latent variables.}
        \label{fig:pw_ablation_k_indep}
    \end{center}
\end{figure*}

\begin{figure*}[!htbp]
    \begin{center}
       \includegraphics[width=0.18\textwidth]{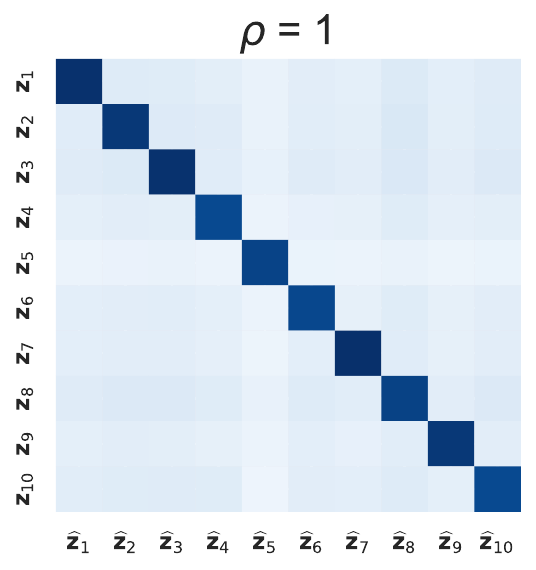}
        \hspace{0.2cm} 
        \includegraphics[width=0.18\textwidth]{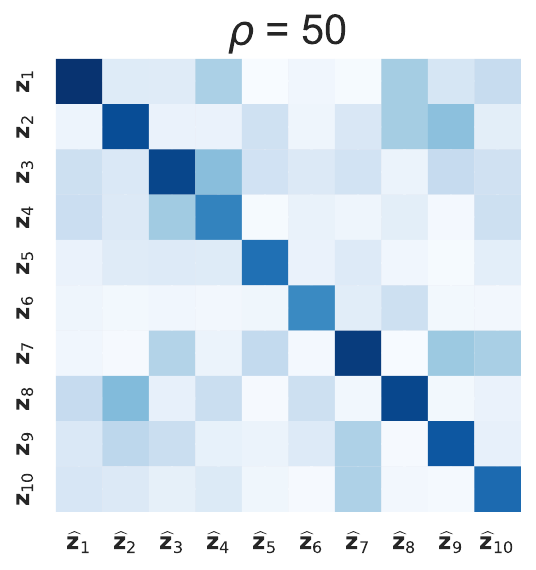}
        \hspace{0.2cm} 
        \includegraphics[width=0.18\textwidth]{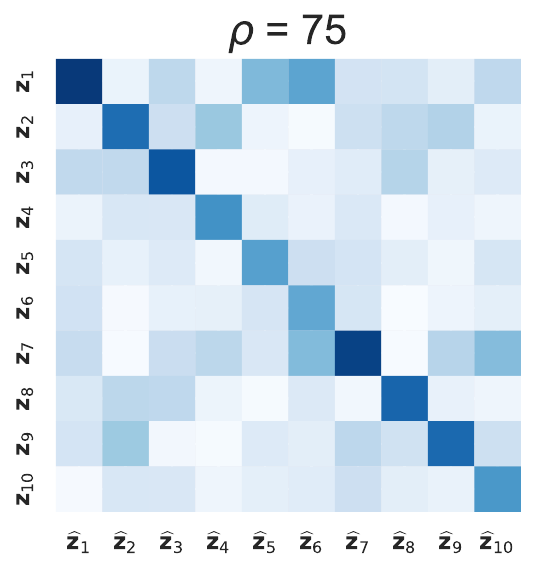}
        \hspace{0.2cm} 
        \includegraphics[width=0.031\textwidth]{Figures/cbar.pdf}
        \caption{Piecewise linear mixing function with linear causal relation and Gaussian noise: ablation study on increasing the ratio of active (unmasked) variables $\rho$ from 1 variable only to 75\% and fixing $n=10$, $m=10$, $\delta=2.0\sigma$, $k=1$. Learning the latent variables is more complicated in the case with a larger portion of active variables.}
        \label{fig:pw_ablation_rho_indep}
    \end{center}
\end{figure*}

\begin{figure*}[!htbp]
    \begin{center}
        \includegraphics[width=0.18\textwidth]{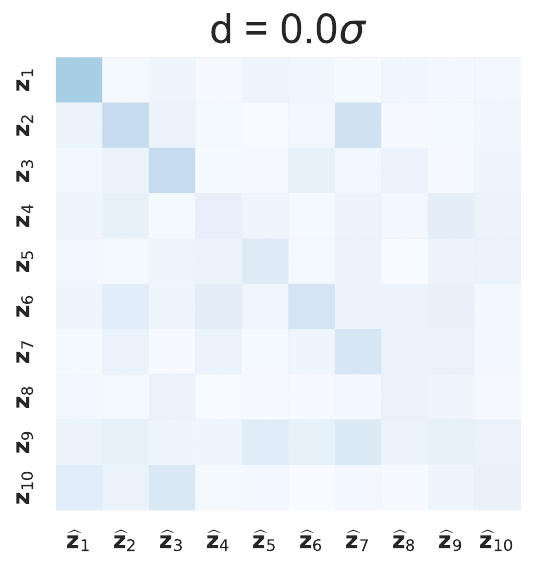}
        \hspace{0.1cm} 
        \includegraphics[width=0.18\textwidth]{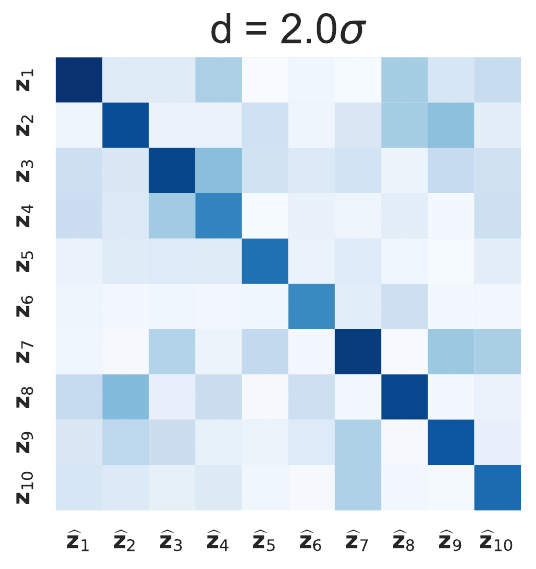}
        \hspace{0.1cm} 
        \includegraphics[width=0.18\textwidth]{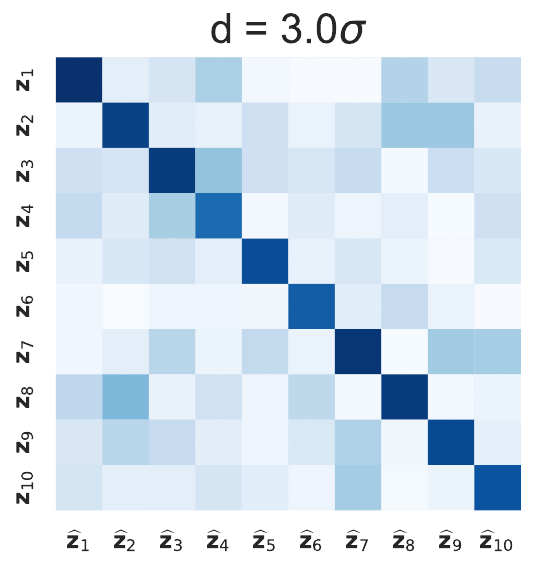}
        \hspace{0.1cm} 
        \includegraphics[width=0.18\textwidth]{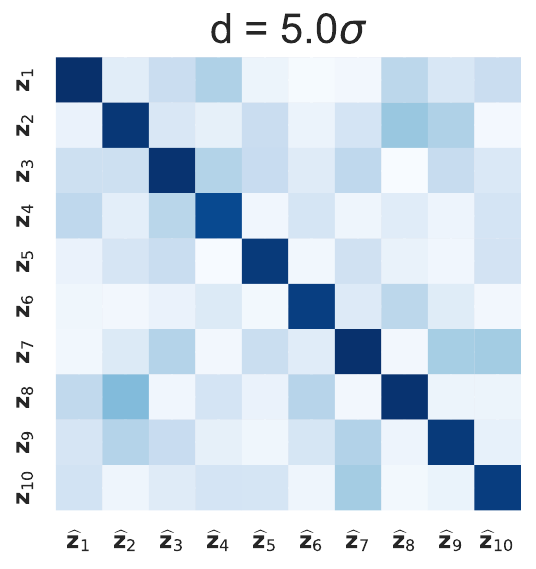}
        \hspace{0.1cm} 
        \includegraphics[width=0.18\textwidth]{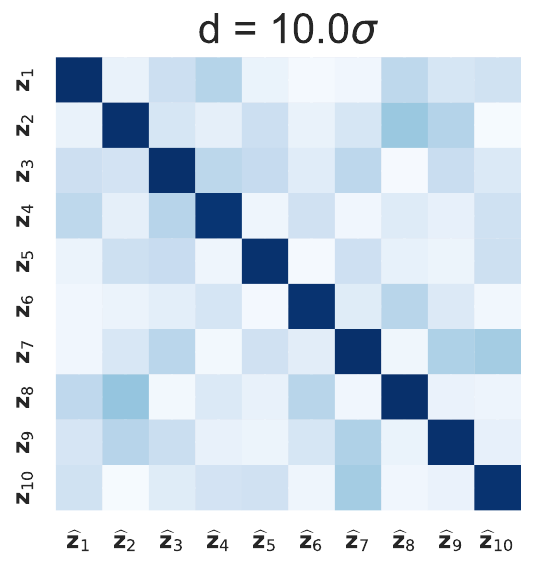}
        \includegraphics[width=0.031\textwidth]{Figures/cbar.pdf}
        \caption{Piecewise linear mixing function with linear causal relation and Gaussian noise: ablation study on increasing the distance between masked value and mean of latents from $0.0 \sigma$ to $10.0 \sigma$, where $\sigma$ is the standard deviation of latents, and fixing $n=10$, $m=10$, $\rho=50$, $k=1$. Learning the latent variables is more complicated in the case with a smaller distance.}
        \label{fig:pw_ablation_delta_indep}
    \end{center}
\end{figure*}
\FloatBarrier

\subsubsection{Over-parameterization of the number of causal representations}
\label{app: over-parameterization}
Our theoretical result assume that we know the number of causal variables in advance, but in practice (and as often also the case in other CRL methods), we can show empirically that if we overestimate the true number of latent variables, we will still be able to identify the true causal variables, while some of them will be unused. We provide ablation study below, which seem to confirm this empirically.

In the experiments below we consider $n=10$ ground truth causal variables, and $nn$ latent variables that we estimate, where $nn \geq n$. For all experiments, we consider an average over 3 random seeds. We use a causal graph ER- $k$, where ER-$k$ is a graph with $n \cdot k$ edges, $\delta$ is the distance between mask value and mean of causal variables and $\rho$ denotes the ratio of measured variables. For the piecewise linear case $m$ denotes the number of hidden layers in the MLP. We choose one representative setting for the linear case and one representative setting for the piecewise linear case, and report how the MCC varies when we increase $nn$ from the original size $nn=n$ to double the size $nn=2n$. We report the average MCC over 3 random seeds and a heatmap of the correlation matrix for each of the settings below. 

\begin{table}[h!]
\centering
\begin{tabular}{lllllll}
\hline
 $nn (n=10)$ & $10 (n+0)$ & $12 (n+2)$ & $14 (n+4)$ & $16 (n+6)$ & $18 (n+8)$& $20 (n+10)$\\ 
\hline
MCC       & 0.998$\pm$0.001 & 0.931$\pm$0.056 & 0.965$\pm$0.051 & 0.991$\pm$0.001 & 0.996$\pm$0.001 & 0.978$\pm$0.031 \\
\hline
\end{tabular}
\caption{Average MCCs over 3 random seeds for linear $\fb$ with $n=10$, $\delta=0.0\sigma$, $\rho$=50, $k=1$, Linear Gaussian SCM}
\label{tab: abla_nn_linearmix}
\end{table}

\begin{table}[h!]
\centering
\begin{tabular}{lllllll}
\hline
 $nn (n=10)$ & $10 (n+0)$ & $12 (n+2)$ & $14 (n+4)$ & $16 (n+6)$ & $18 (n+8)$& $20 (n+10)$\\ 
\hline
MCC       & 0.796$\pm$0.026 & 0.798$\pm$0.012 & 0.804$\pm$0.015 & 0.809$\pm$0.012 & 0.814$\pm$0.014 & 0.816$\pm$0.012 \\
\hline
\end{tabular}
\caption{Average MCCs over 3 random seeds for piecewise $\fb$ with $n=10$, $\delta=2.0\sigma$, $\rho$=50, $k=1$, $m=10$, Linear Gaussian SCM}
\label{tab: abla_nn_pwmix}
\end{table}

\begin{figure*}
    \begin{center}
        \includegraphics[width=0.20\textwidth]{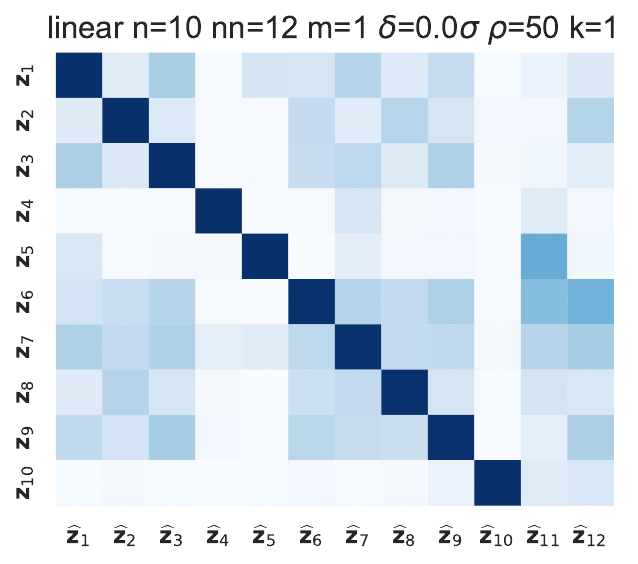}
        \hspace{0.25cm} 
        \includegraphics[width=0.23\textwidth]{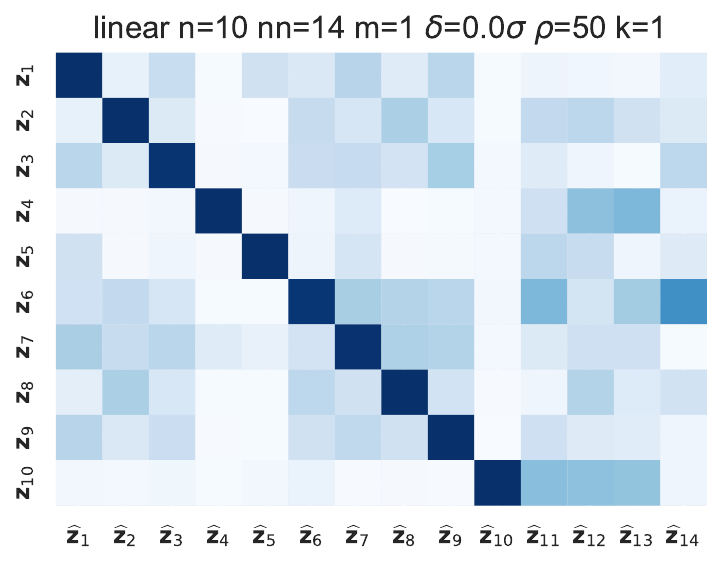}
        \hspace{0.25cm} 
        \includegraphics[width=0.26\textwidth]{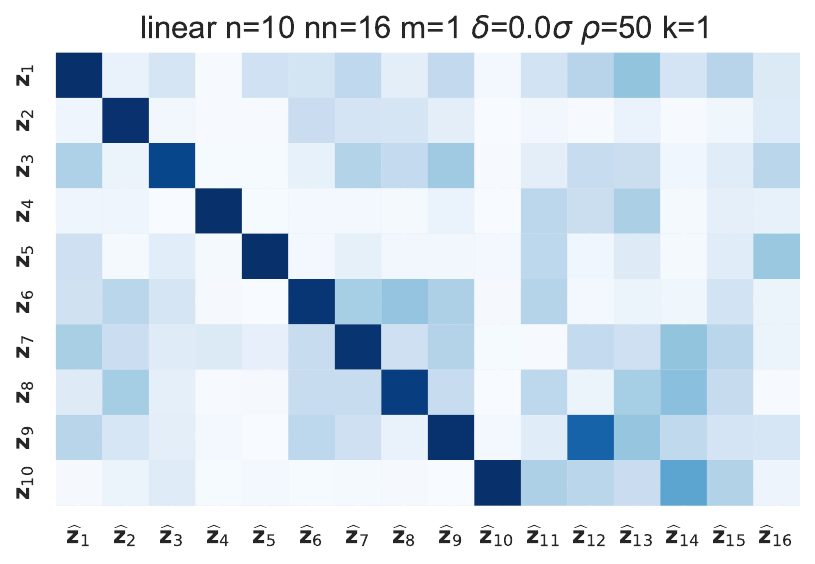}
        \includegraphics[width=0.036\textwidth]{Figures/cbar.pdf}
        \includegraphics[width=0.29\textwidth]{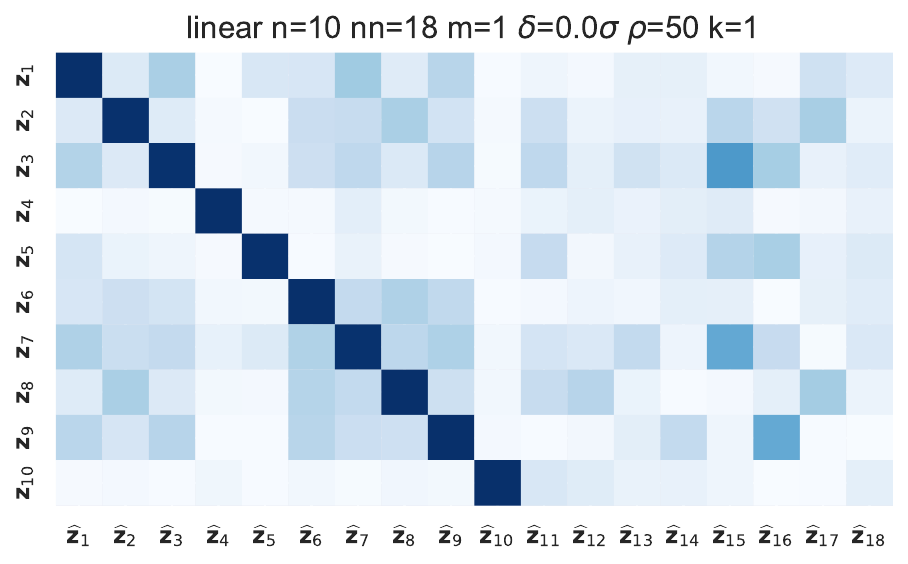}
        \hspace{0.25cm}
        \includegraphics[width=0.32\textwidth]{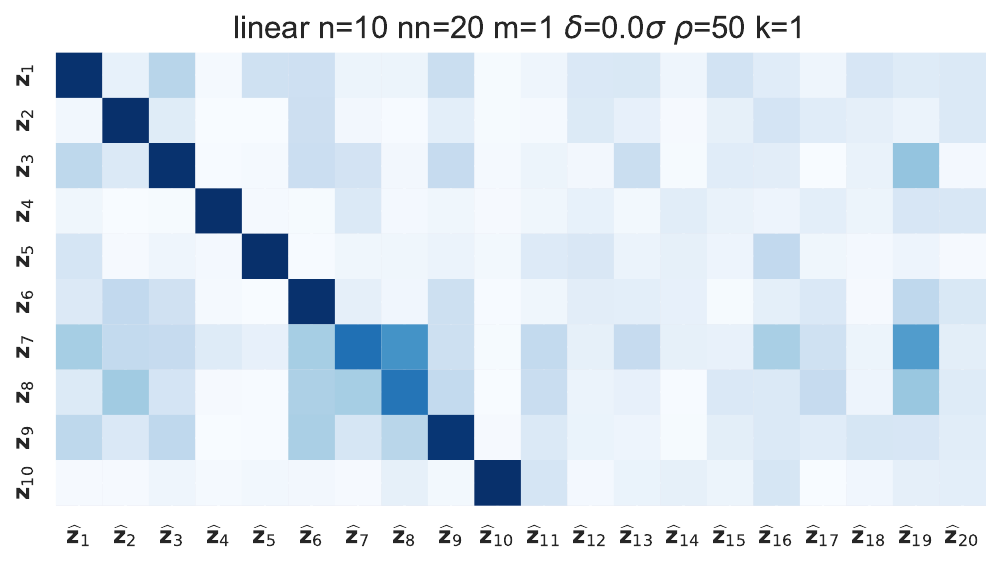}
        \includegraphics[width=0.036\textwidth]{Figures/cbar.pdf}
        \caption{Ablation study on number of overestimated representations $nn$ for linear $\fb$}
        \label{fig:linear_ablation_nn}
    \end{center}
\end{figure*}

\begin{figure*}
    \begin{center}
        \includegraphics[width=0.215\textwidth]{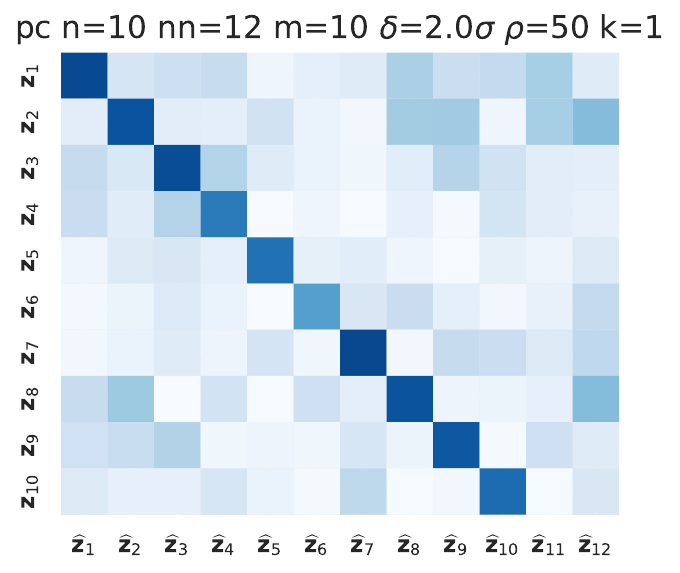}
        \hspace{0.25cm} 
        \includegraphics[width=0.23\textwidth]{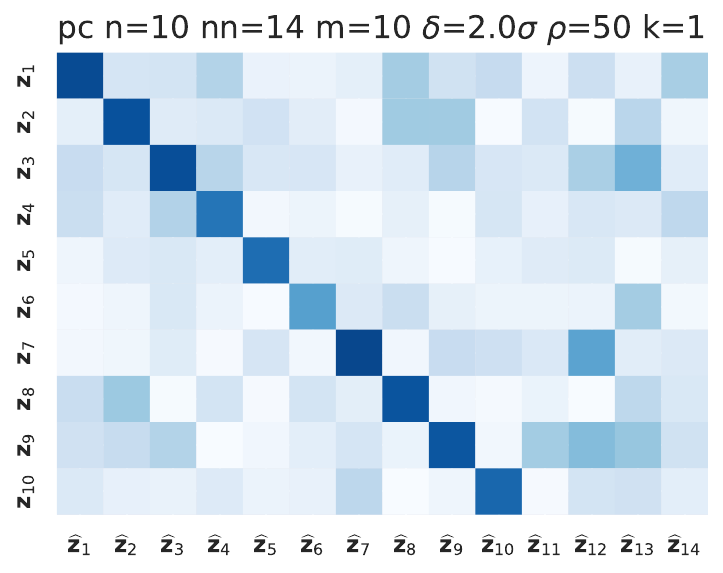}
        \hspace{0.25cm} 
        \includegraphics[width=0.26\textwidth]{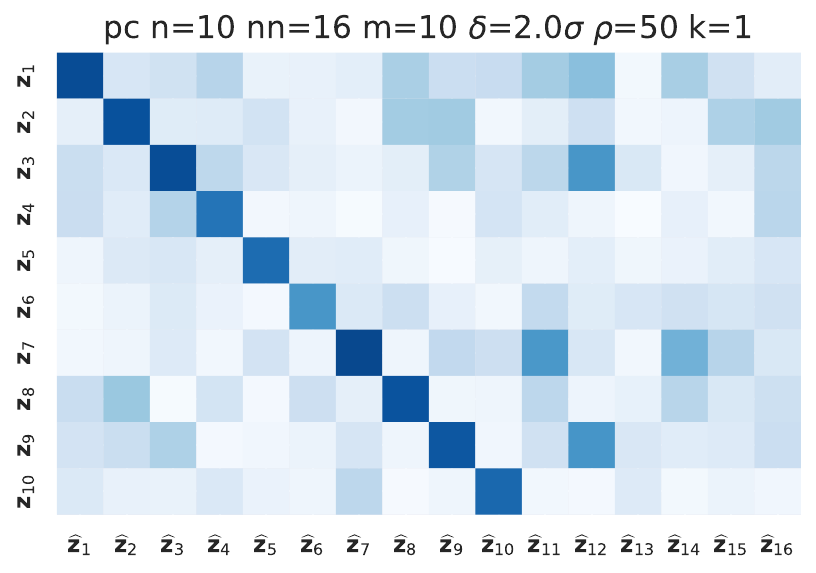}
        \includegraphics[width=0.036\textwidth]{Figures/cbar.pdf}
        \includegraphics[width=0.29\textwidth]{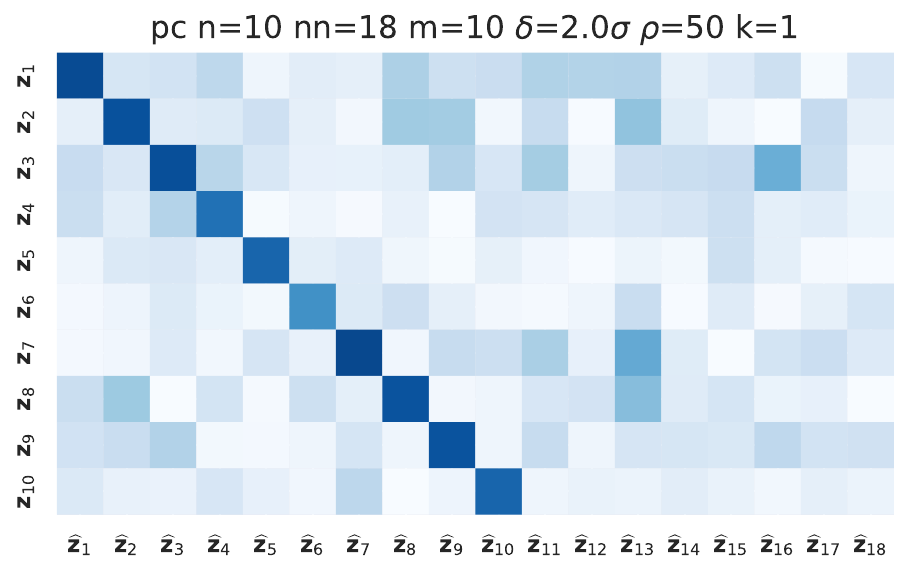}
        \hspace{0.25cm}
        \includegraphics[width=0.32\textwidth]{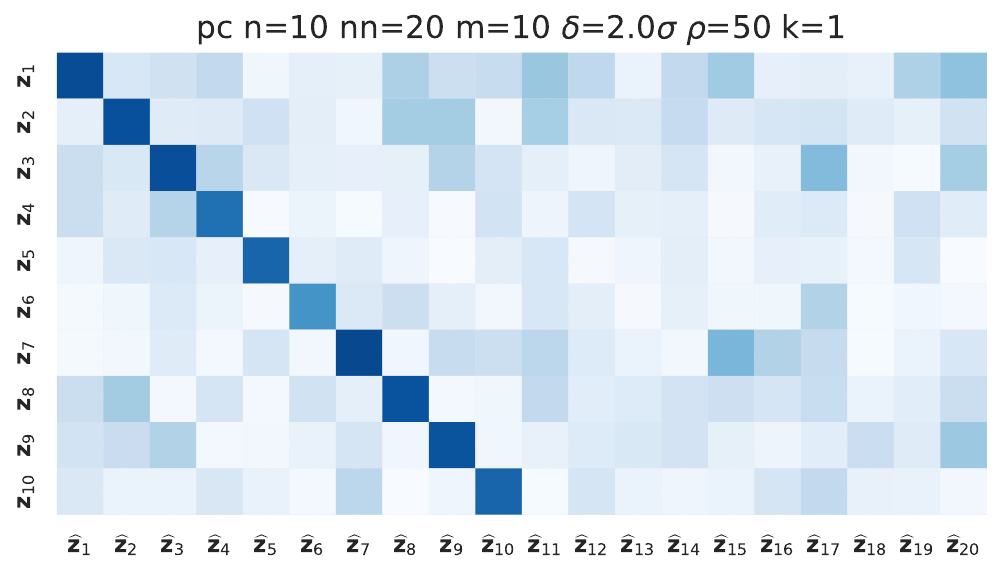}
        \includegraphics[width=0.036\textwidth]{Figures/cbar.pdf}
        \caption{Ablation study on the number of overestimated representations $nn$ for piecewise linear $\fb$ }
        \label{fig:pw_ablation_nn}
    \end{center}
\end{figure*}

As shown in both \cref{tab: abla_nn_linearmix} and \cref{tab: abla_nn_pwmix}, our method seems to perform similarly well, even if we do not know the true number of causal variables, but we overestimate them. We provide the heatmaps of the correlation matrix between $\Zb$ and $\hat{\Zb}$ for one of the seeds with different $nn$'s in \cref{fig:linear_ablation_nn} and \cref{fig:pw_ablation_nn}.

\FloatBarrier

\FloatBarrier

\subsubsection{Causal Discovery on Estimated Latent Causal Variables}
\label{app: graph_learning}
In this section, we evaluate how can we use the latent causal variables we learn as input to  causal discovery methods in order to learn the causal structure.
In causal representation learning often the main difficulty is to identify the causal variables from the observations, while we can then use standard causal discovery methods to estimate an equivalence class of graphs. In this case, intuitively we expect that  the closer the estimated causal variables are to the ground truth, the closer the learned graph to the ground truth causal relations.

Since masking might also mask some causal relations in the original graph, we instead consider a group of data without masked variables and then compare the learned graph to the ground truth.
For this task, we additionally assume a few standard assumptions in causal discovery \citep{Spirtes_2000}, e.g. the causal Markov assumption and the causal faithfulness assumption, which together imply that conditional independences in the data correspond to d-separations in the true underlying causal graph. For simplicity, we also assume causal sufficiency, which in this case means that there are no additional latent confounders of the reconstructed latent variables, and there is no selection bias.

We consider a set of different algorithms for different underlying causal models, based on the parametric assumptions we use in each setting:
\begin{itemize}
    \item Linear Gaussian SCM: PC algorithm with partial correlation tests with significance threshold $\alpha=0.01$ from the \emph{pcalg}~\citep{kalisch2012causal} package,
    \item Linear Exponential SCM: Pairwise LiNGAM~\citep{hyvarinen2013pairwise},
    \item Nonlinear SCM: PC algorithm with Hilbert Schmidt Independence Criterion (HSIC) test with significance threshold $\alpha=0.01$ from the \emph{kpcalg}~\citep{verbyla2017kpcalg} package.
\end{itemize}
While LiNGAM provides in output a Directed Acyclic Graph (DAG), which can be easily compared with the ground truth DAG, the PC algorithm instead outputs a Completely Partially Oriented DAG (CPDAG). In this case, we consider the ground truth CPDAG (the CPDAG of the Markov Equivalence Class that contains the ground truth DAG).

We evaluate the Structural Hamming distance (SHD) with respect to the ground truth causal graph for LiNGAM and to the ground truth CPDAG for PC. For each setting we consider two distances:
\begin{itemize}
    \item $SHD_{\zb^*}$: the DAG or CPDAG learned with the ground truth causal variables $\zb^*$
    \item $SHD_{\hat{\zb}}$ the DAG or CPDAG learned with the estimated causal variables $\hat{\zb}$
\end{itemize}
Since our primary objective is to recover the causal variables, the graph learned with the ground truth $\Zb$ is the optimal result we can attain.
Therefore we calculate the difference between these two SHDs and we use $\Delta_{SHD}$ to denote it. Smaller $\Delta_{SHD}$ implies a closer estimated graph to the optimal one. 

\begin{figure*}
\caption{MCC-$\Delta_{shd}$ of all linear mixing function cases.}
    \begin{center}
        \includegraphics[width=0.35\textwidth]{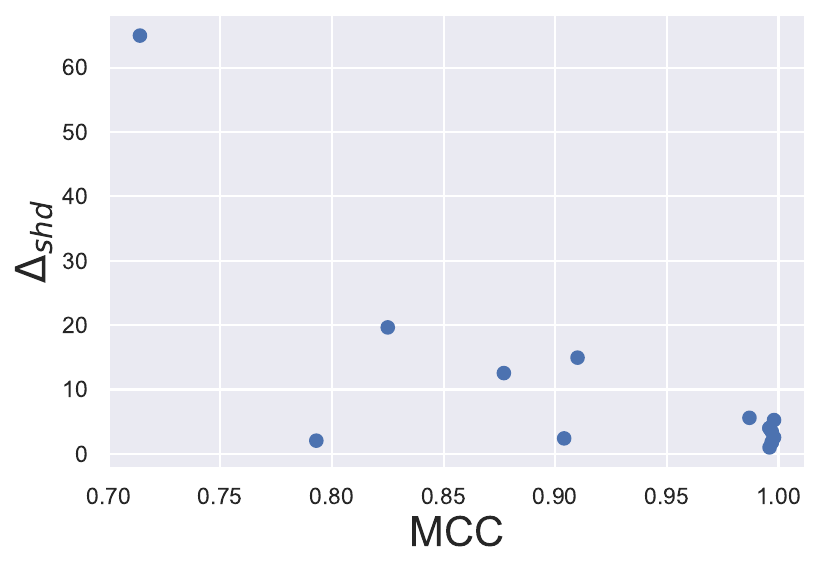}
        
        \label{fig:mcc-shd}
    \end{center}
\end{figure*}

We provide results for our method for linear mixing functions in Table~\ref{tab: linear_learn_graph} and for piecewise linear mixing functions in Table~\ref{tab: pw_our_learn_graph}.
We additionally provide results for the oracle version of our method, which uses the known masks for both settings in Table~\ref{tab: pw_oracle_learn_graph}. 
All of these results show a negative correlation between higher MCC (and hence more accurate estimated representations) and lower $\Delta_{SHD}$ (and hence more accurate estimated causal graphs w.r.t. to the standard causal discovery setting). This is even clear in the scatterplot for the linear case, shown in Fig.~\ref{fig:mcc-shd}. This aligns with the intuition that more well-identified representations tend to lead to graph learning that closely mirrors the one learned from ground truth.

\FloatBarrier

\begin{table}[!htbp]
\caption{Results of learning graph for linear mixing functions.
}
\label{tab: linear_learn_graph}
\begin{center}
\begin{small}
\begin{sc}
\begin{tabular}{ccccccc}
\hline
$n$ & $k$ & SCM &  $\rho$ & $SHD_{\zb^*}$ & $SHD_{\hat{\zb}}$ & $\Delta_{SHD}$     \\ \hline
\textbf{5}    & 1 &Lin. Gauss  & 50 \%  &  0.65$\pm$1.57 &  2.40$\pm$1.90  &  1.75\\
\textbf{10}   & 1 &Lin. Gauss &  50 \%   &  1.9$\pm$1.89 &  5.90$\pm$3.06  &  4.00\\
\textbf{20}   &  1 & Lin. Gauss  & 50 \%  &  7.15$\pm$3.92 &  12.75$\pm$5.56  &  5.60 \\
\textbf{40}   & 1 &Lin. Gauss   & 50 \%   &  13.70$\pm$4.53 &  79.60$\pm$27.08  &  65.90 \\
\hline
10    & \textbf{1} &Lin. Gauss  & 50 \%   &  1.9$\pm$1.89 &  5.90$\pm$3.06  &  4.00\\
10   &  \textbf{2} &Lin. Gauss & 50 \%  &  16.20$\pm$2.38 &  18.60$\pm$4.10  &  2.40\\
10   &  \textbf{3} &Lin. Gauss  & 50 \%  &  27.20$\pm$2.65 &  29.25$\pm$3.29  &  2.05\\
\hline
10   &  \textbf{1} & Lin. Exp   & 50 \%   &  2.95$\pm$2.26 &  8.20$\pm$4.83  &  5.25\\
10    & \textbf{2} & Lin. Exp  & 50 \%  &  3.70$\pm$2.66 &  18.65$\pm$11.81  &  14.95\\
10   & \textbf{3} &Lin. Exp & 50 \%   &  5.20$\pm$4.51 &  24.85$\pm$11.39  &  19.65  \\
\hline
10   &  \textbf{1} & Nonlinear  & 50 \%  &  5.55$\pm$2.50 &  9$\pm$2.94  &  3.45\\
10   &  \textbf{2} & Nonlinear  & 50 \%  &  14.4$\pm$2.04 &  16.25$\pm$3.39  &  1.85\\
10   &  \textbf{3} & Nonlinear  & 50 \%  &  23.4$\pm$2.44 &  25.4$\pm$3.99  &  1\\
\hline
10    & 1 &Lin. Gauss  & \textbf{1\text{var}}  &  2.1$\pm$1.89 &  4.65$\pm$2.89  &  2.55\\
10   &  1 &Lin. Gauss & \textbf{50 \%}   &  1.9$\pm$1.89 &  5.90$\pm$3.06  &  4.00\\
10   &  1 &Lin. Gauss  & \textbf{75 \%}  &  2.15$\pm$1.79 &  14.70$\pm$5.25  &  12.55\\
\hline
\end{tabular}
\end{sc}
\end{small}
\end{center}
\end{table}
\FloatBarrier

\begin{table*}[!htbp]
\caption{Results of learning graph for piecewise linear mixing functions with our method.}

\label{tab: pw_our_learn_graph}
\begin{center}
\begin{small}
\begin{sc}
\begin{tabular}{|c|c|c|c|c|c|c|c|}
\hline
$n$ & $m$ & $k$ &  $\rho$ & $\delta$ & $SHD_{\zb^*}$ & $SHD_{\hat{\zb}}$ & $\Delta_{SHD}$ \\ \hline
5    &  10  & 1  & 50 \%  & 2 &  0.85$\pm$1.79 &  7$\pm$1.89  &  6.15 \\
10   &   10  & 1 & 50 \%  & 2  &  2$\pm$2.08  &  19.45$\pm$2.35  &  17.45\\
20   &  10  & 1  & 50 \% & 2  &  6.8$\pm$3.49 &  46.75$\pm$5.09  &  39.95\\
40   &   10   & 1 & 50 \%  & 2  &  13.3$\pm$4.44 &  113.4$\pm$6.85  &  100.1\\
\hline
10   &  10  & 1 & 50 \%  & 2 &  2$\pm$2.08  &   19.45$\pm$2.35  &  17.45\\
10   &  10  & 2 & 50 \%  & 2  &  15.85$\pm$2.52 &  25.3$\pm$2.32  &  9.45\\
10   &  10   & 3  & 50 \% & 2  &  27.8$\pm$2.33 &  30.75$\pm$2.05  &  2.95\\ 
\hline
10   &  3    & 1 & 50 \%  &2 &  2$\pm$2.08 &  19.05$\pm$2.66  &  17.05\\
10   &  10   & 1  & 50 \%  & 2 &  2$\pm$2.08  &   19.45$\pm$2.35  &  17.45\\
10   &  20   & 1  & 50 \% &2  &  2$\pm$2.08 &  19.5$\pm$2.06  &  17.5\\
\hline 
10   &  10   & 1 & 1\textsc{var}  &2  & 2$\pm$2.08 &  19.5$\pm$1.79  &  17.5\\
10   &  10   & 1 & 50 \%  &2   &  2$\pm$2.08  &   19.45$\pm$2.35  &  17.45\\
10   &  10   & 1 & 75 \% & 2  &  2$\pm$2.08  &  18.8$\pm$2.28  &  16.8\\
\hline
10   &  10   & 1 & 50 \% &0  &  2$\pm$2.08  &  17.9$\pm$2.71  &  15.9\\
10   &  10   & 1 & 50 \% &3   &  2$\pm$2.08  &  18.9$\pm$2.17  &  16.9\\
10   &  10   & 1 & 50 \% & 5   &  2$\pm$2.08  &  18.95$\pm$1.76  &  16.95\\
10   &  10   & 1 & 50 \%  & 10  &  2$\pm$2.08  &  19.05$\pm$1.93  &  17.05\\
\hline
\end{tabular}
\end{sc}
\end{small}
\end{center}
\end{table*}
\FloatBarrier

\begin{table*}[!htbp]
\caption{Results of learning graph for piecewise linear mixing functions with oracle method.}

\label{tab: pw_oracle_learn_graph}
\begin{center}
\begin{small}
\begin{sc}
\begin{tabular}{|c|c|c|c|c|c|c|c|}
\hline
$n$ & $m$ & $k$ &  $\rho$ & $\delta$ & $SHD_{\zb^*}$ & $SHD_{\hat{\zb}}$ & $\Delta_{SHD}$ \\ \hline
5    &  10  & 1  & 50 \%  & 2 &  0.85$\pm$1.79 &  7.1$\pm$1.97  &  6.25 \\
10   &   10  & 1 & 50 \%  & 2  &  2$\pm$2.08  &  21.35$\pm$2.68  &  19.35\\
20   &  10  & 1  & 50 \% & 2  &  6.8$\pm$3.49 &  56.95$\pm$4.17  &  50.15\\
40   &   10   & 1 & 50 \%  & 2  &  13.3$\pm$4.44 &  130.85$\pm$9.84  &  117.55\\
\hline
10   &  10  & 1 & 50 \%  & 2 &  2$\pm$2.08  &  21.35$\pm$2.68  &  19.35\\
10   &  10  & 2 & 50 \%  & 2  &  15.85$\pm$2.52 &  26.1$\pm$2.77  &  10.25\\
10   &  10   & 3  & 50 \% & 2  &  27.8$\pm$2.33 &  29.9$\pm$2.75  &  2.1\\ 
\hline
10   &  3    & 1 & 50 \%  &2 &  2$\pm$2.08 &  19.8$\pm$2.80  &  17.8\\
10   &  10   & 1  & 50 \%  & 2 &  2$\pm$2.08  &  21.35$\pm$2.68  & 19.35\\
10   &  20   & 1  & 50 \% &2  &  2$\pm$2.08 &  21.25$\pm$2.84  &  19.25\\
\hline 
10   &  10   & 1 & 1\textsc{var}  &2  & 2$\pm$2.08 &  21.15$\pm$2.72  &  19.15\\
10   &  10   & 1 & 50 \%  &2   &  2$\pm$2.08  &  21.35$\pm$2.68  &  19.35\\
10   &  10   & 1 & 75 \% & 2  &  2$\pm$2.08  &  21.3$\pm$3.24  &  19.3\\
\hline
10   &  10   & 1 & 50 \% &0  &  2$\pm$2.08  &  19.55$\pm$2.87  &  17.55\\
10   &  10   & 1 & 50 \% &3   &  2$\pm$2.08  &  21.2$\pm$2.74  &  19.2\\
10   &  10   & 1 & 50 \% & 5   &  2$\pm$2.08  &  20.35$\pm$2.74  &  18.35\\
10   &  10   & 1 & 50 \%  & 10  &  2$\pm$2.08  &  19.65$\pm$2.76  &  17.65\\
\hline
\end{tabular}
\end{sc}
\end{small}
\end{center}
\end{table*}
\FloatBarrier

\subsection{Image dataset: Multiple Balls}
\label{app:experiments_multipleballs}
\begin{figure}[!htbp]
    \centering
    \includegraphics[width=.4\linewidth]{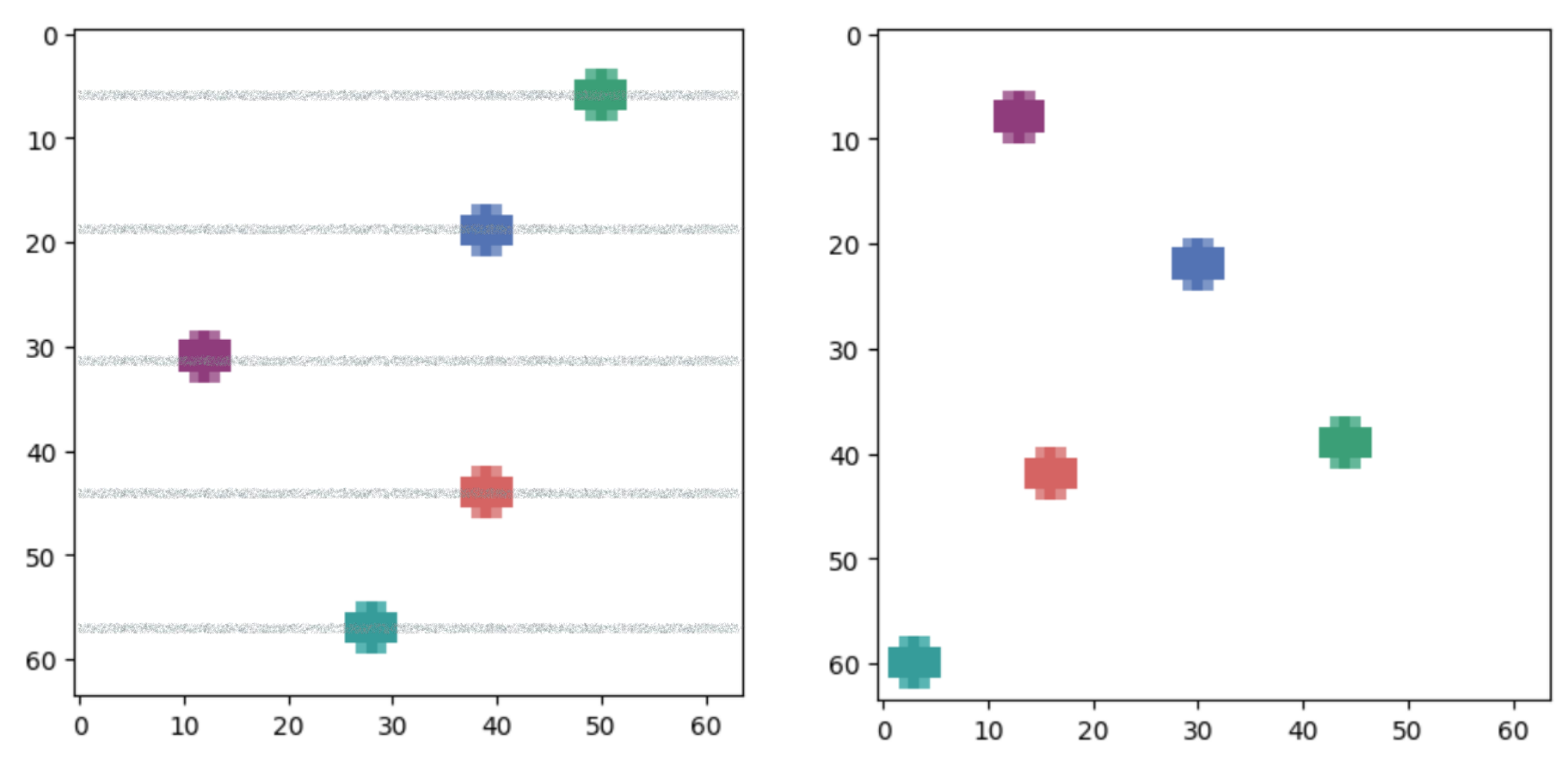}
    \caption{Example images of two settings in multiple ball dataset, based on the rendering code provided by \citep{ahuja2022weakly}. (left) \emph{Missing ball} setting: balls can move along gray paths only and are not visible when they move out of view. (right) \emph{Masked position} setting: balls can move freely inside the frame and their coordinates are masked when they have a specific value $\Mb$; e.g., the left-bottom position represents a masked version of the $x$ and $y$ coordinate for the olive ball.
    }
    \label{fig:balls}
\end{figure}

\paragraph{Data generation process}We use PyGame~\citep{pygame} to render images with size $64 \times 64 \times 3$ as shown in \cref{fig:balls}. The number of balls $b$ varies from 2 to 8. 
 
For the missing ball setting, we generate the $x$-coordinates of each ball $\Cb_j$ for $j\in[b]$ from a truncated normal distribution $\Ncal(0.5,0.1^{2})$ with bounds $(0.1,0.9)$. For the masked position setting, we generate the $x$ and $y$-coordinates of each ball $\Cb_i$ for $j\in[b]$ independently from a truncated 2-dimensional normal distribution $\Ncal(\mub_j,\Sigmab_j)$ with bounds $(0.1,0.9)^2$, where $\mub_i$ is generated from Unif$(0.4,0.6)^2$, and $\Sigmab_j= ((0.01,0.005)(0.005,0.01))$ for all groups. All the aforementioned parameters, including mean and variance/covariance above, are configured to ensure the majority of the samples are located in interval $[0,1]$. This configuration aims to maintain the truncated distribution as close to a Gaussian distribution as feasible.
 
For both settings, we generate the masked causal variables as $(\Zb_k)^K_{k=1} = \yb \cdot \Cb + (1- \yb) \cdot \Mb \in \RR^{K \times n}$. For the missing ball setting, the latent size is equal to the number of balls $\bb$. Therefore, we predefine $b$ masks via strategy ii) in App.~\ref{app: masks}. The mask value is set as $\mathbf{0}$. For the masked position setting, the latent size is $2b$ since both $x$ and $y$-coordinates of each ball are considered. The masked positions for distinct balls are different; avoiding overlap happens if two balls are masked in the same group. The mask value of each latent is predefined by one of these values \{0.05, 0.1, 0.9, 0.95\}.

\FloatBarrier

\subsection{Image dataset: Causal3DIdent}
\label{app:experiments_causal3DIdent}

\begin{figure}[ht]
    \centering
    \includegraphics[width=.6\linewidth]{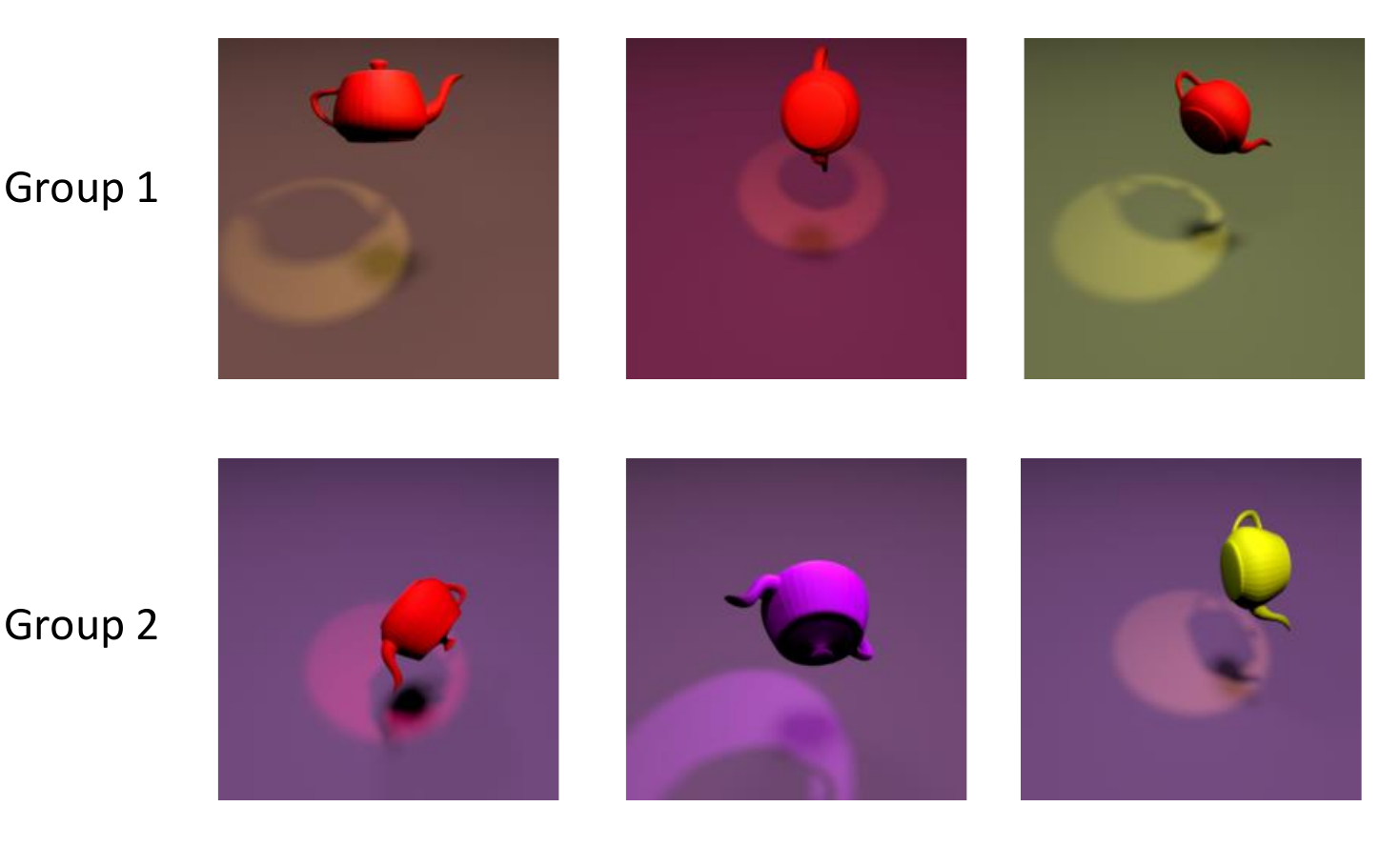}
    \caption{Example images from two groups in PartialCausal3DIdent dataset, resampled from Causal3DIdent datasert provided by \citep{von2021self}. In Group 1, the variable \emph{object hue} is masked to a constant. In Group 2, the variable \emph{background hue} is masked to a constant.}
    \label{fig:partialcausal}
\end{figure}

\begin{figure}[ht]
    \centering
    \includegraphics[width=.4\linewidth]{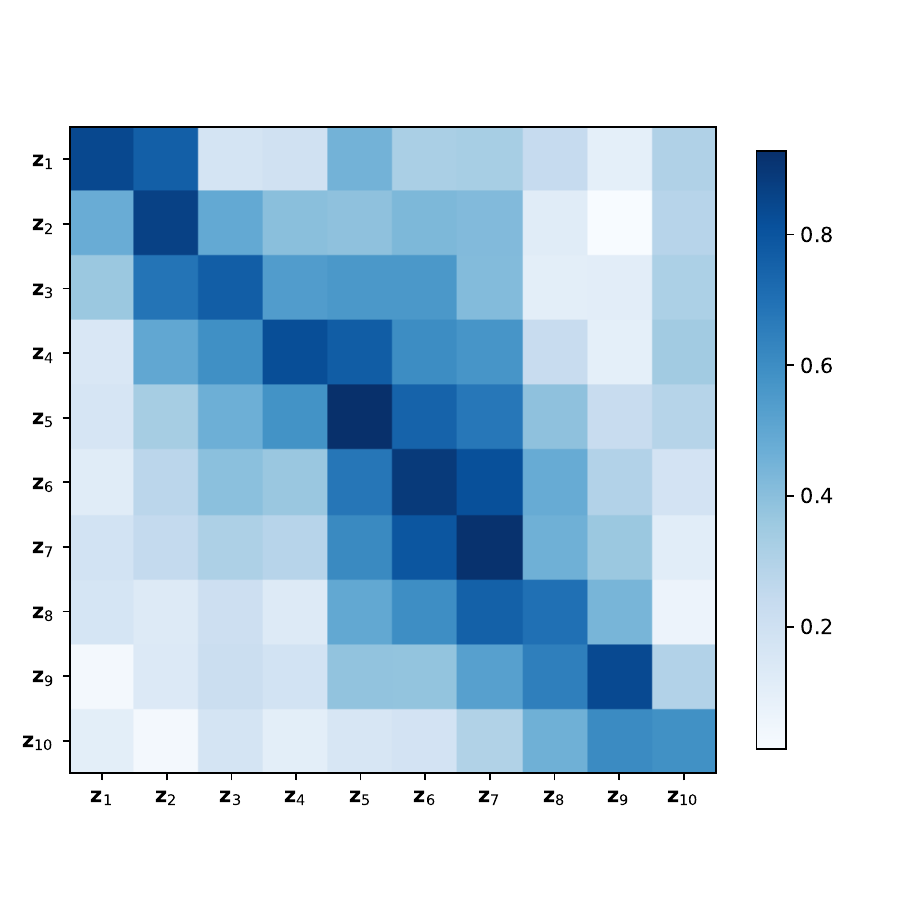}
    \caption{Pearson correlation matrix $\mathrm{Corr}^{n\times n}_{\pi}$ with the permutation $\pi$ between estimated and ground truth latent variables for PartialCausal3DIdent with masking distance $\delta=10\sigma$ for all latent variables. 
    }
    \label{fig:cor_matrix_c3di}
\end{figure}

A qualitative visualization of the result for one object class is presented in Fig.~\ref{fig:cor_matrix_c3di} where estimated and ground truth latent are well aligned on the diagonal. 

Since as we show in \cref{tab:c3di_res}, we did not observe significant performance differences between different object classes, we conduct an ablation study of the masking value $\delta$ only on \textbf{\textsc{Object Class ID} 0}. We show the results in \cref{tab:c3di_delta}. Consistent with the observation in numerical experiments, when we increase the value of $\delta$, which measures the distance between mask value and original mean, our method achieves better identification.

\begin{table}[ht]
    \centering
    \caption{MCC on \textbf{\textsc{object class ID=0}} regarding various masking values $\delta$ with standard deviation $\sigma=0.1$. Results averaged over three random seeds.}
    \begin{sc}
    \begin{small}
    \begin{tabular}{c|cccc}
    \toprule
     $\mathrm{\delta}$  & 0  & $2\sigma$ & $3\sigma$ & $5\sigma$  \\
     \textbf{MCC} (mean $\pm$ std)  & $0.586 \pm 0.056$ & $0.632\pm 0.044$ & $0.700\pm 0.039$  & $0.828\pm 0.011$\\
    \bottomrule
    \end{tabular}
    \end{small}
    \end{sc}
    \label{tab:c3di_delta}
    \end{table} 

\section{Image credits}

The images in Fig.~\ref{fig: example} are adapted from a 
\href{https://vectorportal.com/vector/parking-area.ai/23172}{VectorPortal image}
that is covered by CC BY 4.0. We segmented the cars, and removed some of them or moved their position in the image.

\end{document}